\documentclass{article}

\PassOptionsToPackage{numbers,sort&compress}{natbib}
 \usepackage[preprint]{neurips_2026}

\usepackage[utf8]{inputenc} 
\usepackage[T1]{fontenc}    
\usepackage{hyperref}       
\usepackage{url}            
\usepackage{booktabs}       
\usepackage{amsfonts}       
\usepackage{nicefrac}       
\usepackage{microtype}      
\usepackage{xcolor}         
\usepackage{graphicx}
\usepackage{subcaption}
\usepackage{amsmath}
\usepackage{amssymb}
\usepackage{placeins}

\title{Emergent Causal–Geometric Dynamics \\ Across Depth in Large Language Models}

\author{%
  Shahar Haim, Daniel C McNamee
   \\
  Champalimaud Centre for the Unknown,\\
  Champalimaud Foundation,\\
  Lisbon, Portugal \\
  \texttt{\{shahar.haim,daniel.mcnamee\}@research.fchampalimaud.org} \\
}

\begin{document}

\maketitle

\begin{abstract}

Geometric analyses of large language model (LLM) representations reveal structured variation across depth but remain fundamentally correlational with respect to token prediction formation. Meanwhile, causal interventions expose depth-dependent efficacy profiles without a unifying account of their representational dynamics. A complete account of LLM function requires explaining how representational structure evolves across depth to causally produce predictions. We synthesize these perspectives by combining geometric analysis with mechanistic interventions, explicitly centralizing depth-wise dynamics as the organizing axis for interpreting LLM function. In decoder-only LLMs, we identify a sharp transition from context-processing to prediction-forming computation, accompanied by a more gradual reorganization of representational geometry across layers. This synthesis reveals a late-layer geometric code in which angular structure parameterizes next-token distributional similarity and enables selective causal control over predictions, while representation norms encode information largely decoupled from prediction. Together, our results provide a synthesis of causal and geometric perspectives, yielding a mechanistic account of how control-relevant geometric dynamics across depth transform context into prediction in language models. This perspective reconciles previously puzzling findings and implies that layer-wise function cannot be understood or effectively intervened upon in isolation, but only within the emergent global dynamical structure of the network.

\end{abstract}



\section{Introduction}

Ongoing efforts seek to understand how large language models (LLMs) implement their remarkable predictive capabilities \citep{bubeck2023sparksartificialgeneralintelligence}. Advancing this understanding has important implications for the control of LLMs and for developing new techniques to enhance model performance. Two broad approaches have emerged. One line of work focuses on geometric analyses of representational flow through the residual stream, characterizing how token representations evolve across depth using measures such as similarity \citep{ethayarajh2019contexual,barbero_transformers_2024}, dimensionality \citep{cheng_emergence_2025,skean_layer_2025,sarfati_lines_2024}, curvature \citep{hosseini_straighten_2023,skean_layer_2025}, and isotropy \citep{cai_isotropy_2021,razzhigaev_shape_2024}. This geometric perspective has revealed systematic depth-dependent structure and increasingly abstract representations, but it primarily characterizes how context-dependent representations evolve across depth rather than which geometric features directly govern the output prediction distribution. Thus, this line of work provides increasingly detailed accounts of depth-wise representational dynamics, but remains largely \textit{context-centric}, characterizing how contextual information is encoded across layers while leaving open which aspects of this geometry become mechanistically relevant for next-token prediction.

A complementary line of work adopts a mechanistic, intervention-based perspective, seeking to understand LLM computation by perturbing internal representations and observing changes in predicted output token distributions. These approaches directly target the functional role of internal states in shaping model behavior and have revealed depth-dependent sensitivity to different classes of interventions \citep{turner_steering_2023, engels_not_2025, yu_directions_2025}. However, intervention-based studies typically operate without an explicit representational framework, leaving unclear how the manipulated representations are geometrically organized or which representational degrees of freedom are causally relevant beyond linear separability \citep{mikolov-etal-2013-linguistic, conneau2018probing_sentence_embeddings, saglam_large_2025}. Recent work has begun to use representational geometry to design more targeted interventions \citep{Balest_model_geometry,shafran2026directionsregionsdecomposingactivations,vu2026angular}; nevertheless, it remains unclear how the relevant geometry changes across depth, and why interventions become effective in some layers but not others. Thus, integrating across these lines of research, the key unresolved issue is how the geometric structure of internal representations is re-organized across network depth in order to causally shape prediction.


This distinction is important since representational structure can be descriptively accessible and correlational without being causally operative. Probe techniques \citep{alain2016understanding, conneau2018probing_sentence_embeddings}, for example, can reveal information that is linearly decodable from internal representations, but such information need not be causally involved in how the network generates its output \citep{ravichander-etal-2021-probing, hewitt-liang-2019-designing, elazar-etal-2020-amnesic-probing}. The same issue applies to geometric analyses: the presence of semantic directions, subspaces, or similarity relations does not imply that these geometric degrees of freedom are used to form predictions. Thus, geometric characterization alone is insufficient for mechanistic interpretation. What is needed is an account of what geometric structure is causally relevant to prediction and how it evolves across depth.

We address this challenge by combining geometric analysis with causal interventions in order to specifically target what we term \emph{mechanistic geometry} in LLMs: representational geometry whose structure is directly tied to causal control over prediction. Using this integrated approach, we show that, in the tasks and models studied here, decoder-only LLMs exhibit a sharp depth-wise transition from context-processing to prediction-forming computation. Following this transition, representation space is gradually reorganized into a prediction-centric geometry: angular structure becomes aligned with similarity between next-token distributions and supports selective causal control over predicted token identity, whereas representation norms carry variation that is largely decoupled from prediction similarity. These results show that the transition to prediction-forming computation is not merely a shift in representational content, but the emergence of a distinct geometric structure that directly support causal control over token prediction.

\subsection{Related Work and Background}
We have introduced a growing body of work that has documented systematic depth-dependent changes in the representations of LLMs, characterized by changes in dimensionality, geometry, and decodability \citep{ethayarajh2019contexual, cai_isotropy_2021, hosseini_straighten_2023, sarfati_lines_2024, barbero_transformers_2024, razzhigaev_shape_2024, skean_layer_2025}. However, this literature is almost exclusively input context-centric and does not relate geometric structure to prediction behavior of LLMs.





One study \citealp{lad_remarkable_2025} proposes a functional distinction between early context-centric and later prediction-centric computation in LLMs, based on analyses of attention patterns, single-neuron activity, and non-specific whole-layer ablations, but does not examine representational geometry. More direct causal evidence comes from targeted intervention experiments. We suggest that a retrospective comparison of intervention-based studies indicates a consistent, previously unarticulated, dichotomy whereby interventions derived from input context representations are most effective in early layers \citep{turner_steering_2023, engels_not_2025} and interventions constructed from output prediction-derived representations are most effective in later layers \citep{yu_directions_2025, vu2026angular}. This points to an emergent global functional structure through LLM depth and raises the question: what are the causal-geometric principles by which context representations re-organize into prediction representations? Although the causal role of representational geometry is highlighted by recent intervention methods \citep{Balest_model_geometry,shafran2026directionsregionsdecomposingactivations}, these are applied in a depth-agnostic approach overlooking such global emergent depth-wise dynamics. Our synthesized hypothesis suggest that understanding how representational geometry evolves across depth is fundamental both for explaining layer-specific intervention success and for developing a mechanistic account of evolving computation in LLMs. We therefore examine these depth-wise geometric dynamics and their causal role in model behavior.

\section{Methodology}
\label{methodology}
\paragraph{Models.}
Throughout this study we use three widely used open-source decoder-only LLMs. These models represent different architectural families while having comparable parameter scales: Llama V3.1 8B \cite{dubey_llama3_2024}, Mistral 0.3 7B \cite{jiang_mistral7b_2023}, and Qwen 2.5 7B \cite{qwen_qwen25_2025}.

\subsection{Intervention analyses}
\label{sec:int_setup}
We apply causal interventions to three modular structured tasks in which we have complete experimental control with well-defined finite input and output spaces: Months of the year, Days of the week, and Clock Math. We introduce the intervention setup using the Months task as the primary example, and provide the corresponding details and results for the Days and Clock Math tasks in Appendix~\ref{appendix_interventions}.

\paragraph{The Months task.}
We use the Months task as introduced by Ref.~\citealp{engels_not_2025}: "\textit{Let’s do some calendar math. [\textit{INTERVAL}] months from [\textit{MONTH}] is}" , where \textit{INTERVAL} takes values from \{One, Two,..., Twelve\}, and \textit{MONTH} can be any month of the year. In total, this task has 144 prompts that correspond to every combination of \textit{INTERVAL} and \textit{MONTH}. To simplify, for each prompt we read out only the logits of the 12 month of the year and choose the largest one to be the output. The modular arithmetic design of this task creates a closed input-output space where each prompt maps to a single output, but a single output is appropriate for multiple input prompts. Using the abstracted notation from Ref.~\citealp{engels_not_2025} we can define an input-output pair as $\alpha + \beta = \gamma$ where $\alpha + \beta$ corresponds to the input (starting month and interval), and $\gamma$ corresponds to the output prediction (target month). In this task the models accuracies are: Llama 95\%, Mistral 85\% and Qwen 70\% .

\label{Interventions}
\paragraph{Input-centric intervention.}
An \textit{input-centric intervention} alters the input representation by changing the hidden-layer activations of specific input tokens in the context window (for a schematic depiction, see Appendix~Fig.~\ref{fig:intervention-visual}). In the Months task, this corresponds to perturbing the representation from a given input of the form $\alpha + \beta$ toward $\alpha' + \beta$ or $\alpha + \beta'$, thereby effectively biasing the model toward a new target prediction $\gamma'$.

We calculate steering vectors $V_{L}(\alpha,\alpha')$ per layer $L$ based on centroid differences between token representations. For example, for the tokens corresponding to a source $\alpha$ and a target $\alpha'$, we compute the centroid of all hidden-layer vectors corresponding to occurrences of $\alpha$ and $\alpha'$ in the prompts at a given layer $L$. The steering vector from $\alpha$ to $\alpha'$ at layer $L$ is then defined as
\begin{equation}
V_{L}(\alpha,\alpha') \;=\; \mathbb{E}_{t \sim \alpha'}\!\left[h_{L}(t)\right] \;-\; \mathbb{E}_{t \sim \alpha}\!\left[h_{L}(t)\right] \quad,
\label{eqn:steer_vec}
\end{equation}
where $h_{L}(t)$ denotes the hidden representation of token $t$ at layer $L$. We apply the intervention by adding this vector to the hidden representation of the corresponding input token at layer $L$. To measure the effect of such intervention we quantify the change in models prediction preference between the new and old targets: 
\begin{equation}
    C(\gamma, \gamma') = \left[\operatorname{Logit}_{\text{after}}(\gamma') - \operatorname{Logit}_{\text{after}}(\gamma)\right] - \left[\operatorname{Logit}_{\text{before}}(\gamma') - \operatorname{Logit}_{\text{before}}(\gamma)\right] \quad .
\label{eqn:logit_measure}
\end{equation} If the intervention had no effect we expect a value of zero, and a positive value if the intervention shifted the preference towards the new target.

\paragraph{Output-centric intervention.}
An \textit{output-centric intervention} attempts to directly influence the model’s output prediction by intervening on the hidden-layer activations of the final token, whose representation is used to produce the output. Unlike input-centric interventions, this method steers the model’s output toward a desired target by using representations associated with that outcome. In the Months task, this corresponds to shifting the representation of the last token from one associated with a target prediction $\gamma$ toward that of a different prediction $\gamma'$ via $V_{L}(\gamma,\gamma')$ (Eqn.~\ref{eqn:steer_vec}).

Output-centric steering vectors are computed by taking centroid differences between last-token representations of prompts whose model prediction is $\gamma$ and those whose prediction is $\gamma'$. Concretely, for each layer $L$, we group prompts according to their baseline model prediction and compute the centroids of the hidden representations of the final token for prompts whose prediction is $\gamma$ and those whose prediction is $\gamma'$. The output-centric steering vector is then defined as the difference between these centroids. Unlike the input-centric intervention, which isolates specific context properties, this method directly targets prediction representations. As in the input-centric case, the intervention is applied via vector addition, but here the steering vector is added only to the final token. The effect of this intervention is also quantified by the change in the model's prediction behavior (Eqn.~\ref{eqn:logit_measure}).

\begin{figure}[!tbp]
  \centering
  

  \begin{subfigure}[t]{0.31\textwidth}
    \centering
    \includegraphics[width=\linewidth]{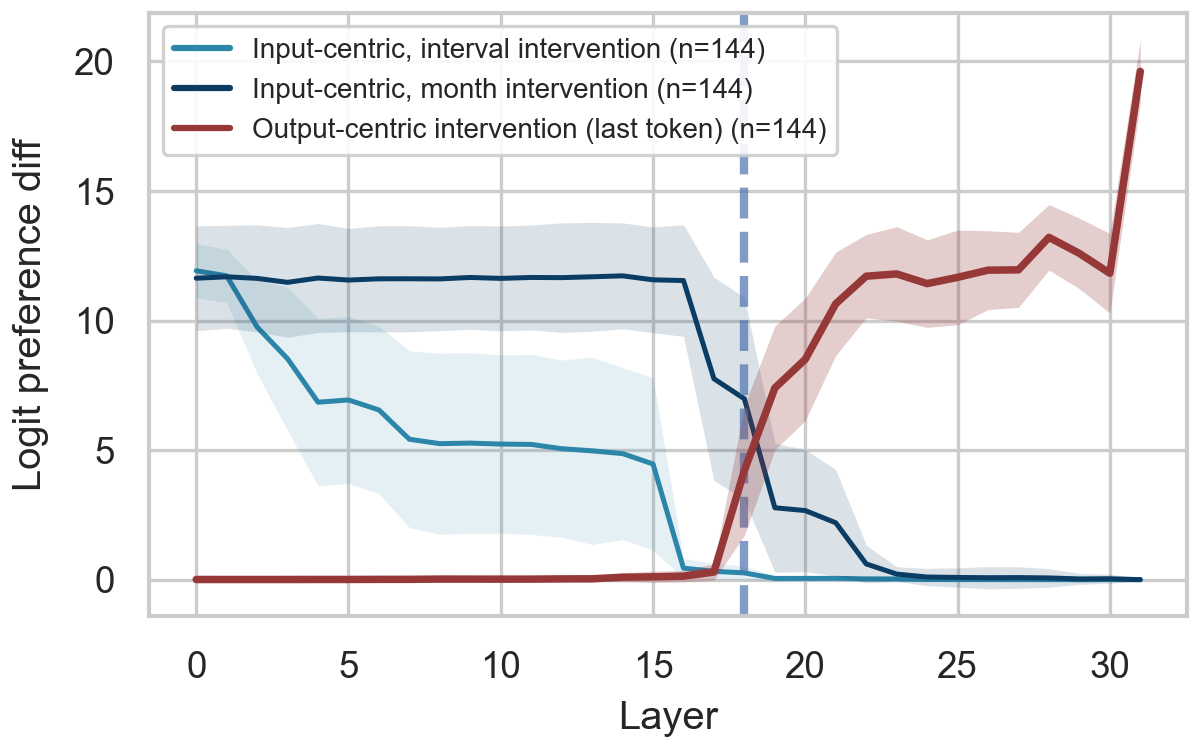}
    \caption{Llama}
    \label{fig:int-llama}
  \end{subfigure}\hfill
  \begin{subfigure}[t]{0.31\textwidth}
    \centering
    \includegraphics[width=\linewidth]{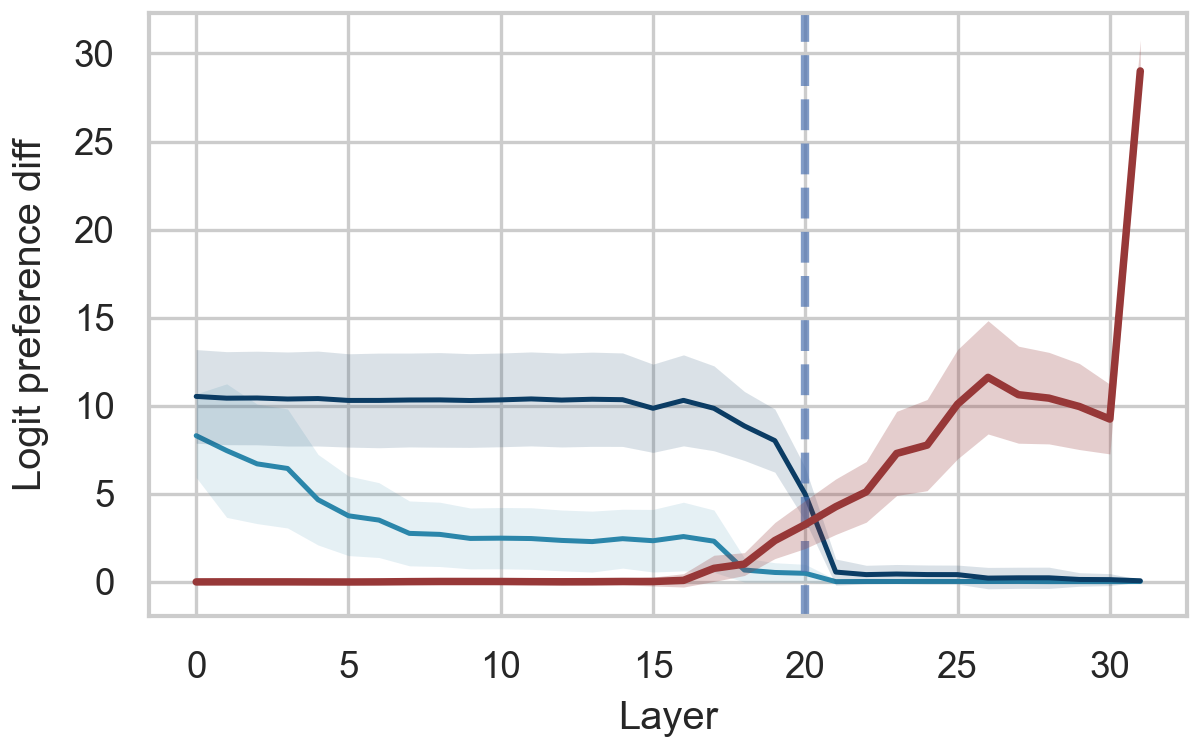}
    \caption{Mistral}
    \label{fig:int-mistral}
  \end{subfigure}\hfill
  \begin{subfigure}[t]{0.31\textwidth}
    \centering
    \includegraphics[width=\linewidth]{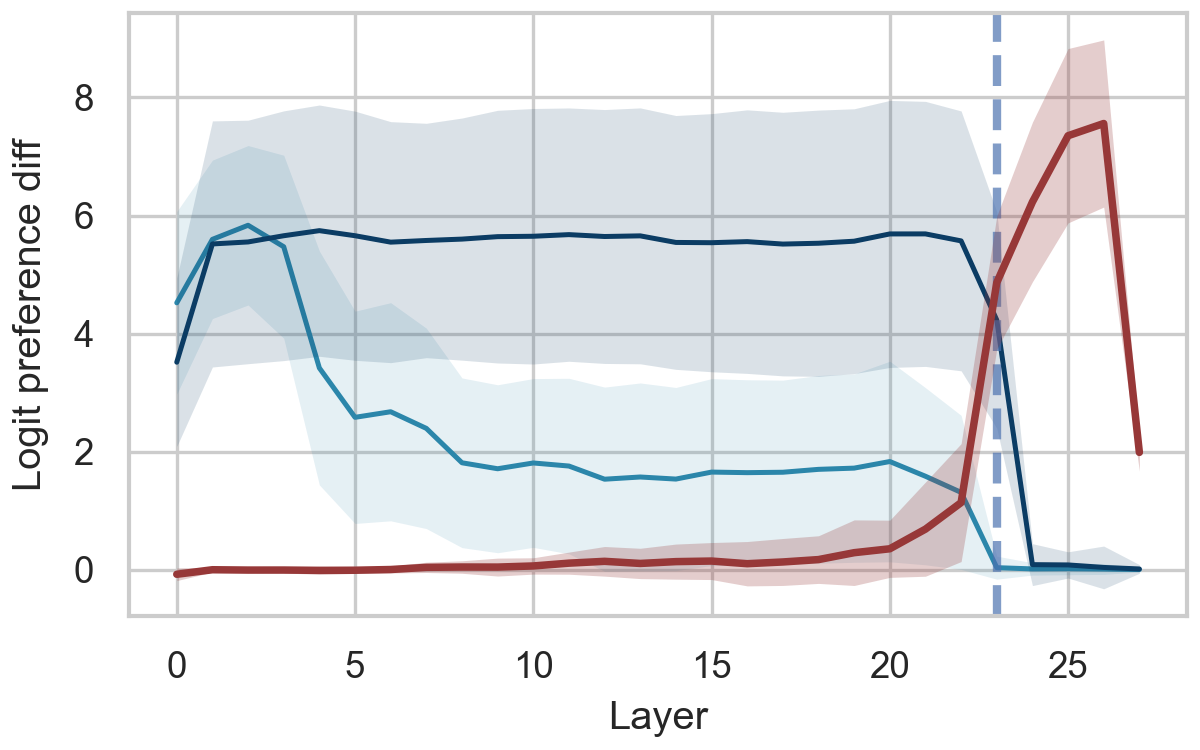}
    \caption{Qwen}
    \label{fig:int-qwen}
  \end{subfigure}
  
  \caption{Layer-wise interventions. Per-layer intervention experiment results. Average logit preference difference: blue curves show input-centric interventions, interval and month; the red curve shows output-centric intervention. The blue vertical line marks the phase-change point. Shading marks ± Standard Deviation. See visual schematic in the appendix Fig.~\ref{fig:intervention-visual}.}
  \label{fig:intervention}
\end{figure}

\label{Representation analyses}
\subsection{Representation analyses}
\label{sec:rep_analyses}
\paragraph{Dataset.}
To inspect the geometry of LLM representations across layers and characterize how they are transformed, we sought a larger and more diverse dataset, since experimental control was less critical than in the interventional experiments. Importantly, a broader and more heterogeneous data distribution enables a more detailed characterization of the representational geometry and its variation across layers. Therefore, we used the wiki-text-103-raw-v1 dataset \cite{merity_wikitext_2017} to collect two sets of sequences: Short context (15 tokens) and long context (500-600 tokens). Each of these sets will be used both as ordered and shuffled sequences. Each set consists of one thousand sequences. We chose our dataset based on length criteria as measured by the original tokenized sequence. For semantic completeness we require all sequences end with the dot at the end of a sentence, thus long contexts have a slightly variable length. We parse the text so that the end of every sequence is ``.'' to keep the final token identical across sequences. For shuffled sequences, we remove the final period, shuffle the words, and then add the period back, ensuring that all sequences end with an identical token. 

\paragraph{Identical vs. non-identical pilot tokens.} In the following, analyses of identical tokens refer to the final token of each sequence. Analyses of non-identical tokens refer to tokens located fourth from the end, which are close to the final token but naturally vary across sequences. We replicate the non-identical-token analysis for tokens located third from the end in the Appendix (Figs. ~\ref{fig:third_from_Normalized_PR}, ~\ref{fig:third_non_identical_correlations}), confirming that the results are not position specific. In the following, we refer to the inspected identical and non-identical tokens at the end of the sequence as \emph{pilot-tokens} ~\citealp{sarfati_lines_2024}. Their contextualized representations serve as proxies for sequence-level information.

\paragraph{Dimensionality analysis.}
For a set of tokens in some LLM layer we measure the distribution of variance across orthogonal dimensions by using the Participation Ratio (PR)\citep{multineuronal_dimensionality}: $\mathrm{PR} = \frac{\left(\sum_{i=1}^d \lambda_i\right)^2}{\sum_{i=1}^d \lambda_i^2}$ where $\lambda_i = \sigma^2_i$
and $\sigma_i$  are the matrix singular values. Here, specifically, we decompose matrices who's entries are the hidden layer activations of a set of tokens. Higher PR indicates that variance is distributed across a larger number of orthogonal directions. Due to the sensitivity of variance based decompositions to outliers, we perform the decomposition after normalizing each token vector. The normalized results preserve the same trends in data while attenuating outlier driven distortions. We provide the unnormalized analyses results in the appendix for comparison (Fig.~\ref{fig:PR}).

\section{Results}

\paragraph{Two phases of computation in LLMs.}
To study input context-centric and output prediction-centric representational dynamics we first seek to implicate representations across LLM layers as causally contributing to either. To do so, we intervene on hidden layer activations of three LLMs performing the Months task with two types of interventions: (1) an \textit{input-centric intervention} and (2) an \textit{output-centric intervention} (see Section~\ref{sec:int_setup}). For each prompt, we fix a source representation --- either the input month/interval or the model’s predicted output month --- and apply separate interventions toward each of the alternative centroids possible for that token, one target at a time. For example, the representation of the input month ``January'' is intervened upon and shifted towards each of the other eleven months using the steering vector $V_L(\text{``January''}, \text{``February''})$ (Eqn.~\ref{eqn:steer_vec}) for the target month ``February''.
Strikingly, across models, we find that applying a layer-wise \textit{input-centric} intervention to the starting month token has dominant effect during an early phase of the models. This early phase occupies roughly the first two-thirds of layers. In stark contrast, applying a layer-wise \textit{output-centric} intervention to the last token has a dominant effect during the late phase of the models, occupying roughly the last third of layers. To ensure the observed effects are unrelated to some coupling between the manipulated input tokens and predicted tokens (both being months), we apply the \textit{input-centric} intervention to the interval tokens, instead of the input month tokens. Again, the effect is observable during the early phase of the models (Fig.~\ref{fig:intervention}). In order to test whether this sharp depth-wise input-to-output transition is specific to the Months task, or is representative of a broader principle, we replicated our approach in two additional tasks, Days and Clock Math, and observed the same bi-phasic depth-wise structure suggesting that this is a general feature of LLM context-to-prediction processing (Appendix section~\ref{appendix_interventions}; Fig.~\ref{fig:add-intervention-extras}).

The intervention experiments reveal a clear functional dissociation across depth, with early layers selectively sensitive to input-centric perturbations and later layers selectively sensitive to output-centric perturbations. This pattern suggests that early layers perform computations that depend directly on contextual input representations, whereas, after a sharp depth-wise transition, later layers operate on transformed representations in a more prediction-oriented manner. To examine how this functional dissociation is reflected in the underlying geometric organization, we first analyze the dimensionality structure of token representations across layers in a more diverse dataset using the Participation Ratio.


\label{Context- and token-dependent dimensionality across layers}
\paragraph{Context- and token-dependent dimensionality across layers.}
To isolate the contribution of contextual variability to token representations, we compare the geometric structure of pilot-token representations between \textit{identical} tokens appearing at the end of different sequences and \textit{non-identical} tokens positioned nearby in the sequence (see Section~\ref{sec:rep_analyses}). In this setting, the participation ratio (PR) quantifies the effective dimensionality of representational variability across tokens, capturing how variation induced by context and token identity is distributed across dimensions. Such comparison contrasts the organizations of two different token groups that share highly similar (overlapping) context, but vary in their identity.

\begin{figure}[t]
  \centering

  \begin{subfigure}[t]{0.31\textwidth}
    \centering
    \includegraphics[width=\linewidth]{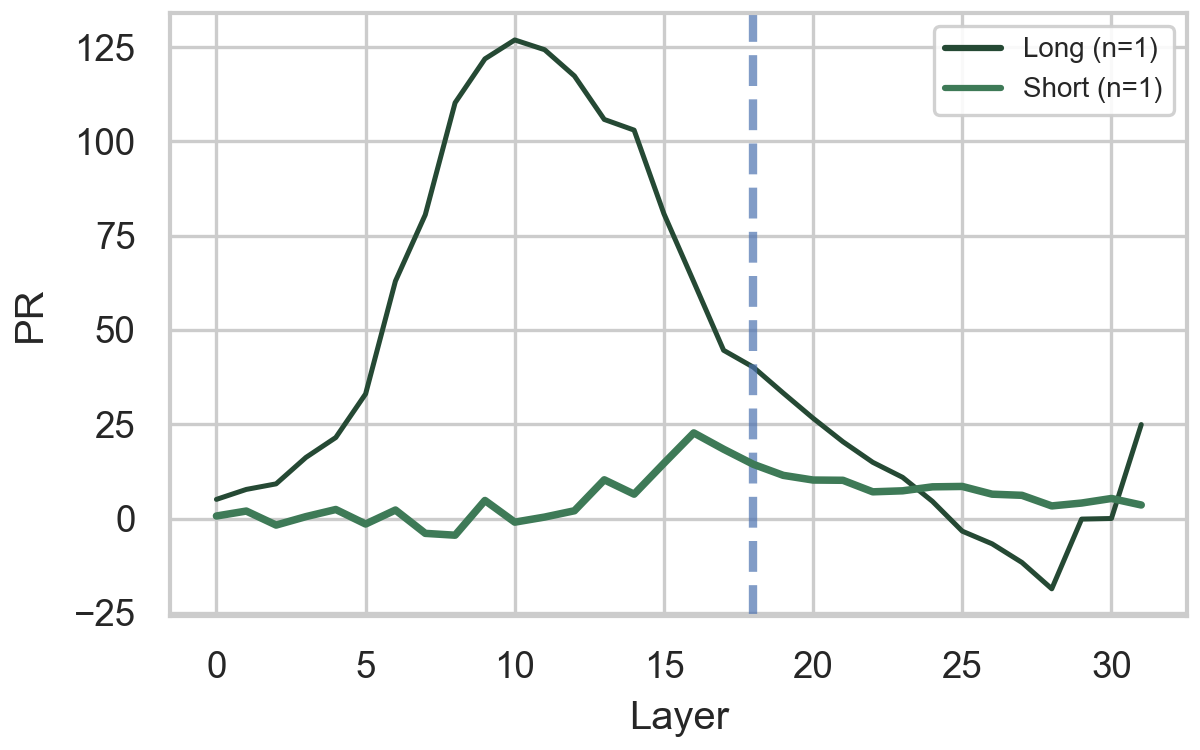}
    \caption{Llama: Identical tokens PR}
    \label{fig:pr-llama-identical}
  \end{subfigure}\hfill
  \begin{subfigure}[t]{0.31\textwidth}
    \centering
    \includegraphics[width=\linewidth]{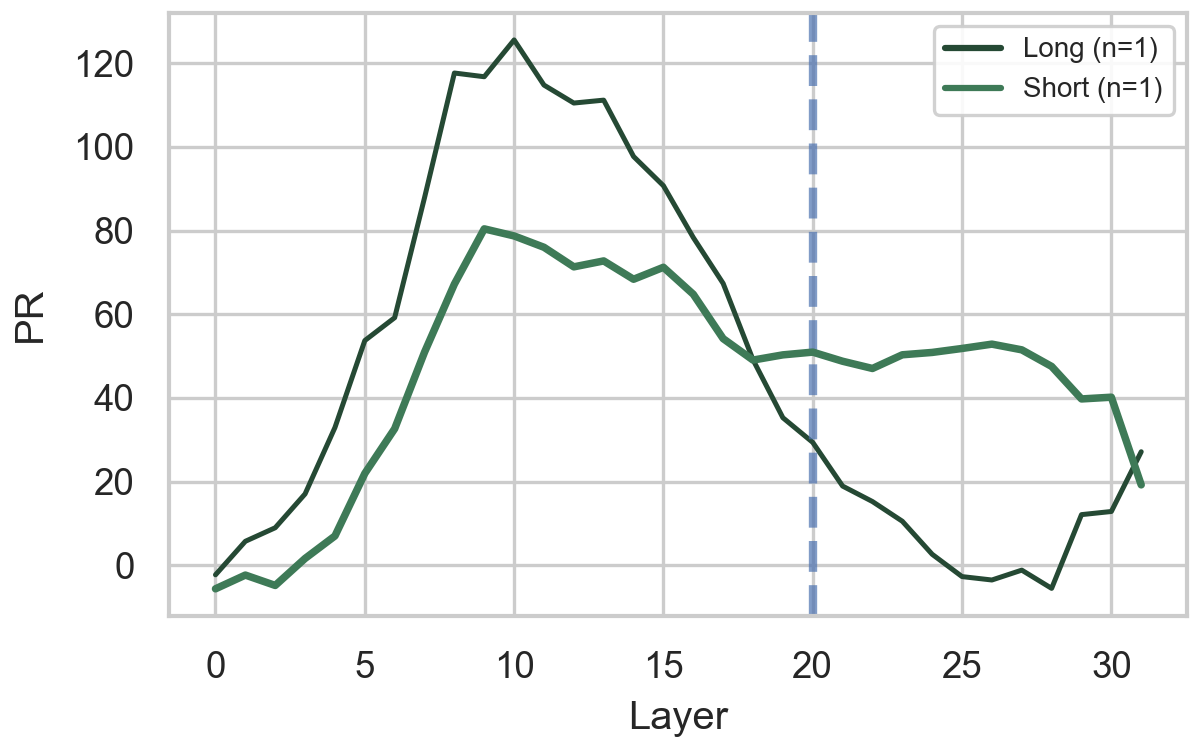}
    \caption{Mistral: Identical tokens PR}
    \label{fig:pr-mistral-identical}
  \end{subfigure}\hfill
  \begin{subfigure}[t]{0.31\textwidth}
    \centering
    \includegraphics[width=\linewidth]{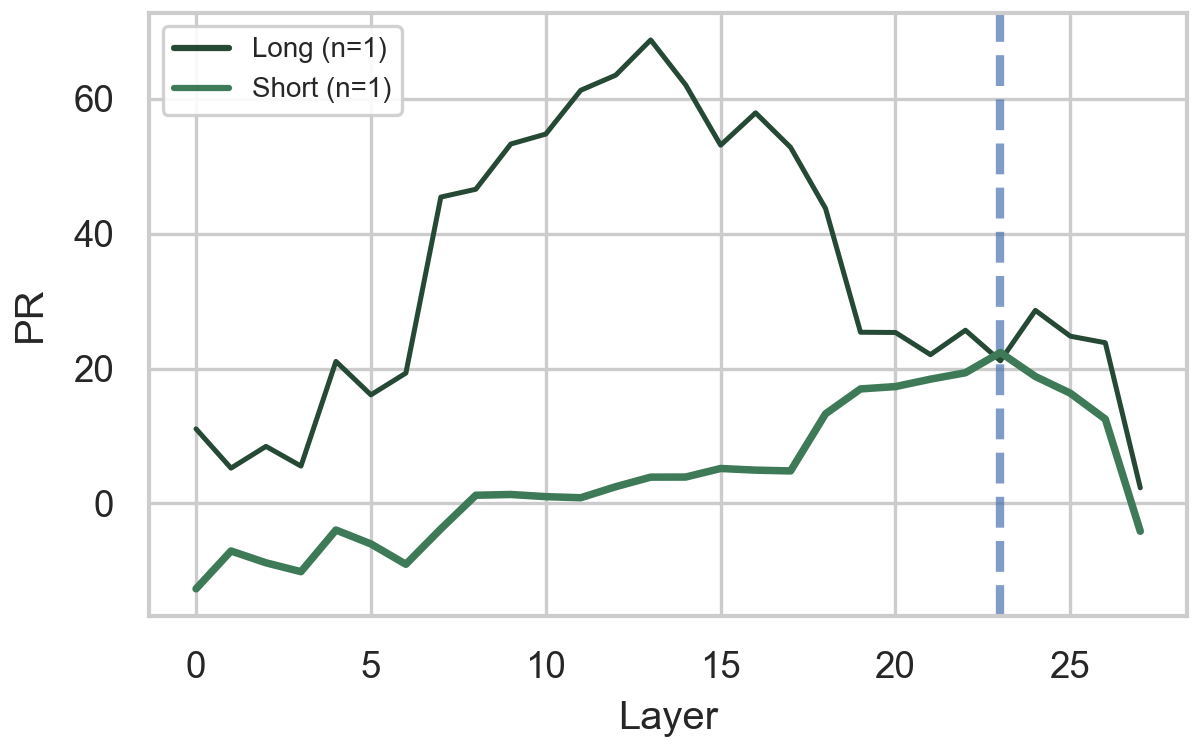}
    \caption{Qwen: Identical tokens PR}
    \label{fig:pr-qwen-identical}
  \end{subfigure}

  \vspace{0.6em}

  \begin{subfigure}[t]{0.31\textwidth}
    \centering
    \includegraphics[width=\linewidth]{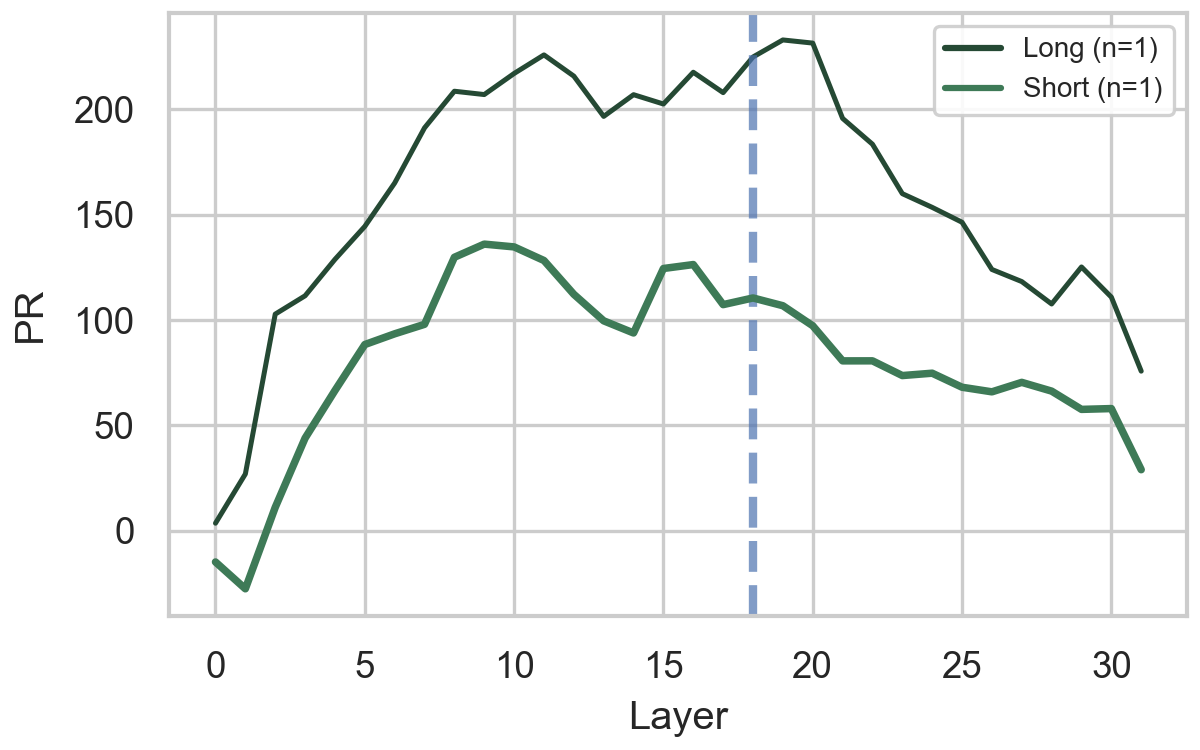}
    \caption{\footnotesize Llama: Non-identical tokens PR}
    \label{fig:pr-llama-nonidentical}
  \end{subfigure}\hfill
  \begin{subfigure}[t]{0.31\textwidth}
    \centering
    \includegraphics[width=\linewidth]{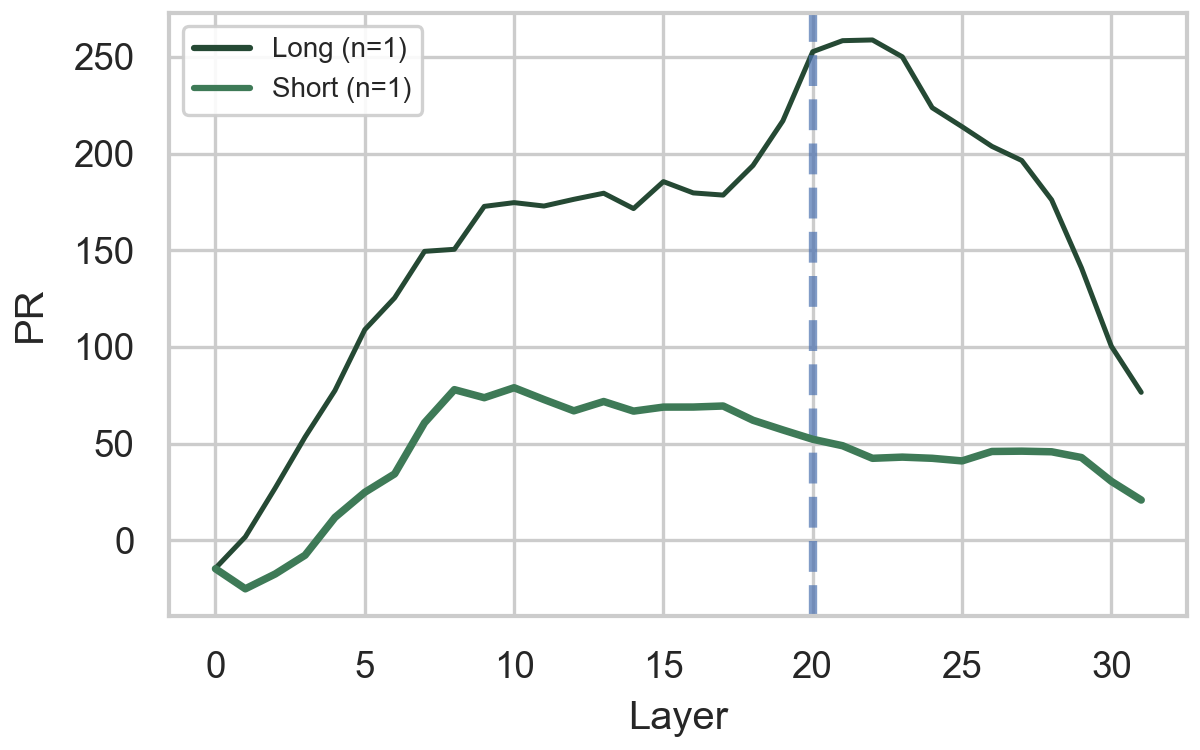}
    \caption{\footnotesize Mistral: Non-identical tokens PR}
    \label{fig:pr-mistral-nonidentical}
  \end{subfigure}\hfill
  \begin{subfigure}[t]{0.31\textwidth}
    \centering
    \includegraphics[width=\linewidth]{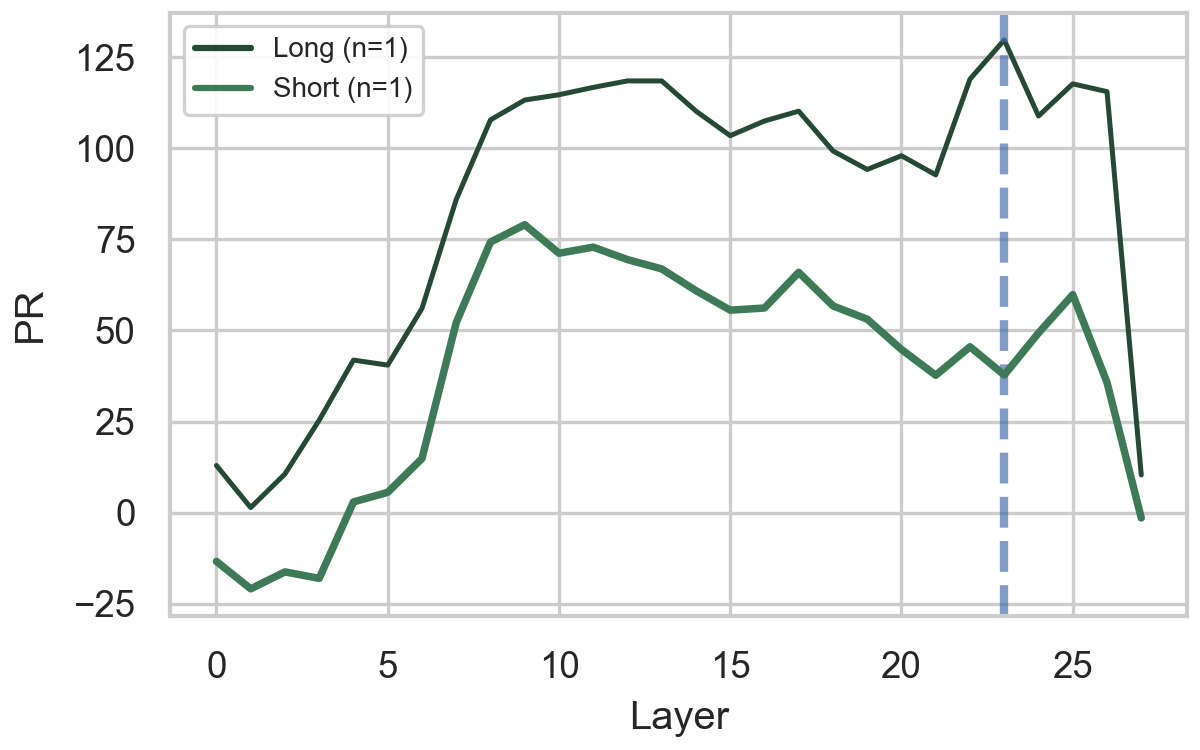}
    \caption{\footnotesize Qwen: Non-identical tokens PR}
    \label{fig:pr-qwen-nonidentical}
  \end{subfigure}

  \caption{
    Per-layer participation ratio (PR) of normalized pilot-token representations for ordered long and short sequences, shown after baselining by the corresponding shuffled-sequence condition (the original ordered and shuffled curves are shown in Fig.~\ref{fig:Normalized_PR})
    \textbf{Top row}: identical tokens.
    \textbf{Bottom row}: non-identical tokens.
    Columns correspond to models (Llama, Mistral, Qwen, left to right).
    The blue vertical line marks the perturbation-based phase-change point.
  }
  
  \label{fig:diff_Normalized_PR}
\end{figure}

We examine the dimensionality of the subspace occupied by these two token sets under four contextual conditions: long or short sequences that are either ordered or shuffled. These conditions allow us to independently control for context length and contextual coherence. For each condition, token set, and layer, we compute the PR for both unnormalized and normalized token representations. Normalization yields highly similar trends across layers while producing more stable and consistent PR estimates for Mistral. We therefore focus on normalized PR results in the main text and provide unnormalized results in Appendix~Fig.~\ref{fig:PR}. To more directly isolate the effect of contextual coherence, we further baseline the PR of ordered sequences by subtracting the corresponding shuffled condition, yielding a differential measure that emphasizes coherence-related differences (Fig.~\ref{fig:diff_Normalized_PR}). For completeness, the original normalized PR curves without baseline subtracting are shown in Appendix~Fig.~\ref{fig:Normalized_PR}.

A clear contrast emerges between the PR profiles of identical and non-identical tokens. Identical-token representations undergo an expansion followed by contraction of their effective dimensionality in the early layers, after which PR stabilizes with depth. In contrast, non-identical tokens exhibit broader and often multi-peaked PR profiles, reflecting variability driven jointly by token identity and contextual differences across layers. For both token sets, PR is modulated by contextual coherence, with ordered sequences tending to occupy higher-dimensional subspaces than shuffled ones (Fig.~\ref{fig:diff_Normalized_PR}; see also Figs.~\ref{fig:Normalized_PR}, \ref{fig:third_from_Normalized_PR}, \ref{fig:PR}), suggesting that coherent context induces richer representational variability.

As a control, we measure PR for short, ordered sequences in untrained models (Figs.~\ref{fig:Normalized_PR}, \ref{fig:PR}). These exhibit qualitatively different PR profiles for both identical and non-identical tokens, indicating that the observed patterns are learned properties of trained models rather than architectural artifacts.

These results reveal a depth-dependent reorganization of representational geometry consistent with the intervention findings. Early layers exhibit context-sensitive variability, particularly for identical tokens, while later layers exhibit more stable dimensional structure. Alongside the non-identical token results, this points to a geometric organization that depends jointly on context and token identity. In line with this, greater token diversity is associated with greater diversity in prediction distributions, reflected in increased pairwise dissimilarity between prediction distributions (Fig.~\ref{fig:pred_dist_sim}).

From a mechanistic geometry perspective, the goal is to characterize the relationship between representational geometry and the mechanisms of model prediction. Thus, we next compare embedding Euclidean and angular distances with distributional dissimilarities in token prediction distributions.

\paragraph{Two-component geometric coding in the prediction-centric phase.}

Previous analyses identified a context-dependent and token-sensitive spatial reorganization during the first, context-centric computational phase in LLMs. These reorganized representations are propagated to the subsequent, prediction-centric phase. To characterize how this geometry relates to model outputs, we analyze representations with respect to final prediction distributions using the same two sets of pilot tokens. 
For each layer and condition, we compute pairwise Euclidean and angular distances between all token representations within each pilot-token set, and quantify prediction differences by the symmetric Kullback-Leibler (KL) divergence between the corresponding output distributions. We then compute the layer-wise Spearman correlation between representational distance and prediction divergence.

\begin{figure}[t]
  \centering

  \begin{subfigure}[t]{0.31\textwidth}
    \centering
    \includegraphics[width=\linewidth]{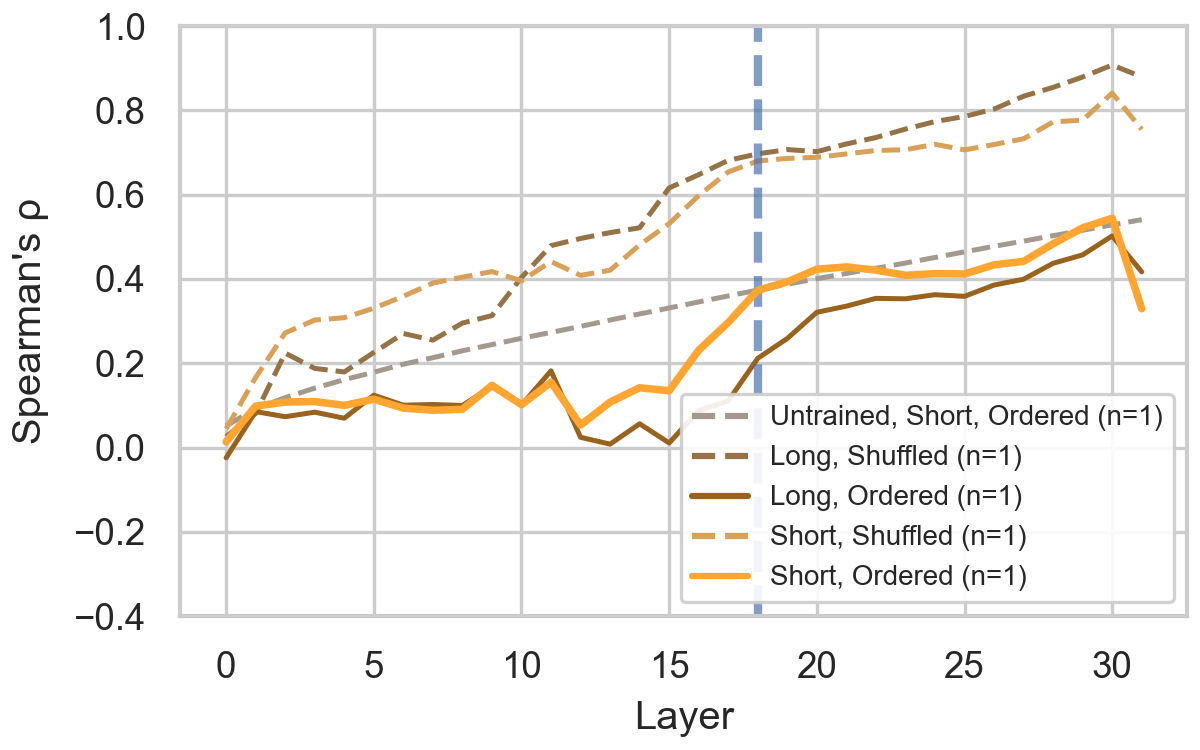}
    \caption{Llama: angular distances, predictions correlation}
    \label{fig:ang-llama-ang}
  \end{subfigure}\hfill
  \begin{subfigure}[t]{0.31\textwidth}
    \centering
    \includegraphics[width=\linewidth]{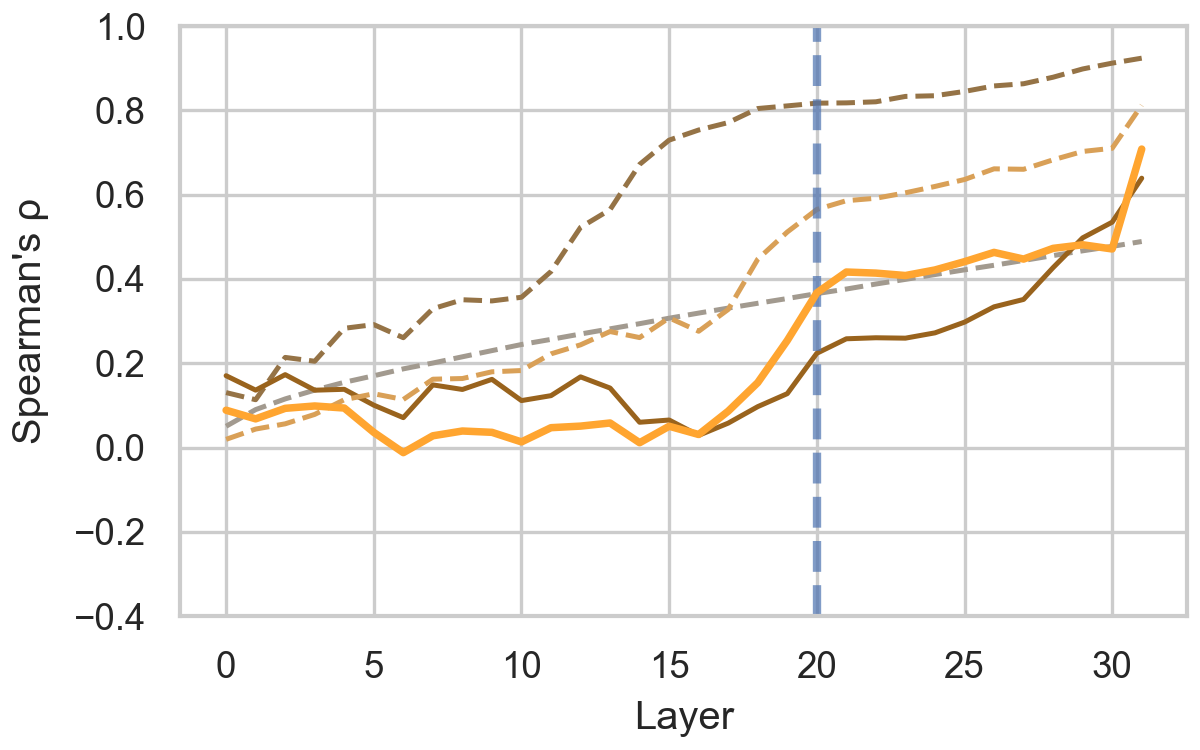}
    \caption{Mistral: angular distances, predictions correlation}
    \label{fig:ang-mistral-ang}
  \end{subfigure}\hfill
  \begin{subfigure}[t]{0.31\textwidth}
    \centering
    \includegraphics[width=\linewidth]{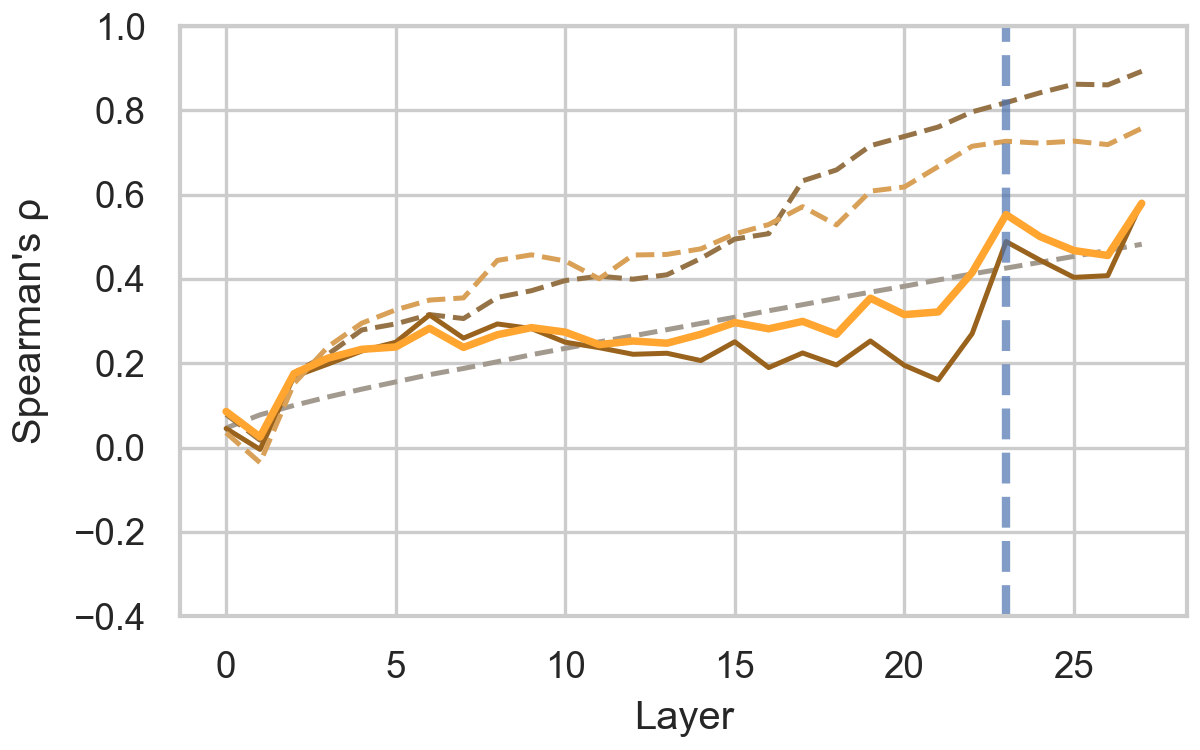}
    \caption{Qwen: angular distances, predictions correlation}
    \label{fig:ang-qwen-ang}
  \end{subfigure}

  \vspace{0.6em}

  \begin{subfigure}[t]{0.31\textwidth}
    \centering
    \includegraphics[width=\linewidth]{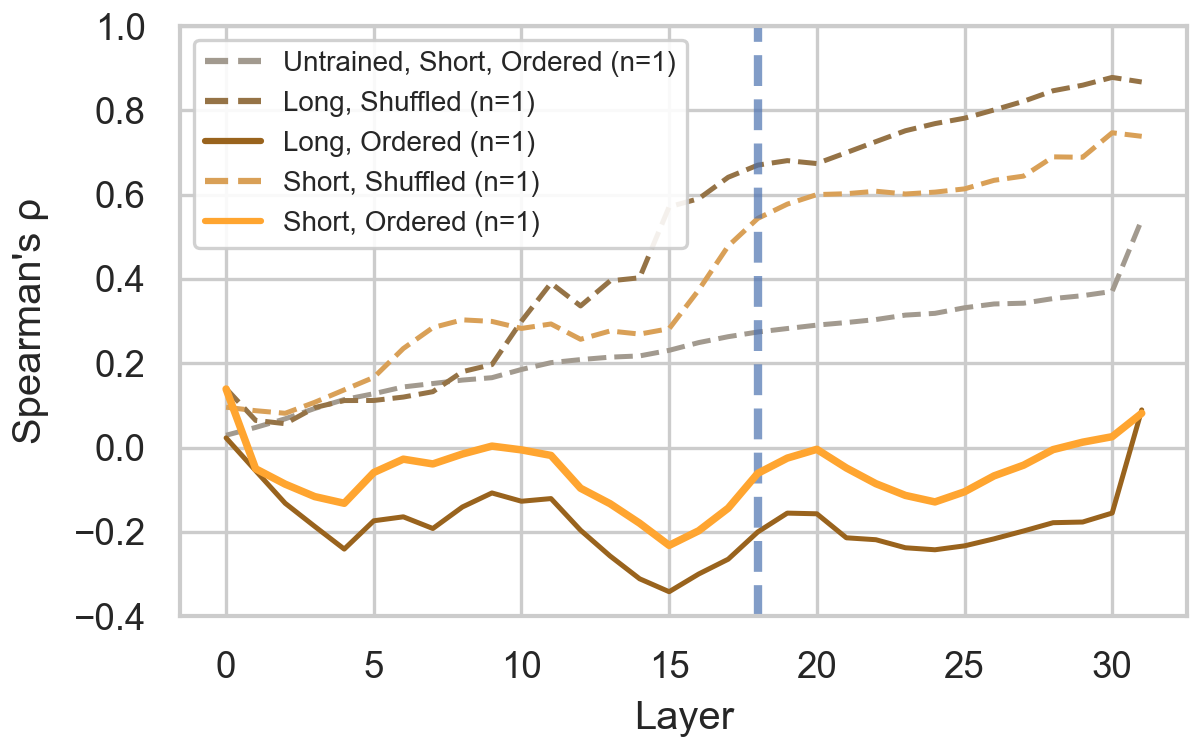}
    \caption{Llama: Euc. distances, predictions correlation}
    \label{fig:ang-llama-euc}
  \end{subfigure}\hfill
  \begin{subfigure}[t]{0.31\textwidth}
    \centering
    \includegraphics[width=\linewidth]{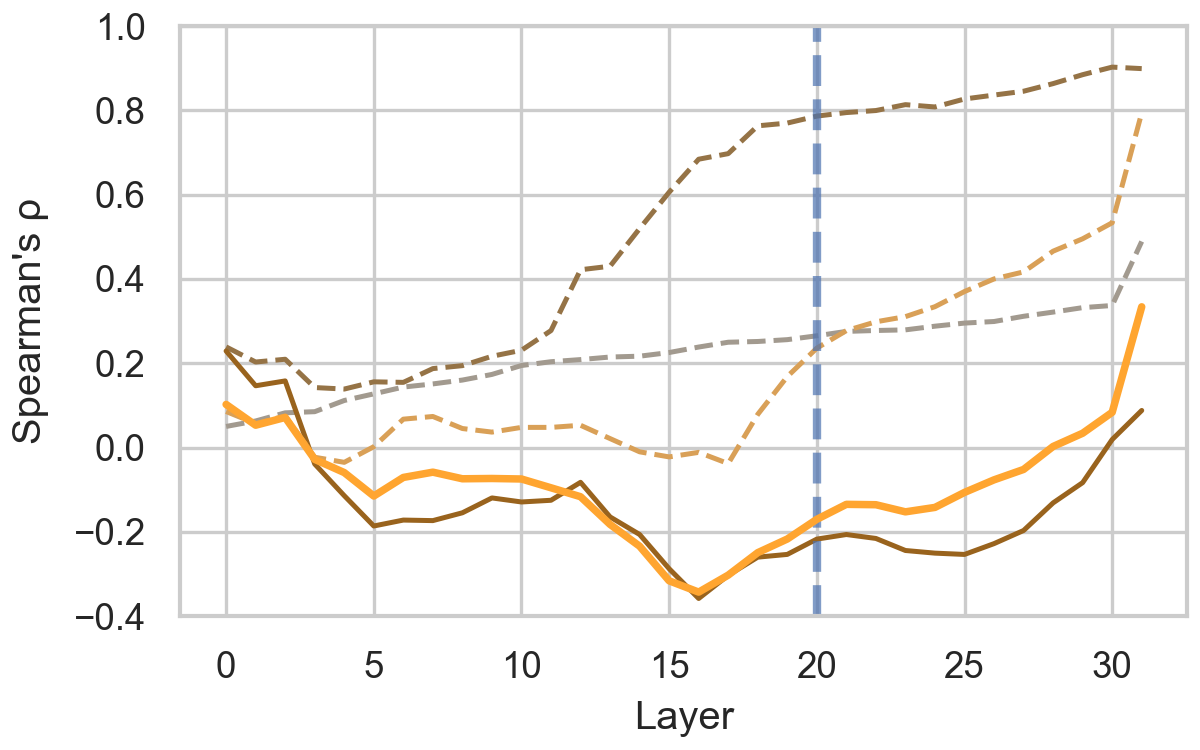}
    \caption{Mistral: Euc. distances, predictions correlation}
    \label{fig:ang-mistral-euc}
  \end{subfigure}\hfill
  \begin{subfigure}[t]{0.31\textwidth}
    \centering
    \includegraphics[width=\linewidth]{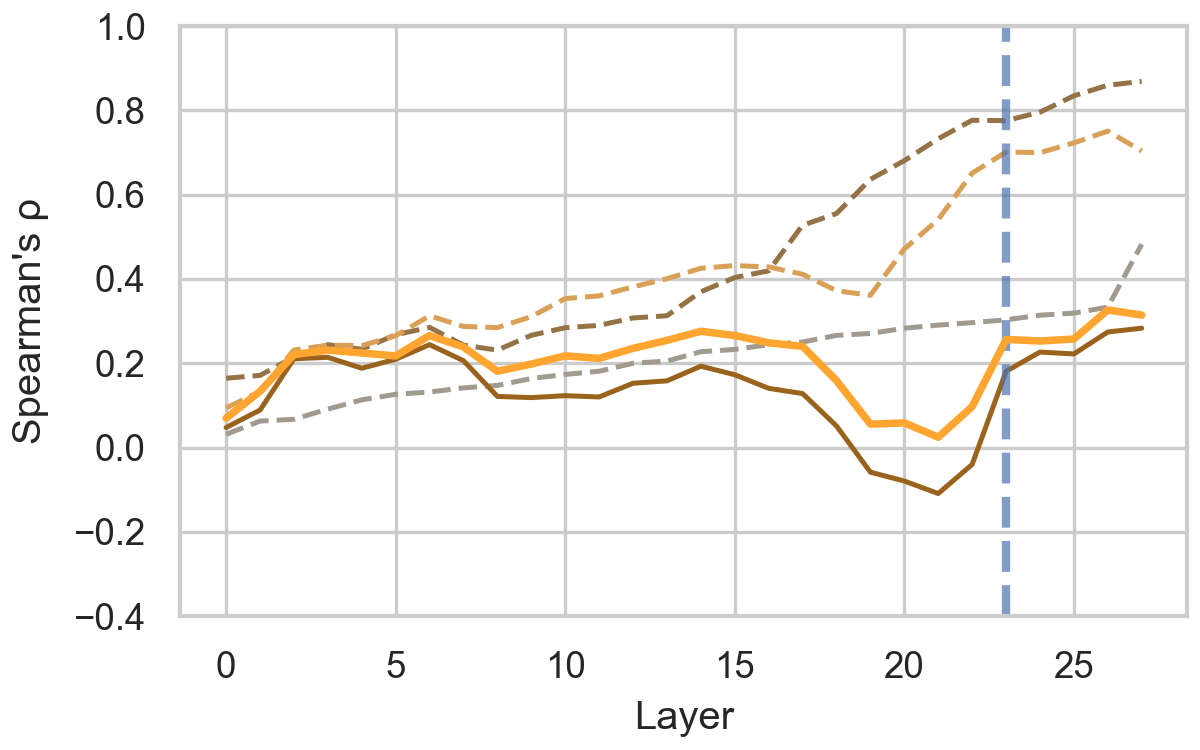}
    \caption{Qwen: Euc. distances, predictions correlation}
    \label{fig:ang-qwen-euc}
  \end{subfigure}

  \caption{
    Per-layer correlation between non-identical tokens' pairwise distances and pairwise prediction-distribution symmetric KL divergence, shown across layers.
    \textbf{Top row}: angular distances.
    \textbf{Bottom row}: Euclidean distances.
    Rows correspond to distance metrics, and columns correspond to models.
    The blue vertical line marks the perturbation-based phase-change point.
  }
  
  \label{fig:non_identical_correlations}
\end{figure}

For identical tokens, both Euclidean and angular distances exhibit gradually increasing positive correlations with prediction divergence across depth (Fig.~\ref{fig:identical_correlations}), indicating that geometric proximity increasingly mirrors prediction similarity as representations propagate. In contrast, for non-identical tokens this correspondence breaks down (Figs.~\ref{fig:non_identical_correlations}, ~\ref{fig:third_non_identical_correlations}). Angular distances in ordered sequences remain uncorrelated with prediction divergence in early layers, but begin to exhibit a clear positive correlation during the second, prediction-centric phase. This transition aligns with the depth at which output-centric interventions become effective, suggesting that first-phase reorganization induces angular geometry that becomes predictive of output distributions in the second phase.

By contrast, Euclidean distances for non-identical tokens show less stable relationships with prediction divergence, fluctuating between anti-correlation and de-correlation, where de-correlation trends are observed more dominantly during the second phase. Specifically, in the late layers, all three models exhibit an upward trend from anti-correlation toward de-correlation. This indicates that Euclidean proximity is largely orthogonal to predictive similarity in this regime. Although Euclidean $\|X - Y\|_2^2$ and angular distances $\theta$ are directly related, $\|X - Y\|_2^2 = \|X\|_2^2 + \|Y\|_2^2 - 2\|X\|_2\|Y\|_2\cos(\theta)$, a divergence between them reflects variability in vector norms, i.e. $\|X\|_2^2$ and  $\|Y\|_2^2$. The observed dissociation therefore indicates that, for non-identical tokens, norm fluctuations contribute strongly to Euclidean distances while remaining largely uncoupled from prediction similarity.

This interpretation is consistent with our earlier perturbation results. In that setting, Euclidean output-centric interventions applied to the same \textit{identical} final token were effective despite altering both norm and direction. Because the readout is taken from a fixed token whose angular alignment with prediction-relevant structure is preserved, simultaneous changes in norm and direction do not disrupt the underlying predictive geometry. The present analysis generalizes this observation by showing that, across non-identical tokens, predictive similarity in the prediction-centric phase is primarily indexed by angular organization rather than by overall Euclidean proximity. Results for shuffled sequences show no consistent angular–prediction alignment, serving as a negative control and indicating that the observed effects depend on structured context rather than token identity alone.

Together, these findings support a bi-phasic depth-wise computational organization and point to a two-component geometric coding scheme in the prediction-centric phase: angular organization increasingly reflects token prediction distributions, while representation norms encode token- and context-specific information that does not directly influence predictive similarity.
This leads to a causal-geometric hypothesis: interventions that selectively modify angular geometry should be effective in the prediction-centric phase, but not in the context-centric phase. We test this hypothesis.

\paragraph{Angular interventions selectively affect the prediction-centric phase.}
\label{alternative_intervention_results}

Previous results indicated that angular organization in the second computational phase carries prediction-aligned information. This led to the hypothesis that angular interventions should be effective in the prediction-centric phase of LLM computation only, but not in the context-centric phase. We test this hypothesis using a purely angular intervention on the Months task.

Unlike the previous interventions, we apply a purely directional intervention. Prior to computing centroids, we normalize all representation vectors, so that centroid vectors become average direction vectors, thus avoiding leaking norm information. For each token representation we intervene on (whether input- or output-centric), we then normalize the token vector, replace its direction with that of the desired centroid direction, and finally rescale it to its original norm. This procedure alters only the representation’s direction while preserving its original magnitude.

\begin{figure}[t]
  \centering

  \begin{subfigure}[t]{0.31\textwidth}
    \centering
    \includegraphics[width=\linewidth]{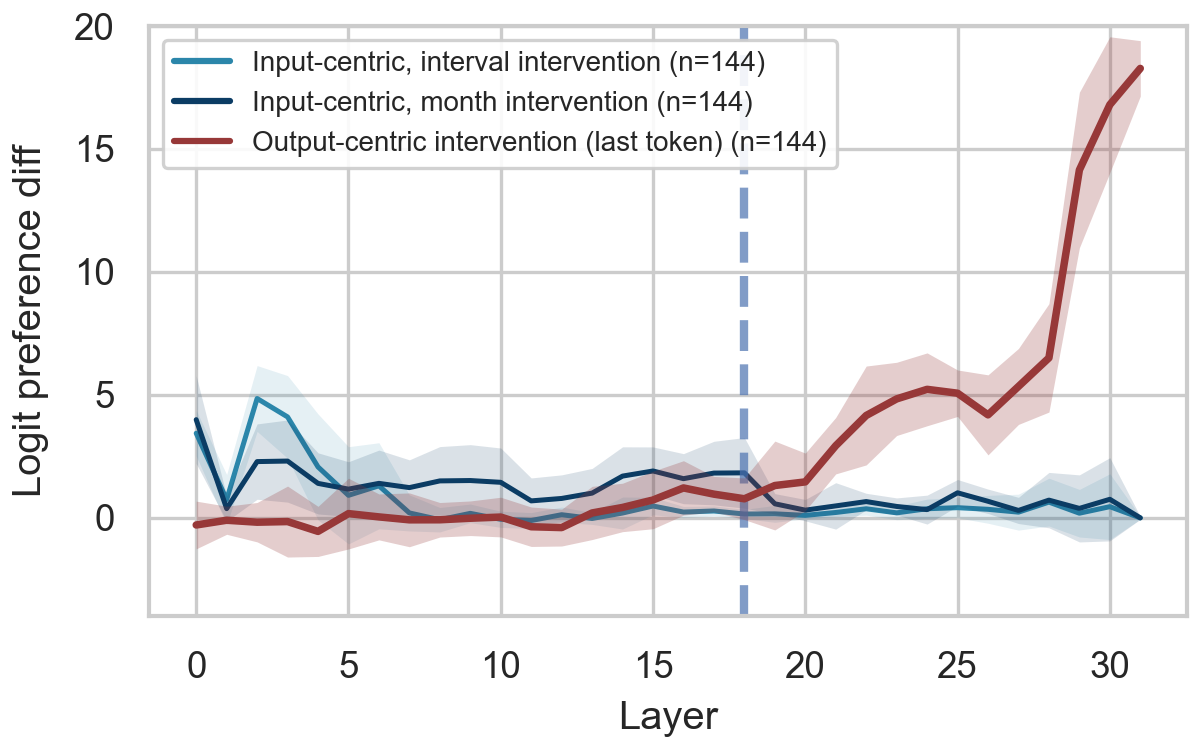}
    \caption{Llama (angle)}
    \label{fig:angle-llama}
  \end{subfigure}\hfill
  \begin{subfigure}[t]{0.31\textwidth}
    \centering
    \includegraphics[width=\linewidth]{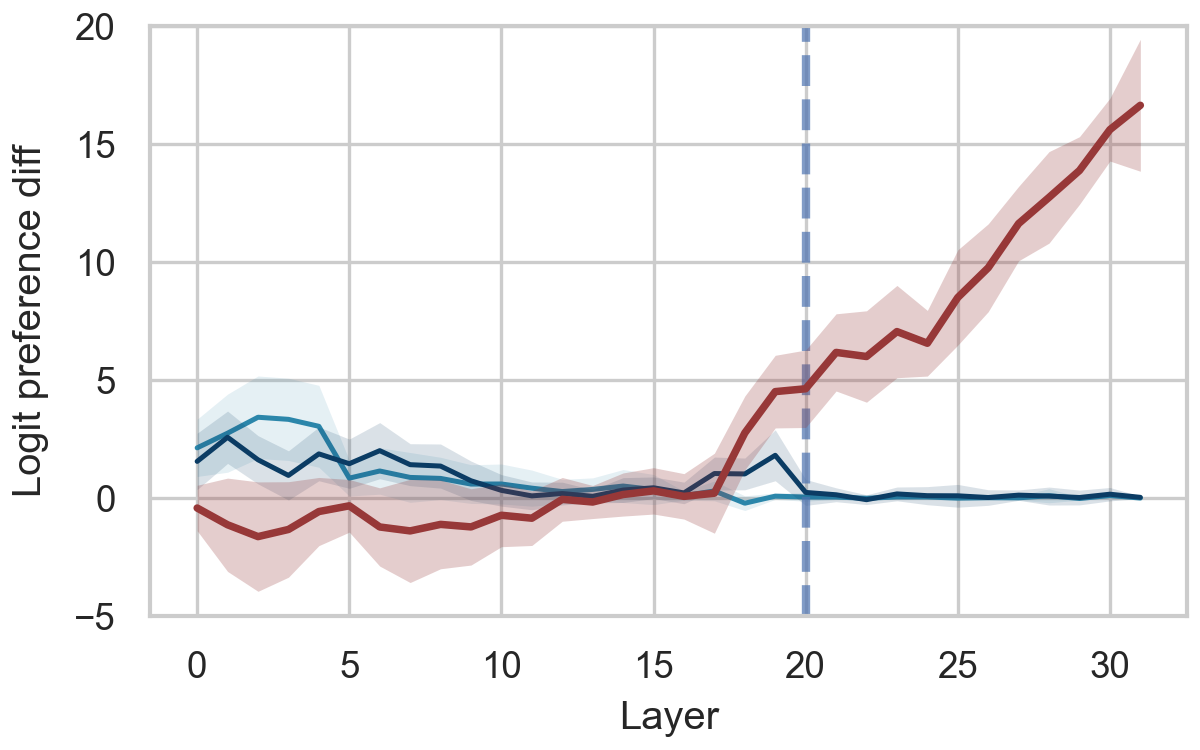}
    \caption{Mistral (angle)}
    \label{fig:angle-mistral}
  \end{subfigure}\hfill
  \begin{subfigure}[t]{0.31\textwidth}
    \centering
    \includegraphics[width=\linewidth]{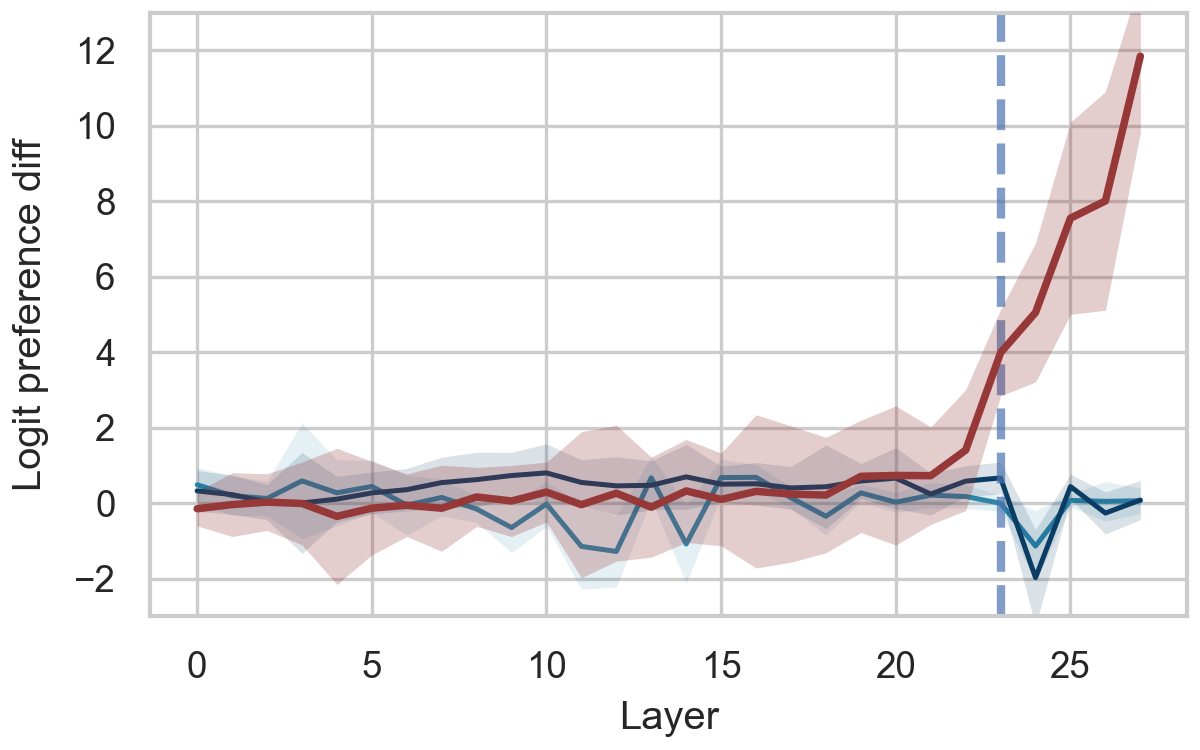}
    \caption{Qwen (angle)}
    \label{fig:angle-qwen}
  \end{subfigure}

  \begin{subfigure}[t]{0.31\textwidth}
    \centering
    \includegraphics[width=\linewidth]{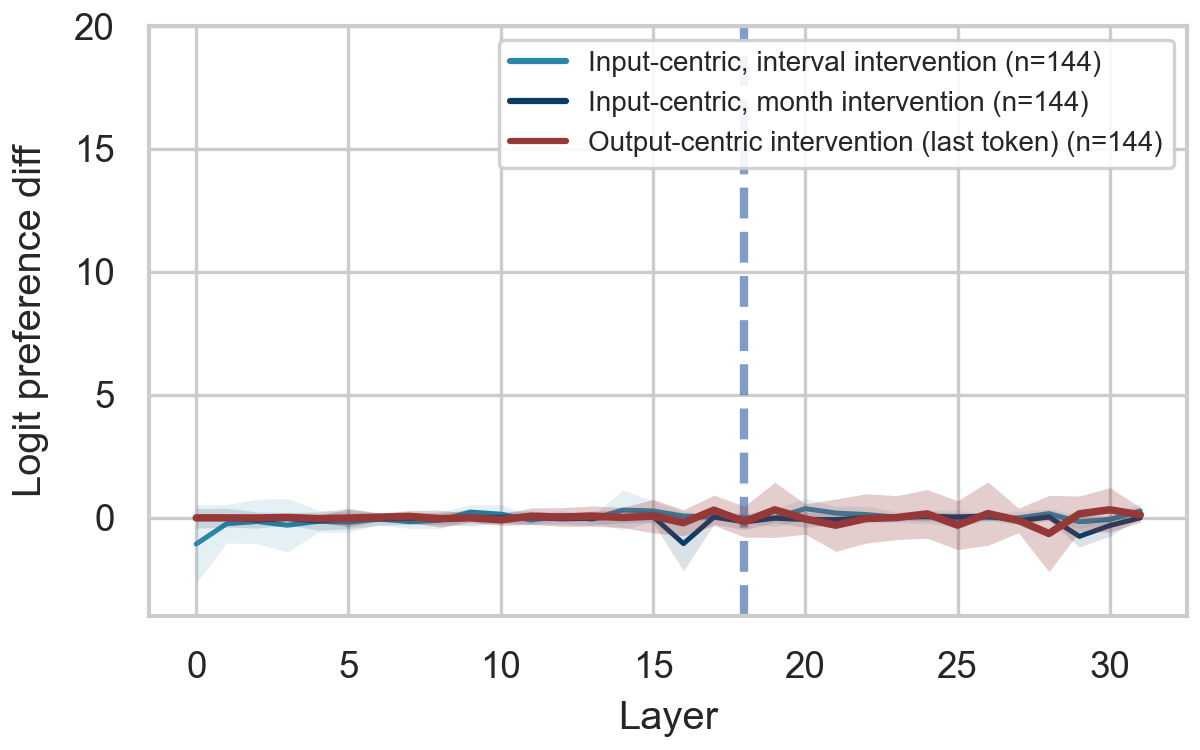}
    \caption{Llama (norm)}
    \label{fig:norm-llama}
  \end{subfigure}\hfill
  \begin{subfigure}[t]{0.31\textwidth}
    \centering
    \includegraphics[width=\linewidth]{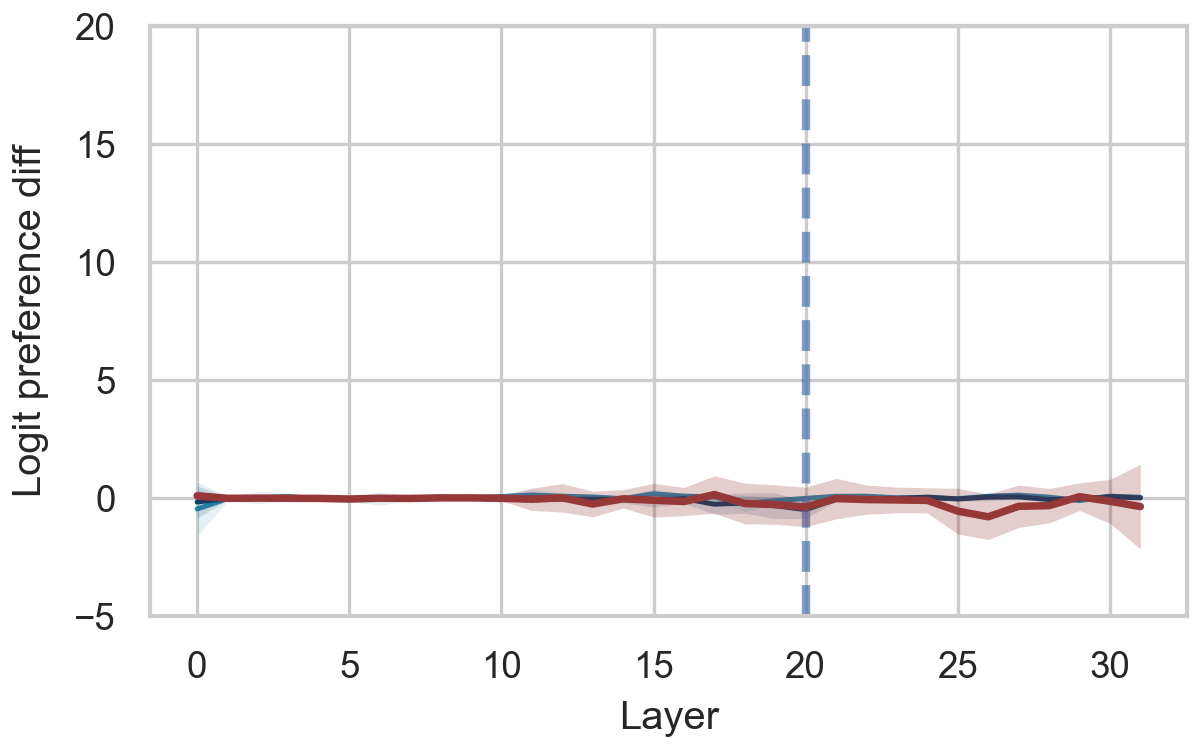}
    \caption{Mistral (norm)}
    \label{fig:norm-mistral}
  \end{subfigure}\hfill
  \begin{subfigure}[t]{0.31\textwidth}
    \centering
    \includegraphics[width=\linewidth]{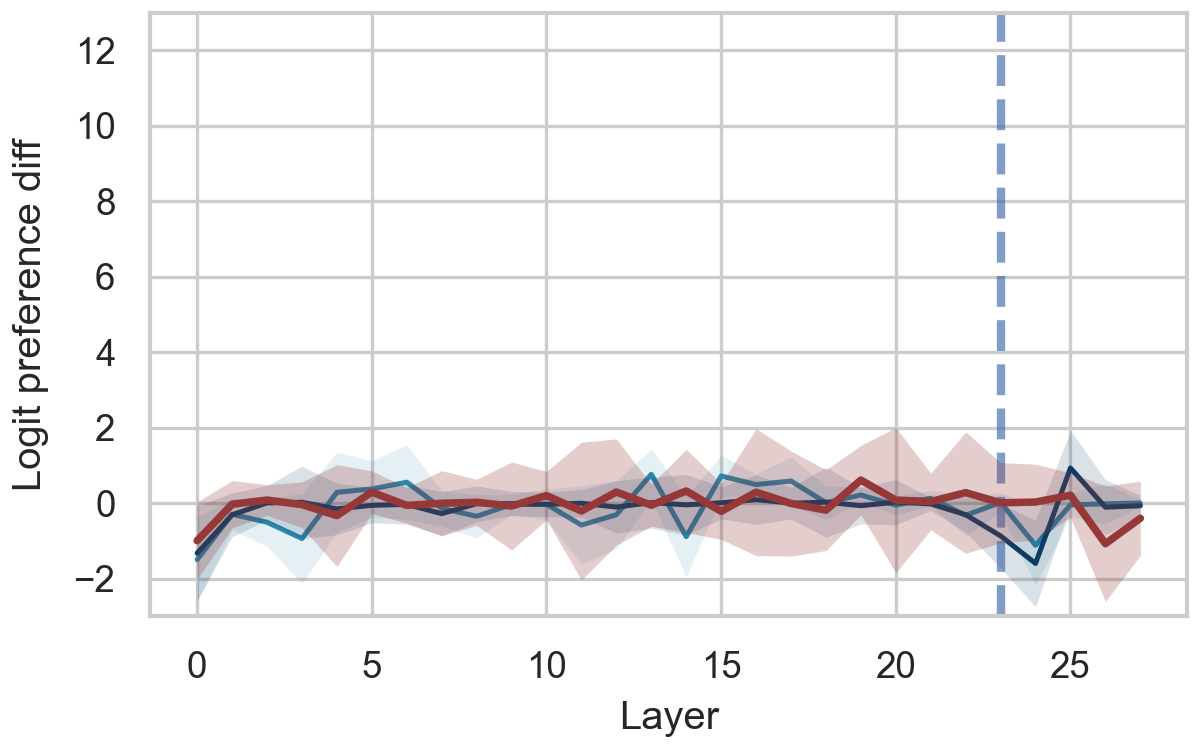}
    \caption{Qwen (norm)}
    \label{fig:norm-qwen}
  \end{subfigure}

  \caption{
    Per-layer intervention experiment results.
    \textbf{Top row}: pure angular interventions.
    \textbf{Bottom row}: pure norm interventions.
    Average logit preference difference is shown:
    blue curves correspond to input-centric interventions, interval and month;
    red curves correspond to output-centric interventions.
    The blue vertical line marks the original perturbation-based phase-change point. Shading marks ± Standard Deviation.
  }  
  \label{fig:angle_norm_interventions}
\end{figure}

In line with our predictions, input-centric angular interventions are largely ineffective across layers, whereas output-centric angular interventions remain potent in the second, prediction-centric phase (Fig.~\ref{fig:angle_norm_interventions}). To further test this dissociation, we additionally perform a purely norm-based intervention, in which token representations are rescaled to match the mean norm of the corresponding centroid vectors. This intervention is ineffective across layers and intervention types, apart from small fluctuations consistent with noise (Fig.~\ref{fig:angle_norm_interventions}). We replicate this qualitative pattern in two additional tasks, Days and Clock Math (Section~\ref{extra_angular_norm_int}; Figs.~\ref{fig:angle_norm_interventions_dow} and~\ref{fig:angle_norm_interventions_cm}).

One possible explanation for this dissociation follows directly from the structure of the final readout. Ignoring bias terms for simplicity, the output distribution for a token representation $h_L$ at the final layer can be written as $p(\text{output tokens}) = \operatorname{Softmax}(W_{\text{out}}h_L)$, where $W_{\text{out}}$ is the un-embedding matrix. Changing notation to separate norm and direction, we can write $p(\text{output tokens}) = \operatorname{Softmax}(W_{\text{out}}\lVert h_L\rVert_2\hat{h}_L)$. This decomposition makes explicit that the direction of $h_L$ determines the relative logit values across vocabulary entries, while the norm $\lVert h_L\rVert_2$ acts as a global scaling factor analogous to an inverse temperature, modulating the sharpness of the output distribution. This provides a suggestive mechanistic rationale for why selectively modifying angular structure is sufficient to alter prediction identity in the prediction-centric phase which can be further investigated in future work.

\section{Conclusion}
\vspace{-10pt}

A central contribution of this work is to clarify the relationship between two prominent but often disconnected approaches to analyzing large language models: geometric analyses of internal representational dynamics \citep{ethayarajh2019contexual,barbero_transformers_2024, cheng_emergence_2025,skean_layer_2025,sarfati_lines_2024, hosseini_straighten_2023,skean_layer_2025, cai_isotropy_2021,razzhigaev_shape_2024} and intervention-based analyses of mechanism \citealp{turner_steering_2023, engels_not_2025, yu_directions_2025,vu2026angular,Balest_model_geometry}. Our results provide a synthesis of these perspectives for the geometric quantities and tasks studied here, revealing how their relationship is structured by depth-wise computational dynamics. Geometric analyses recover structured organization in latent space, but our intervention results show that the presence of such structure is not, by itself, indicative of causal relevance for prediction. While all layers exhibit structured and decodable geometry, early and late layers vary in the type of interventions they are sensitive to. Most notably, selective causal control over prediction emerges primarily for late-layer angular organization. Thus, the geometric structure studied here is broadly present across layers, but only becomes mechanistically operative for prediction steering in later layers, where it is embedded within the network’s global dynamical structure and directly shapes output.

An open question raised by our findings concerns the architectural and optimization biases that may favor angular, rather than norm-based, coding in prediction-forming layers. Modern transformer architectures make extensive use of normalization schemes such as RMSNorm, which explicitly normalize or constrain representation magnitudes while preserving directional information \citep{zhang2019rmsnorm}. While normalization alone does not imply angular coding, it plausibly creates favorable conditions under which direction becomes a more reliable and functionally meaningful signal than magnitude. More generally, understanding how network architecture design principles, such as normalization, self-attention, and residual connections interact to promote angularly organized prediction geometry remains an important direction for future work \citep{he2016deep, vaswani2017attention, zhang2019rmsnorm}.

Our results also connect naturally to the widespread use of angular distance (e.g., cosine similarity) as a measure of semantic similarity \citep{mikolov-etal-2013-linguistic} and as a characteristic feature of representational geometry \citep{hosseini_straighten_2023}. Angular similarity has long been observed to correlate with semantic relatedness in word embeddings and contextual representations ~\citep{diegosimon2024polarcoordinaterepresentssyntax}. Recently the functional role of this geometry in decoder-based LLMs has been exemplified, yet its depth-dependent performance remained puzzling ~\citep{vu2026angular}. Our findings provide a mechanistic grounding for this practice in late layers of decoder-only LLMs: angular proximity does not merely reflect semantic similarity but parametrizes similarity between output prediction distributions. 


These distinctions have direct implications for the understanding and control of large language models \citep{gilpin2018explaining_explanations, pan2025trainingfree_alignment}. Much prior work implicitly assumes that decodable or separable structure, whether semantic, behavioral, or safety-related \citep{saglam_large_2025}, constitutes a suitable target for intervention. Our results highlight a limitation of this assumption: decodable representational structure may be geometrically redundant with respect to the mechanisms that actually govern prediction. As a result, interventions that target easily decodable features may fail to generalize or may exert only indirect control \citep{ravichander-etal-2021-probing, hewitt-liang-2019-designing, elazar-etal-2020-amnesic-probing}. By contrast, interventions aligned with causally meaningful geometry, such as late-layer angular structure, offer a more principled basis for reliable control e.g. geometry-aware, layer-specific adapters \citep{hu2021lora}. Furthermore, our results suggest that evaluating intervention strategies should involve not only detectability, but the characterization of the dynamic mechanistically-relevant geometry.



\vspace{-5pt}
\paragraph{Limitations and Future Work.} This work has several limitations. First, our investigation focuses on a single class of modular tasks, a specific large natural text dataset, and three model families, and it remains unclear how broadly the identified causal-geometric schema generalizes across tasks, modalities, or training regimes. Second, while we identify a sharp transition from context-processing to prediction-forming computation, the relationship between early-layer context geometry (e.g. dimensionality expansion) and the emergence of prediction-aligned angular codes is not yet understood. One possibility is that early layers prepare a high-dimensional contextual basis from which late layers select and stabilize angularly organized predictive directions, but this hypothesis remains speculative. Finally, our analysis emphasizes representations averaged across contexts and does not address the context-dependent within-sequence dynamics.


\vspace{-5pt}
\paragraph{Broader Impacts.} The work is foundational in nature and does not propose new deployment, safety, or policy interventions; its primary impact is to improve scientific understanding of how LLMs compute predictions. Our findings may inform future approaches to model interpretation, fine-tuning, and control of LLMs.

\bibliographystyle{unsrtnat}
\bibliography{references}

\newpage
\appendix

\section*{Appendix}
\addcontentsline{toc}{section}{Appendix}

\renewcommand{\thefigure}{\thesection.\arabic{figure}}
\setcounter{figure}{0}
\FloatBarrier
\section{Additional intervention results}
\label{appendix_interventions}
\subsection{Experimental setup}
\paragraph{Days of the week task.}
We replicate our intervention results on the Days of the week task \citealp{engels_not_2025}: "\textit{Let's do some days of the week math. [\textit{INTERVAL}] days from [\textit{DAY}] is}" , where \textit{INTERVAL} takes values from \{One, Two,..., Seven\}, and \textit{DAY} can be any day of the week. In total, this task has 49 prompts that correspond to every combination of \textit{INTERVAL} and \textit{DAY}. Here, we only readout the logits corresponding to days of the week. In this task the models accuracies are: Llama 86\%, Mistral 59\% and Qwen 77\% .

\paragraph{Clock math task.}
In addition to the two previous tasks we also replicate our intervention results on a simple clock arithmetic task - Clock math: "\textit{Let's do some clock math. [\textit{INTERVAL}] hours from [\textit{HOUR}] o'clock is}" , where \textit{INTERVAL} takes values from \{One, Two,..., Twelve\}, and similarly \textit{HOUR} takes values from \{one, two,..., twelve\}. In total, this task has 144 prompts that correspond to every combination of \textit{INTERVAL} and \textit{HOUR}. Here, we only readout the logits corresponding to the hour strings. In this task the models accuracies are: Llama 95\%, Mistral 80\% and Qwen 76\% .

\begin{figure}[t]
  \centering
  \includegraphics[width=\textwidth]{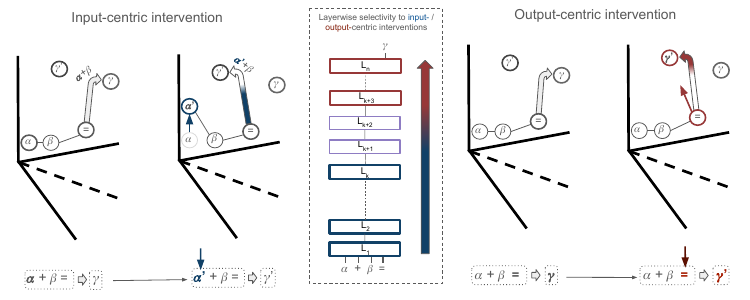}
  \caption{
    Visual illustration of two intervention approaches applied to a modular task. A schematic of an input sequence is presented in latent space, first, with its natural prediction output, and then with the intervention applied. This is visualized both for input- (left) and output-centric (right) interventions. Inset: Schematic of layer-wise efficacy for both approach.
  }
  \label{fig:intervention-visual}
\end{figure}

\subsection{Additional Intervention results}

\begin{figure}
  \centering

  \begin{subfigure}[t]{0.31\textwidth}
    \centering
    \includegraphics[width=\linewidth]{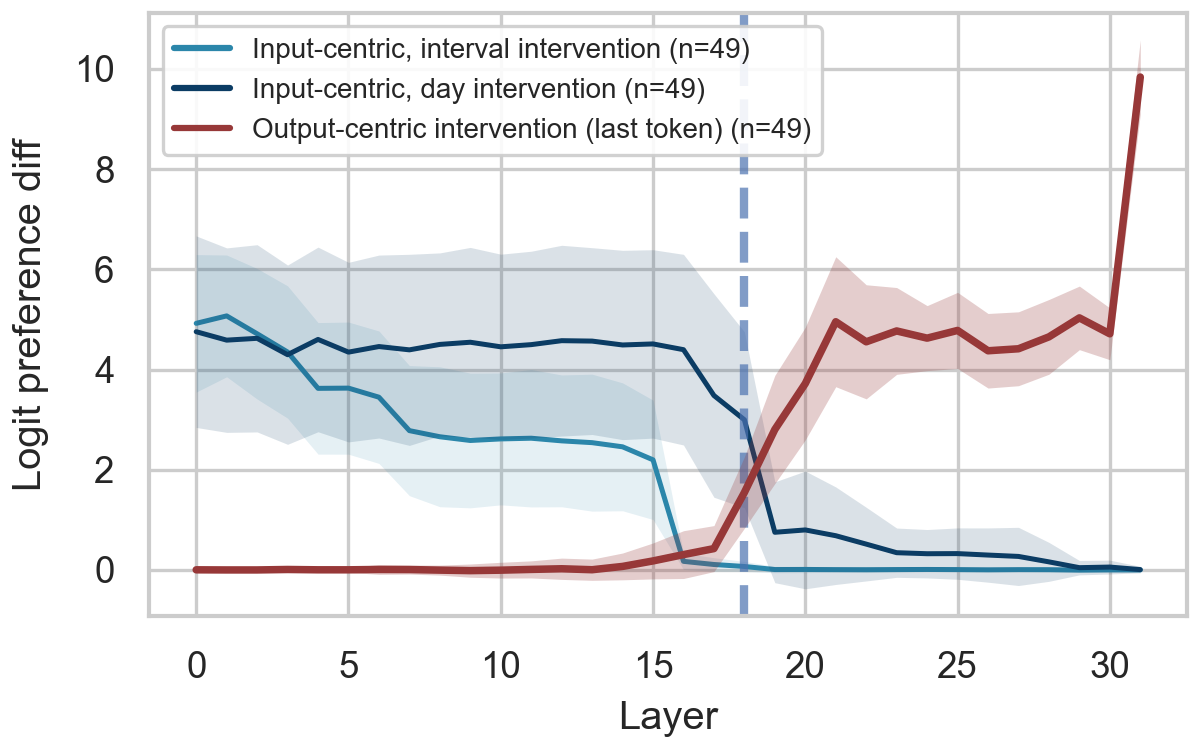}
    \caption{Llama - DOW}
    \label{fig:dow-int-llama}
  \end{subfigure}\hfill
  \begin{subfigure}[t]{0.31\textwidth}
    \centering
    \includegraphics[width=\linewidth]{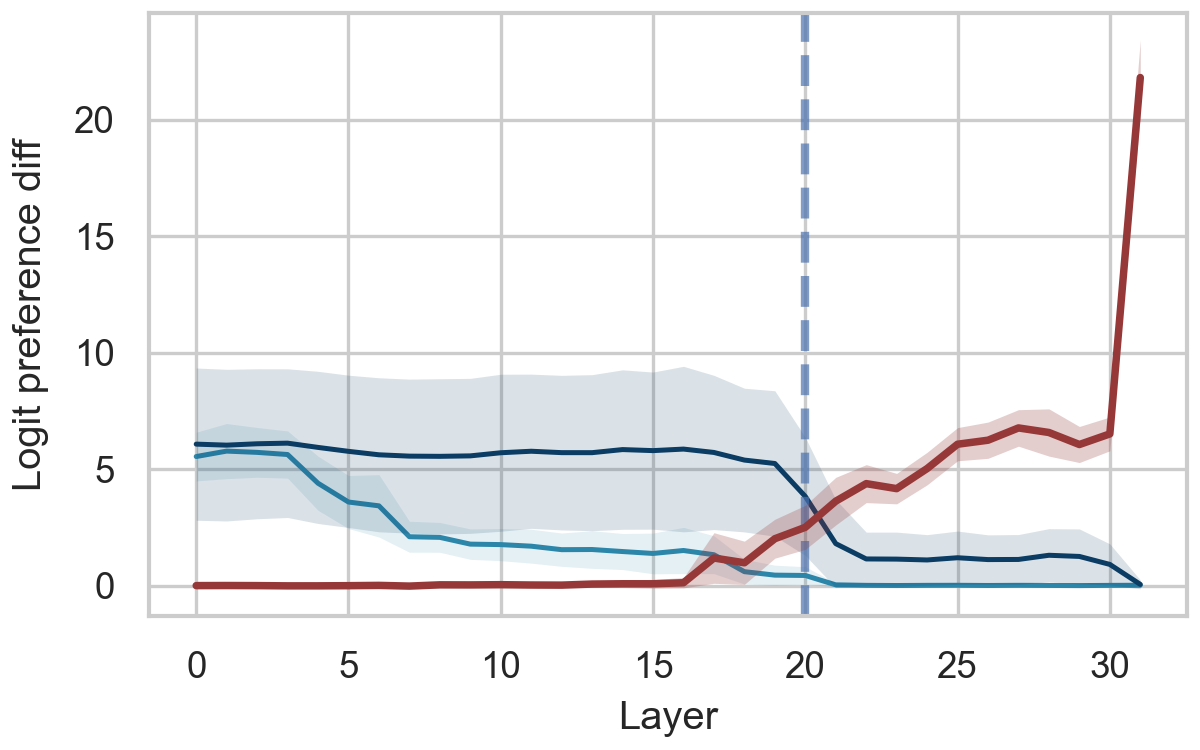}
    \caption{Mistral- DOW}
    \label{fig:dow-int-mistral}
  \end{subfigure}\hfill
  \begin{subfigure}[t]{0.31\textwidth}
    \centering
    \includegraphics[width=\linewidth]{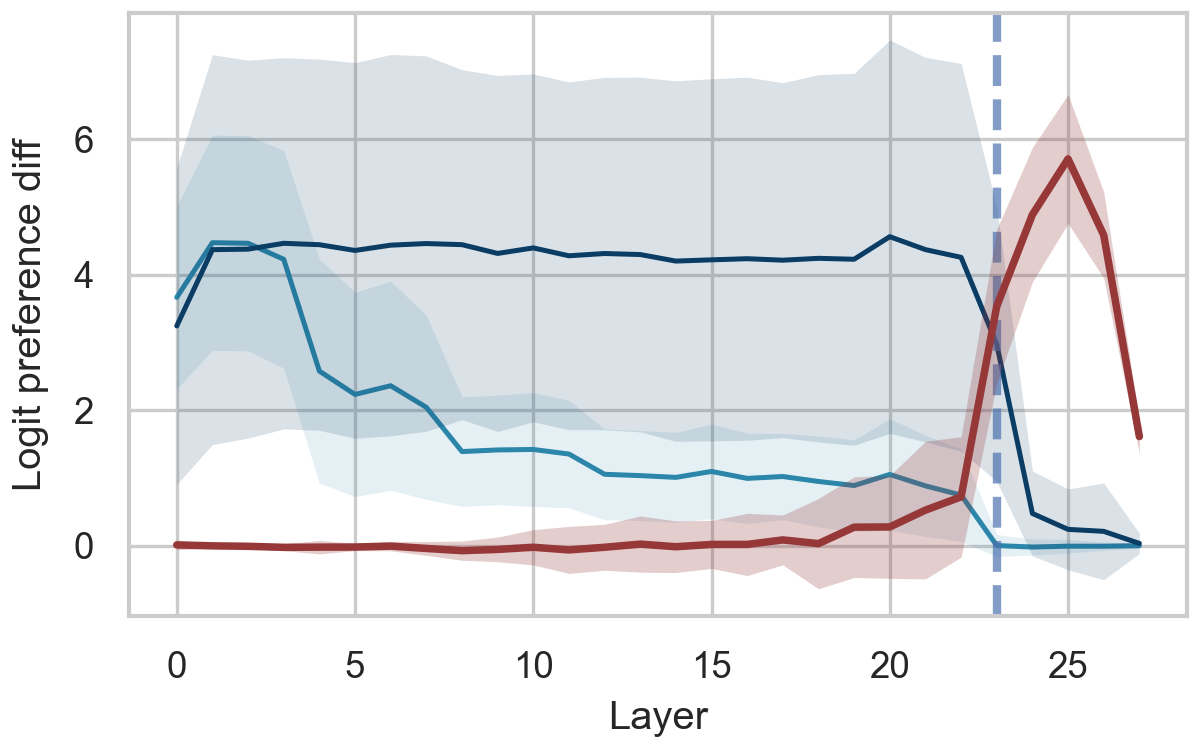}
    \caption{Qwen - DOW}
    \label{fig:dow-int-qwen}
  \end{subfigure}

  \vspace{0.6em}

  \begin{subfigure}[t]{0.31\textwidth}
    \centering
    \includegraphics[width=\linewidth]{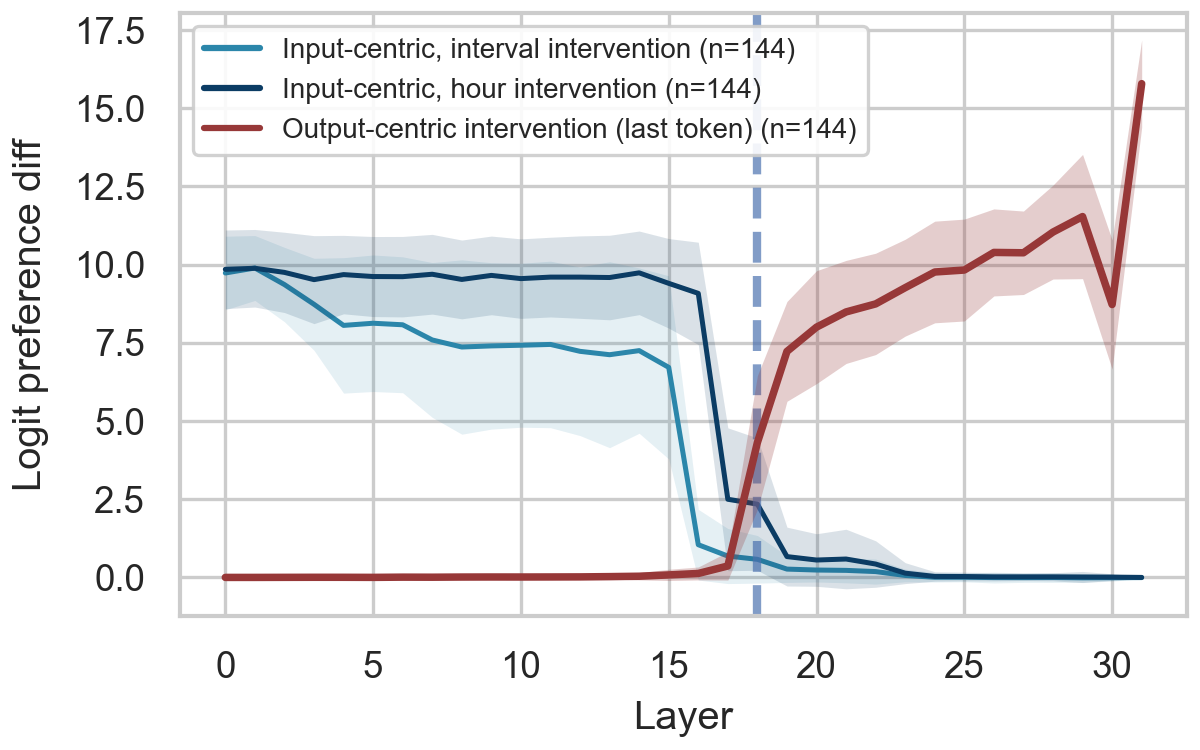}
    \caption{Llama - CM}
    \label{fig:cm-int-llama}
  \end{subfigure}\hfill
  \begin{subfigure}[t]{0.31\textwidth}
    \centering
    \includegraphics[width=\linewidth]{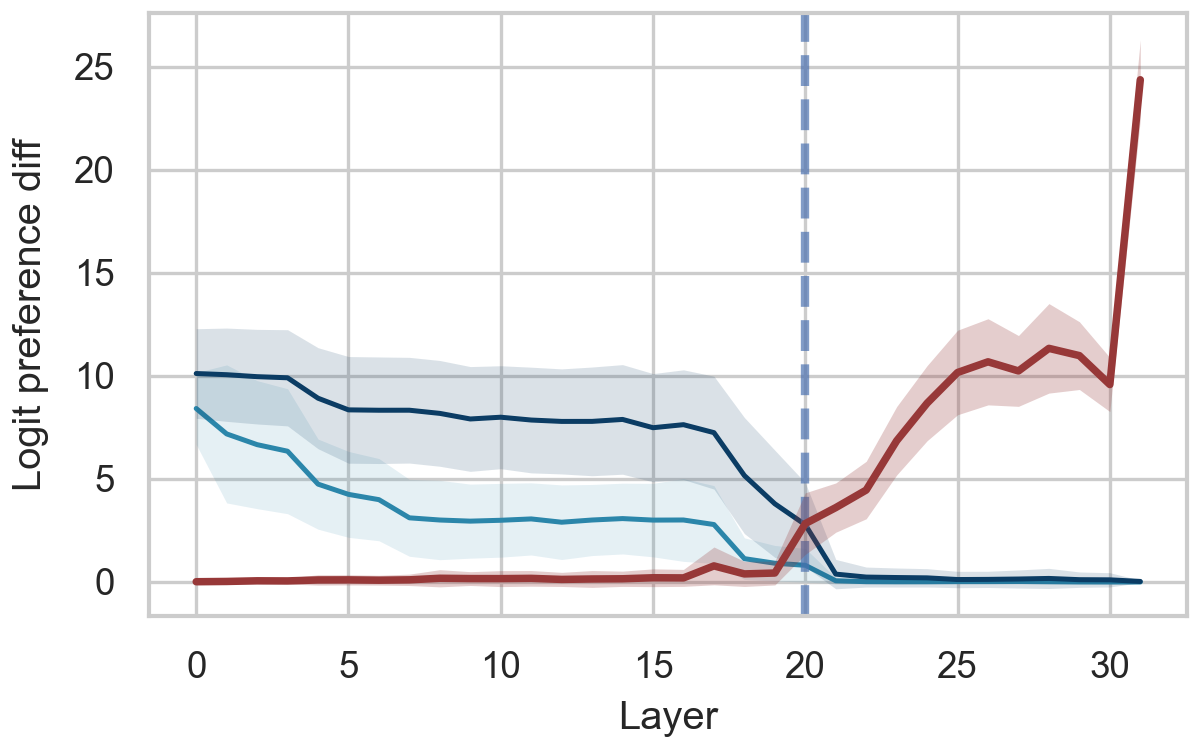}
    \caption{Mistral- CM}
    \label{fig:cm-int-mistral}
  \end{subfigure}\hfill
  \begin{subfigure}[t]{0.31\textwidth}
    \centering
    \includegraphics[width=\linewidth]{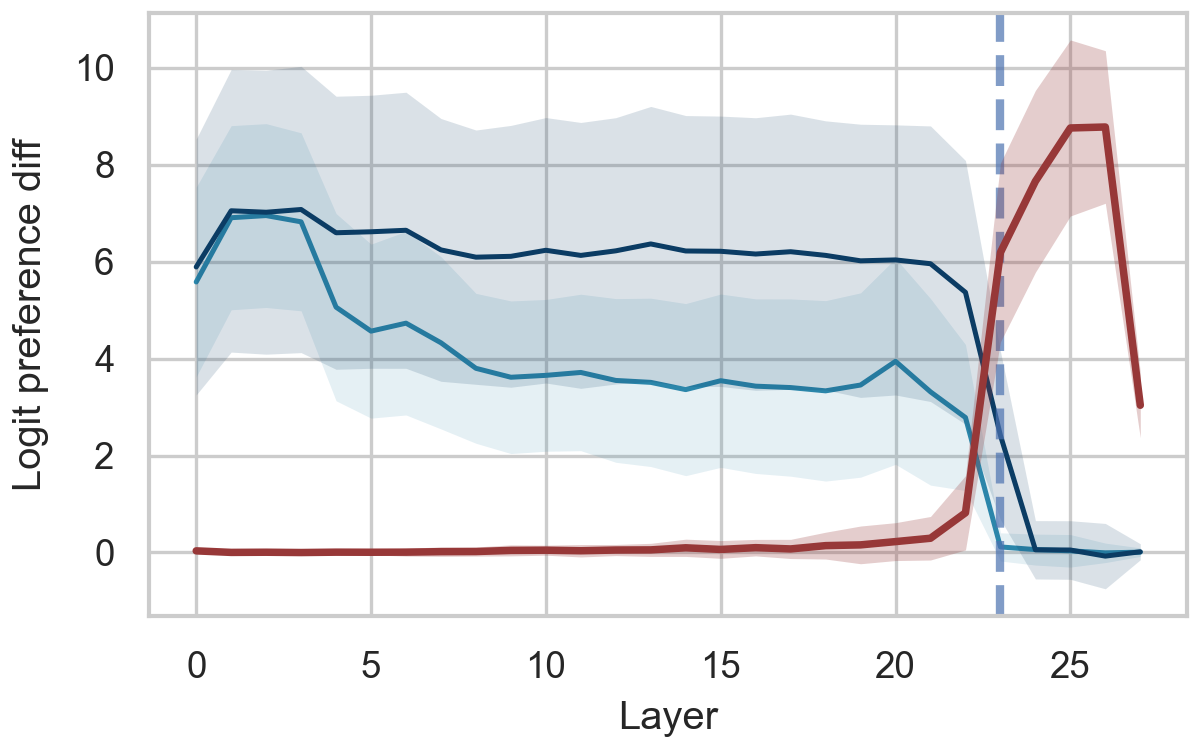}
    \caption{Qwen - CM}
    \label{cm-fig:cm-int-qwen}
  \end{subfigure}  

  \caption{Per-layer intervention experiment results on Days Of the Week (DOW; top), and Clock Math (CM; bottom) tasks. Average logit preference difference: blue curves show input-based interventions, interval and month; the red curve shows output-based intervention. The blue vertical line marks the phase-change point.}
  \label{fig:add-intervention-extras}
\end{figure}

\paragraph{Euclidean intervention results. }
We applied the intervention techniques introduced in Section~\ref{Interventions} to the Days and Clock Math tasks, and observed the same qualitative bi-phasic pattern found for Months (Fig.~\ref{fig:add-intervention-extras}; compare Fig.~\ref{fig:intervention}). In all three tasks, input-centric interventions have larger effects in earlier layers, while output-centric interventions have larger effects in later layers.

We quantified the transition between these regimes by defining a change point as the intermediate layer that minimizes the absolute difference between the output-centric and input-centric (interval) intervention curves. This change point was consistent across Months, Days, and Clock Math. The same conclusion was obtained using a lightly smoothed zero-crossing criterion, for which the estimated change points agreed within 2.5\% of model depth across tasks.

\begin{figure}[t]
  \centering

  \begin{subfigure}[t]{0.31\textwidth}
    \centering
    \includegraphics[width=\linewidth]{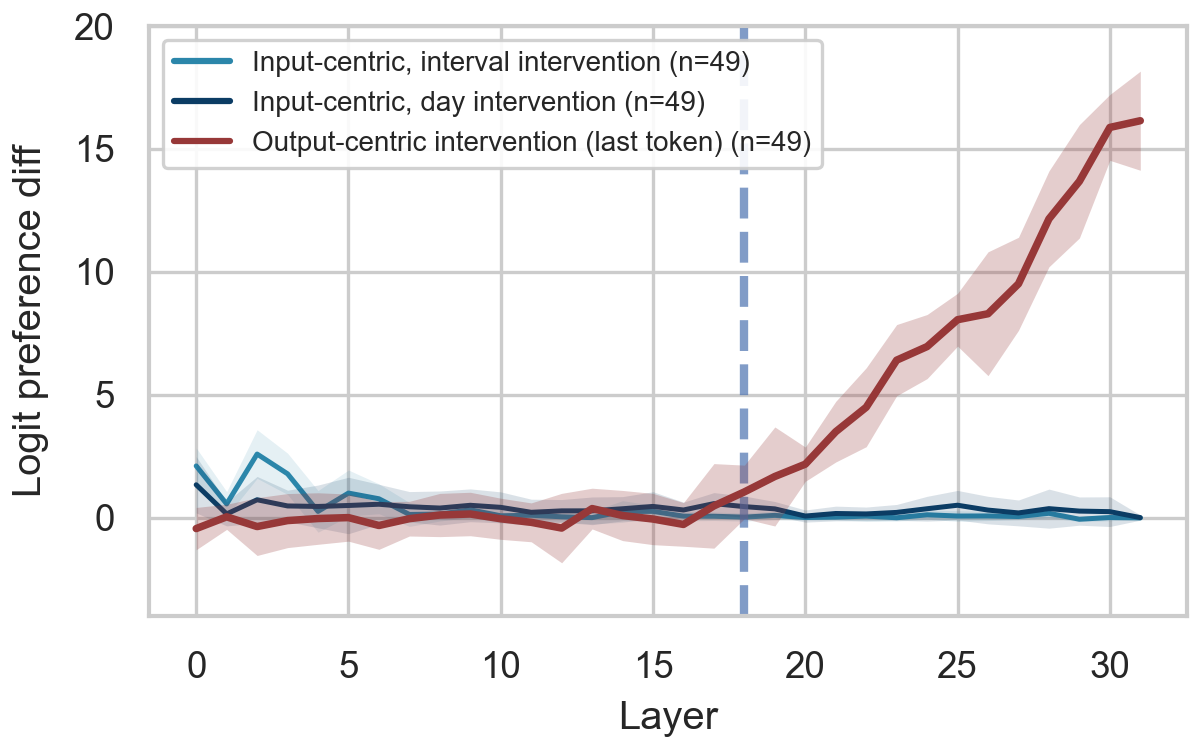}
    \caption{Llama (angle)}
    \label{fig:dow-angle-llama}
  \end{subfigure}\hfill
  \begin{subfigure}[t]{0.31\textwidth}
    \centering
    \includegraphics[width=\linewidth]{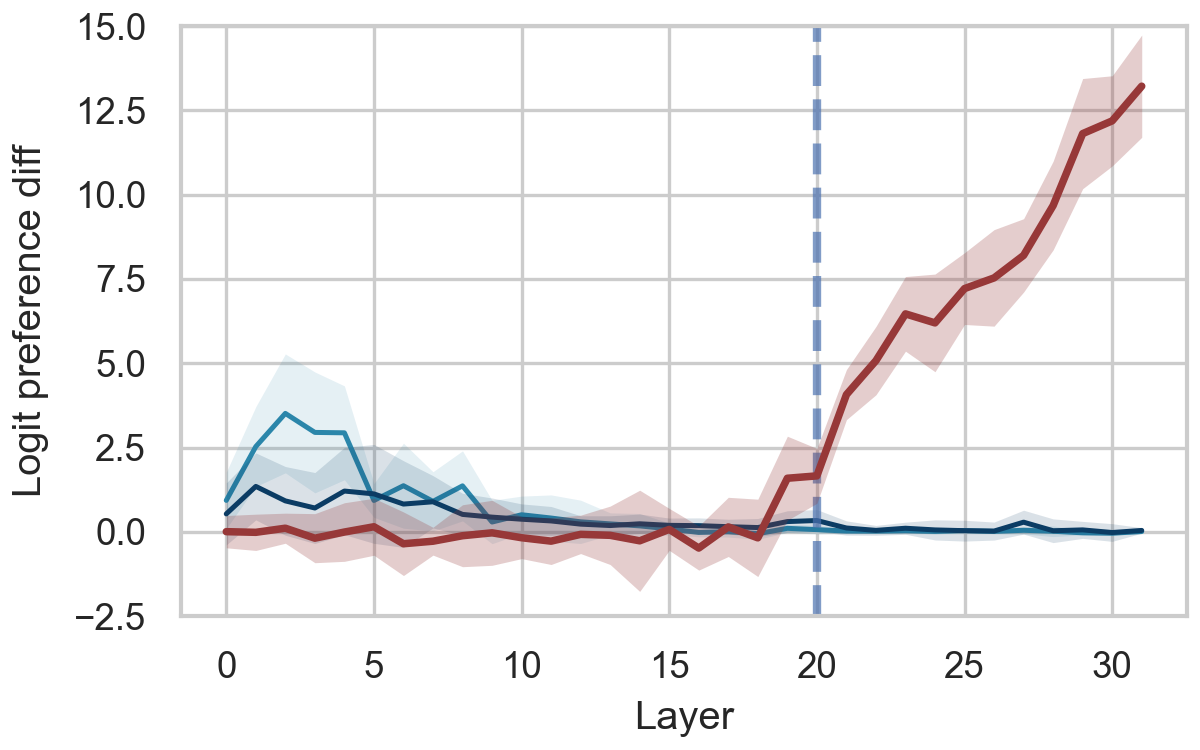}
    \caption{Mistral (angle)}
    \label{fig:dow-angle-mistral}
  \end{subfigure}\hfill
  \begin{subfigure}[t]{0.31\textwidth}
    \centering
    \includegraphics[width=\linewidth]{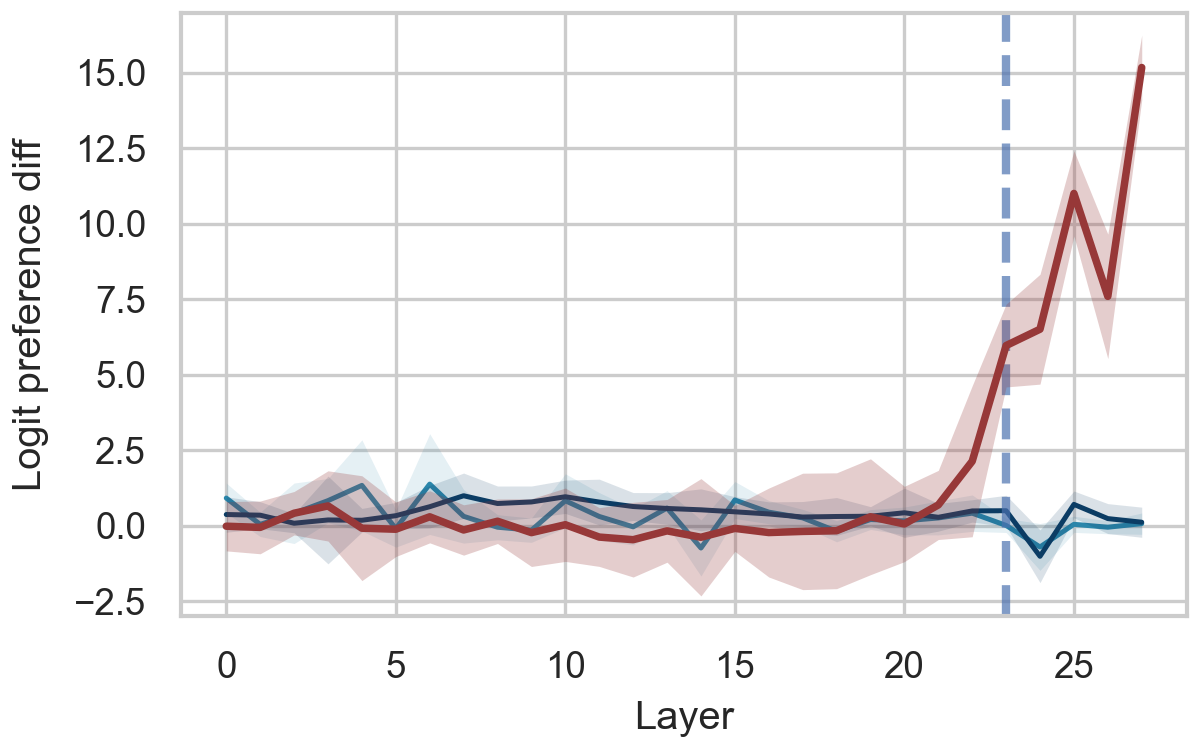}
    \caption{Qwen (angle)}
    \label{fig:dow-angle-qwen}
  \end{subfigure}

  \vspace{0.6em}

  \begin{subfigure}[t]{0.31\textwidth}
    \centering
    \includegraphics[width=\linewidth]{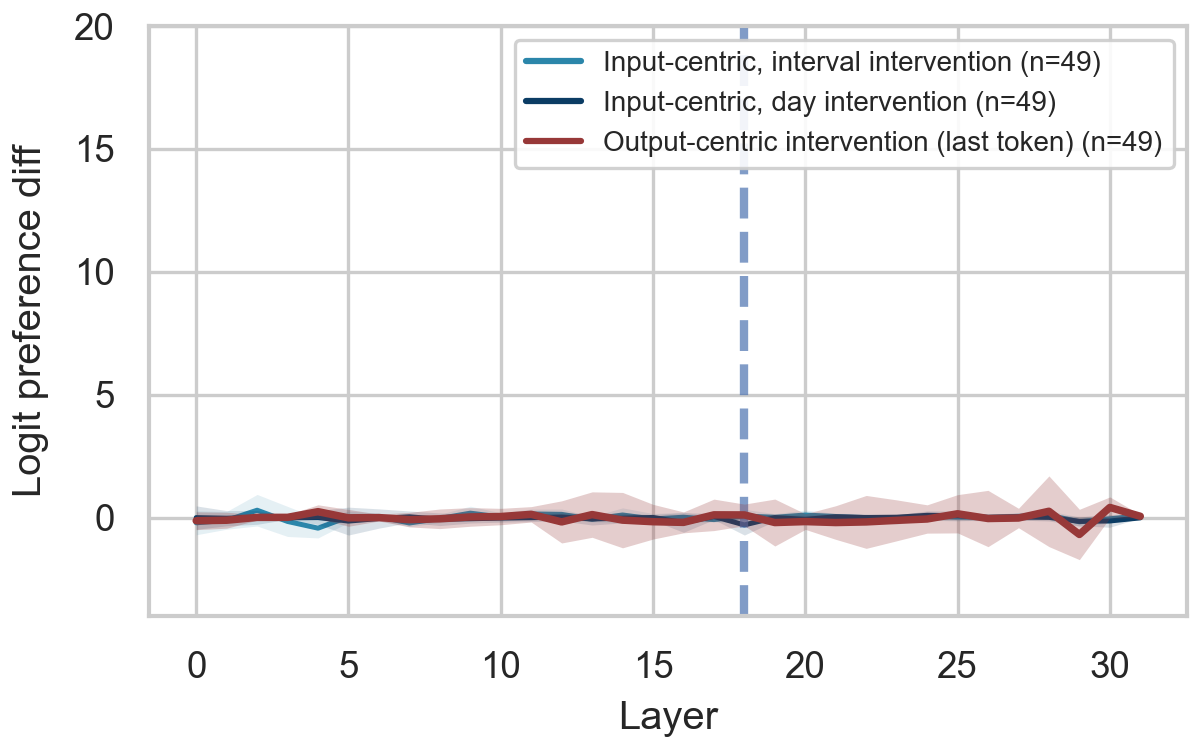}
    \caption{Llama (norm)}
    \label{fig:dow-norm-llama}
  \end{subfigure}\hfill
  \begin{subfigure}[t]{0.31\textwidth}
    \centering
    \includegraphics[width=\linewidth]{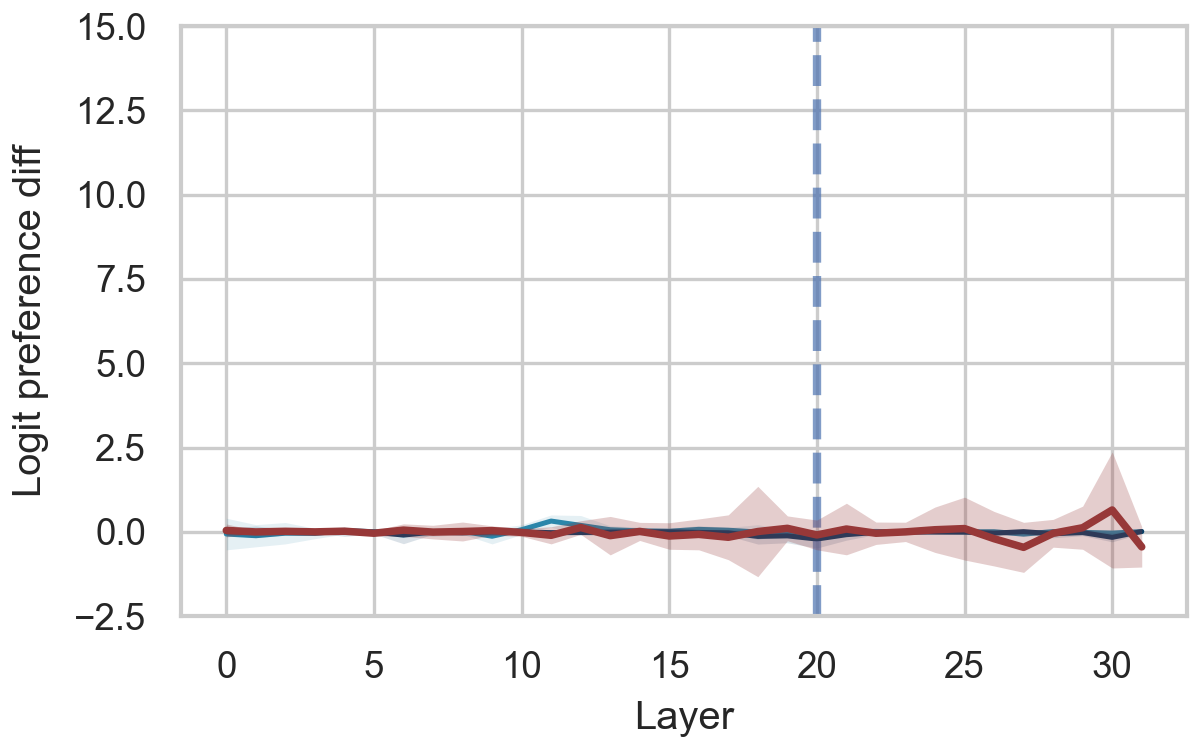}
    \caption{Mistral (norm)}
    \label{fig:dow-norm-mistral}
  \end{subfigure}\hfill
  \begin{subfigure}[t]{0.31\textwidth}
    \centering
    \includegraphics[width=\linewidth]{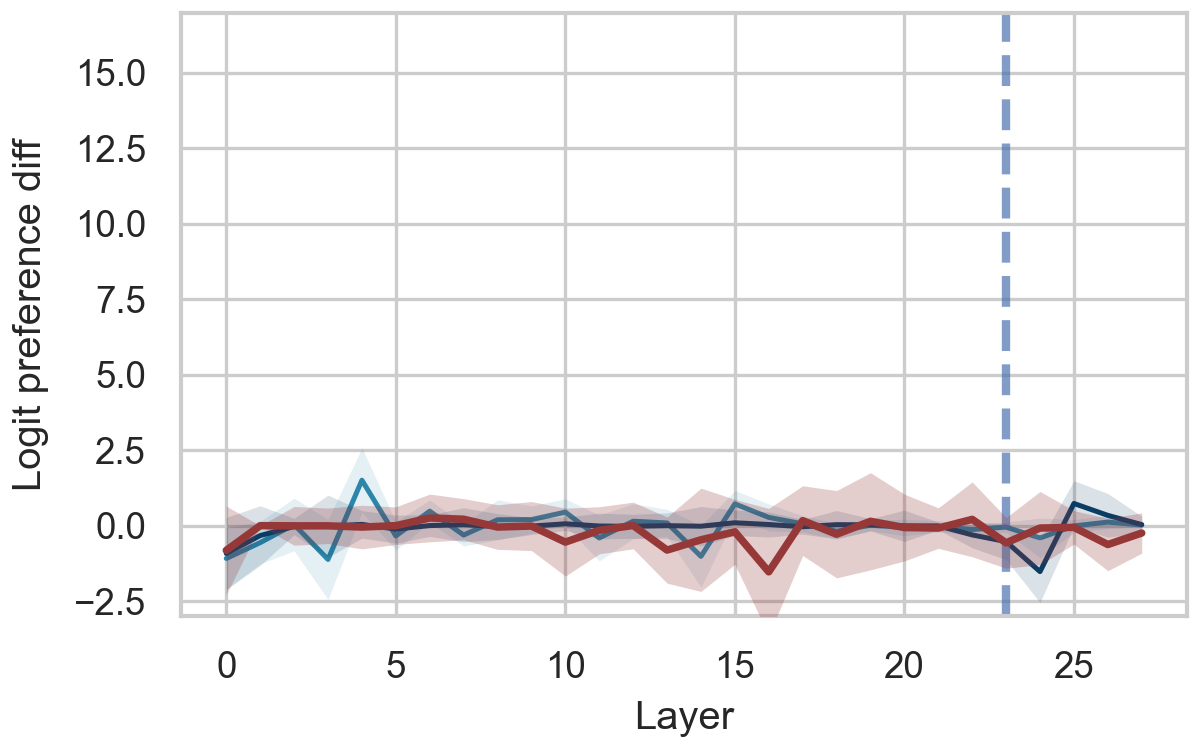}
    \caption{Qwen (norm)}
    \label{fig:dow-norm-qwen}
  \end{subfigure}

  \caption{
    Per-layer intervention experiment results on Days of the week task.
    \textbf{Top row}: pure angular interventions.
    \textbf{Bottom row}: pure norm interventions.
    Average logit preference difference is shown:
    blue curves correspond to input-based interventions, interval and month;
    red curves correspond to output-based interventions.
    The blue vertical line marks the original perturbation-based phase-change point.
  }
  
  \label{fig:angle_norm_interventions_dow}
\end{figure}

\begin{figure}[t]
  \centering

  \begin{subfigure}[t]{0.31\textwidth}
    \centering
    \includegraphics[width=\linewidth]{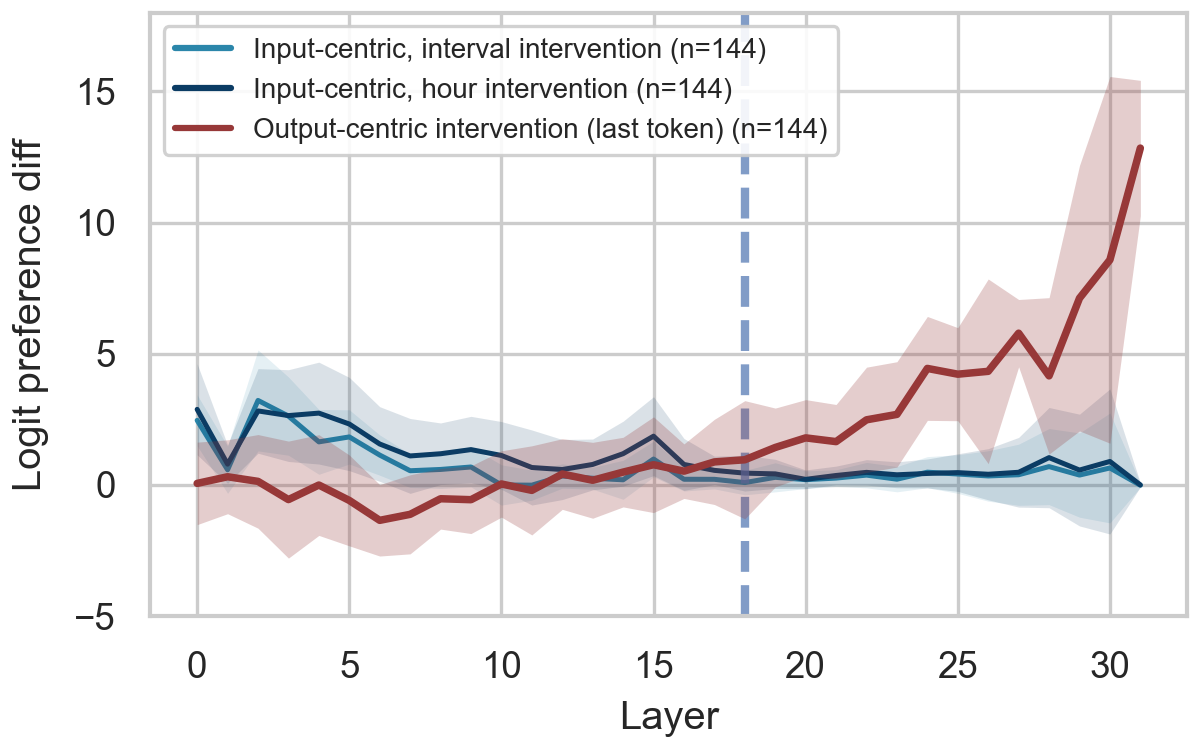}
    \caption{Llama (angle)}
    \label{fig:cm-angle-llama}
  \end{subfigure}\hfill
  \begin{subfigure}[t]{0.31\textwidth}
    \centering
    \includegraphics[width=\linewidth]{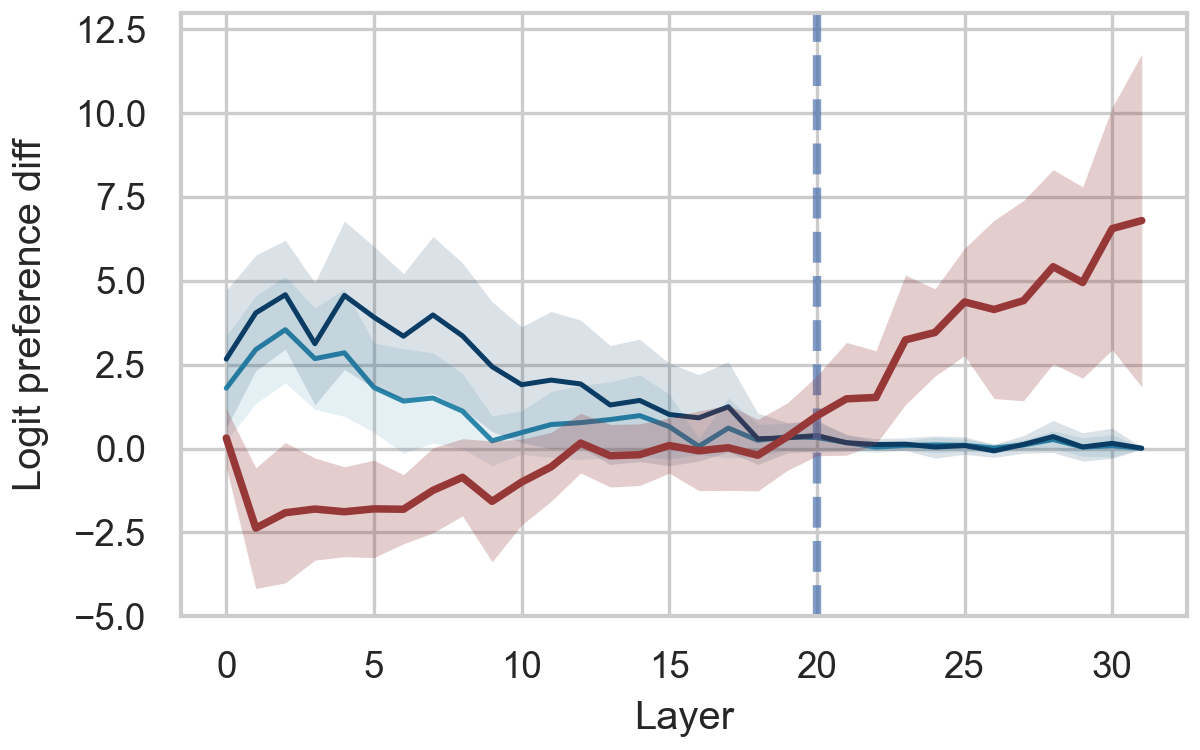}
    \caption{Mistral (angle)}
    \label{fig:cm-angle-mistral}
  \end{subfigure}\hfill
  \begin{subfigure}[t]{0.31\textwidth}
    \centering
    \includegraphics[width=\linewidth]{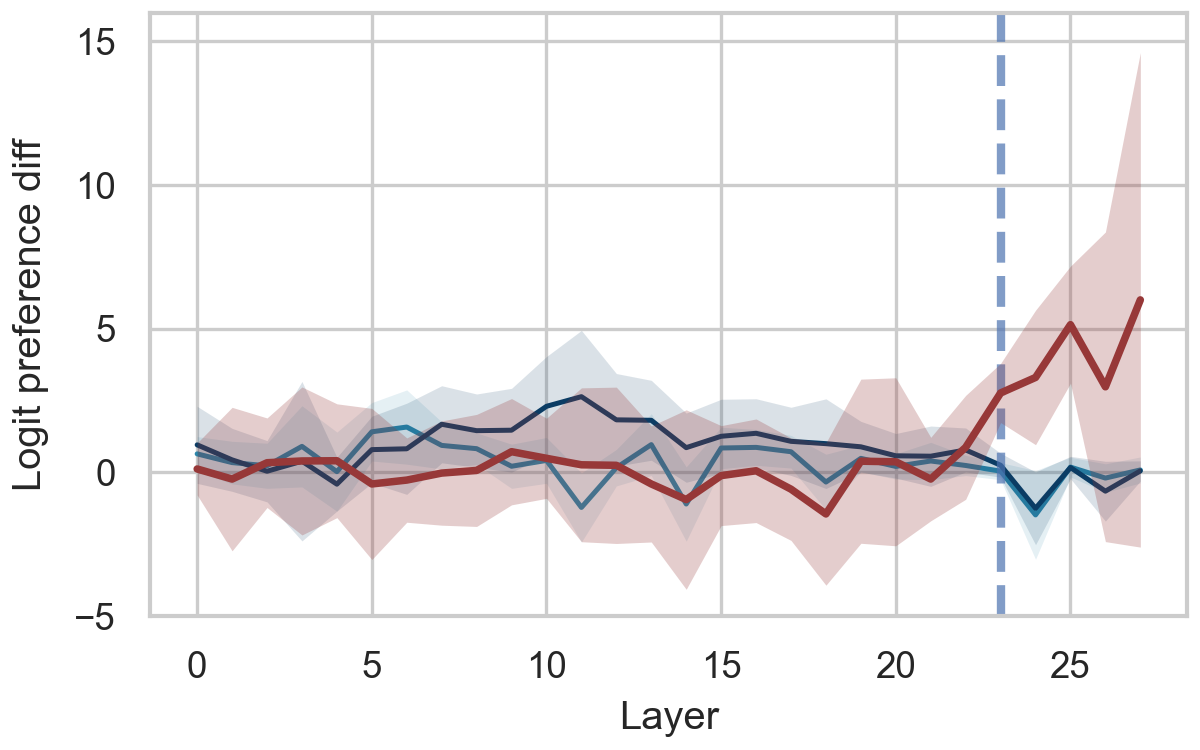}
    \caption{Qwen (angle)}
    \label{fig:cm-angle-qwen}
  \end{subfigure}

  \vspace{0.6em}

  \begin{subfigure}[t]{0.31\textwidth}
    \centering
    \includegraphics[width=\linewidth]{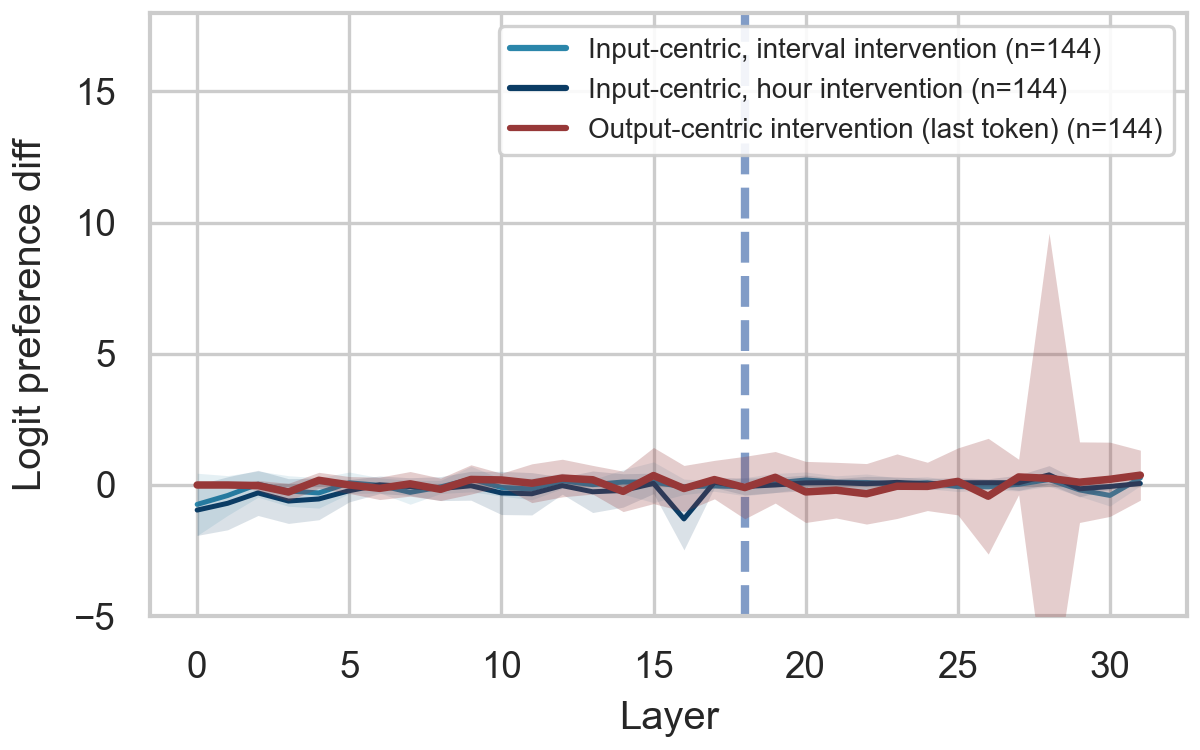}
    \caption{Llama (norm)}
    \label{fig:cm-norm-llama}
  \end{subfigure}\hfill
  \begin{subfigure}[t]{0.31\textwidth}
    \centering
    \includegraphics[width=\linewidth]{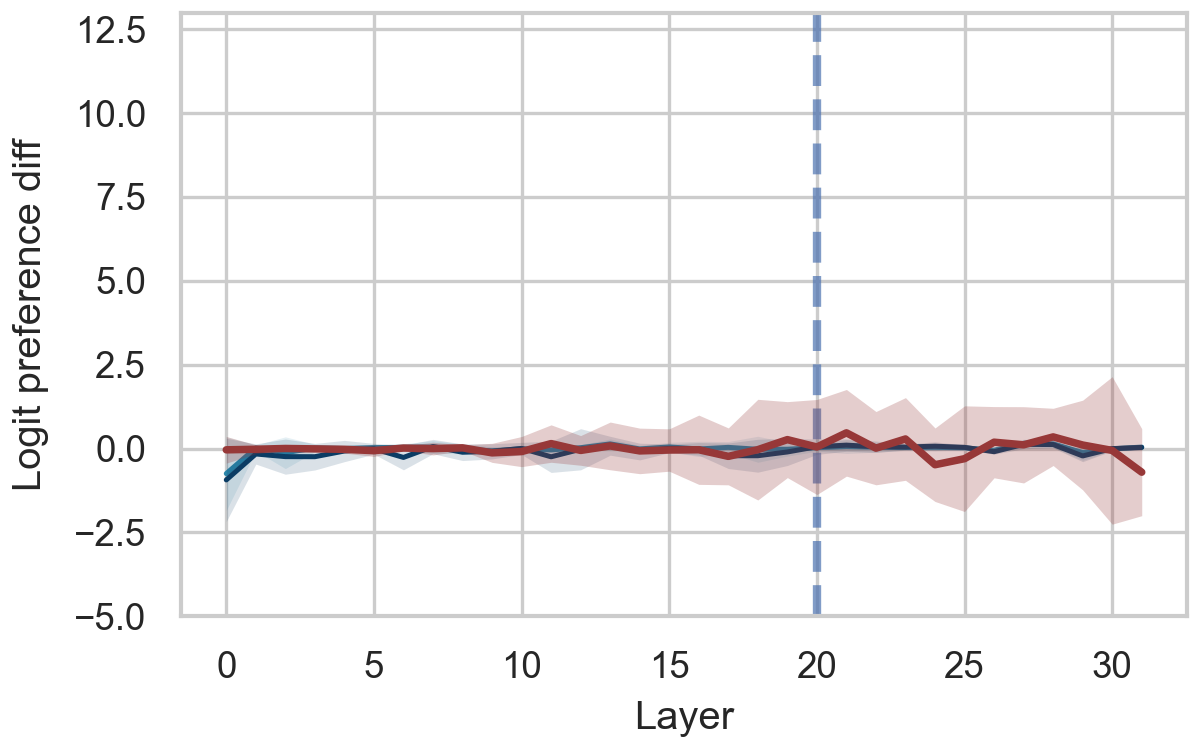}
    \caption{Mistral (norm)}
    \label{fig:cm-norm-mistral}
  \end{subfigure}\hfill
  \begin{subfigure}[t]{0.31\textwidth}
    \centering
    \includegraphics[width=\linewidth]{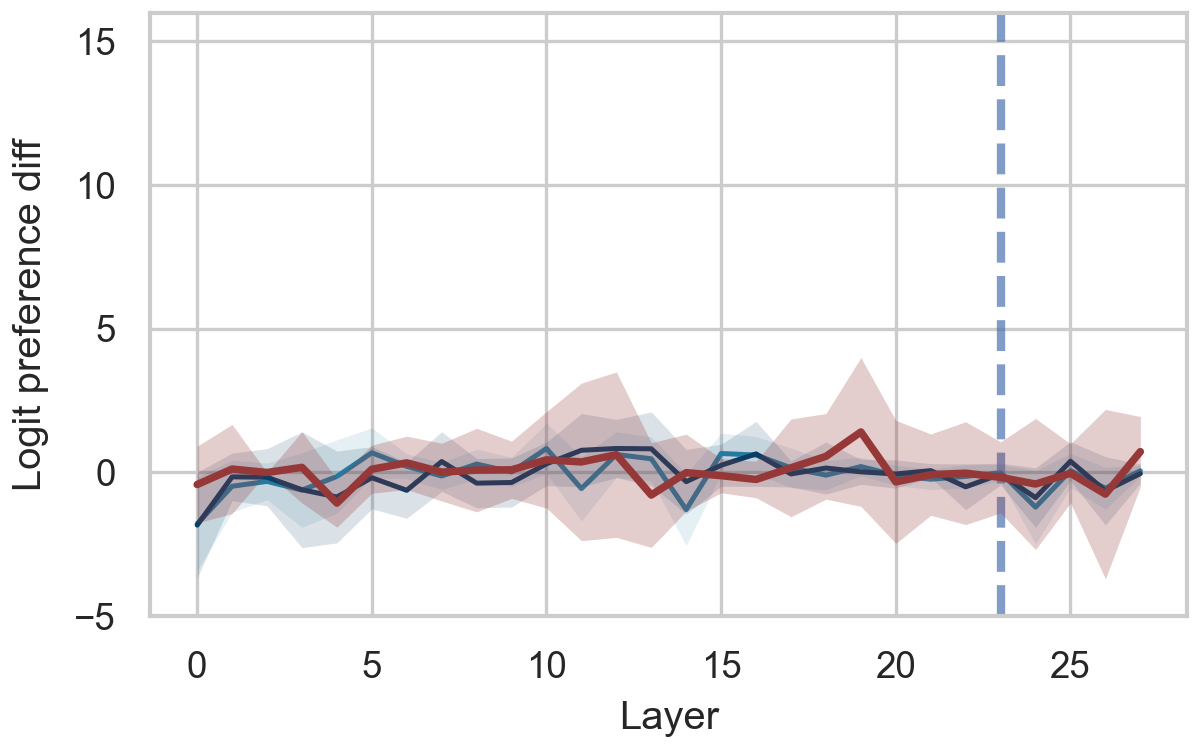}
    \caption{Qwen (norm)}
    \label{fig:cm-norm-qwen}
  \end{subfigure}

  \caption{
    Per-layer intervention experiment results on Clock math task.
    \textbf{Top row}: pure angular interventions.
    \textbf{Bottom row}: pure norm interventions.
    Average logit preference difference is shown:
    blue curves correspond to input-based interventions, interval and month;
    red curves correspond to output-based interventions.
    The blue vertical line marks the original perturbation-based phase-change point.
  }
  
  \label{fig:angle_norm_interventions_cm}
\end{figure}

\paragraph{Angular and norm intervention results.}
\label{extra_angular_norm_int}
We further applied the pure angular and norm interventions introduced in Section~\ref{alternative_intervention_results} to the Days and Clock Math tasks (Figs.~\ref{fig:angle_norm_interventions_dow} and~\ref{fig:angle_norm_interventions_cm}). These results qualitatively match the pattern observed for the Months task (Fig.~\ref{fig:angle_norm_interventions}). Norm-based interventions produce little or no systematic effect across layers, apart from small fluctuations consistent with noise. Input-centric angular interventions are also largely ineffective, with only minimal, localized effects in the early intervention phase for one model. In contrast, output-centric angular interventions are consistently effective in the second phase.

\section{Additional Figures}

\begin{figure}[t]
  \centering

  \begin{subfigure}[t]{0.31\textwidth}
    \centering
    \includegraphics[width=\linewidth]{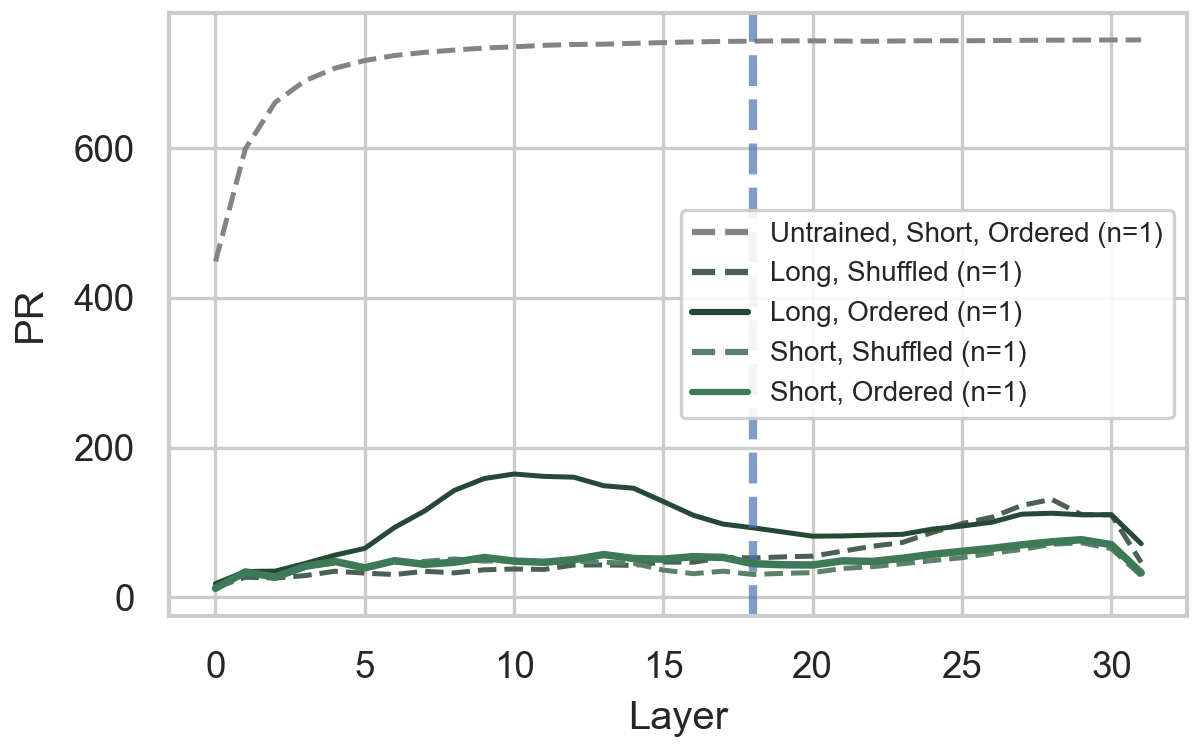}
    \caption{Llama: Identical tokens PR}
    \label{fig:pr-llama-identical}
  \end{subfigure}\hfill
  \begin{subfigure}[t]{0.31\textwidth}
    \centering
    \includegraphics[width=\linewidth]{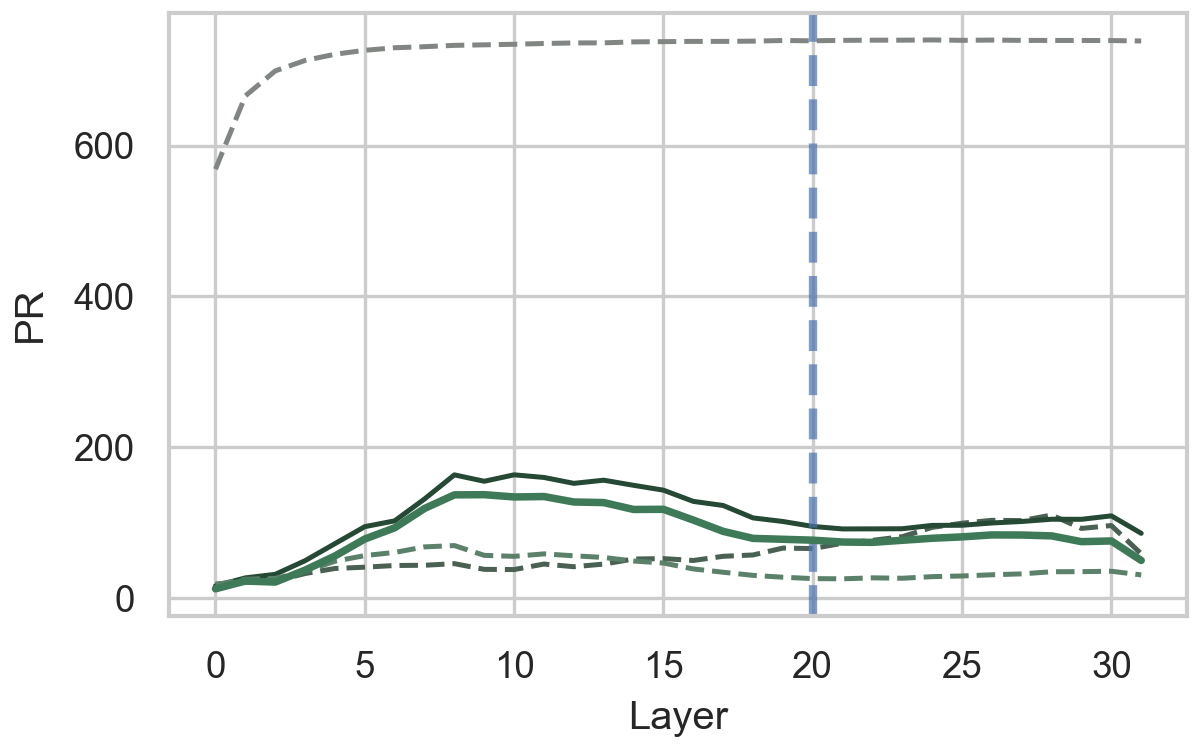}
    \caption{Mistral: Identical tokens PR}
    \label{fig:pr-mistral-identical}
  \end{subfigure}\hfill
  \begin{subfigure}[t]{0.31\textwidth}
    \centering
    \includegraphics[width=\linewidth]{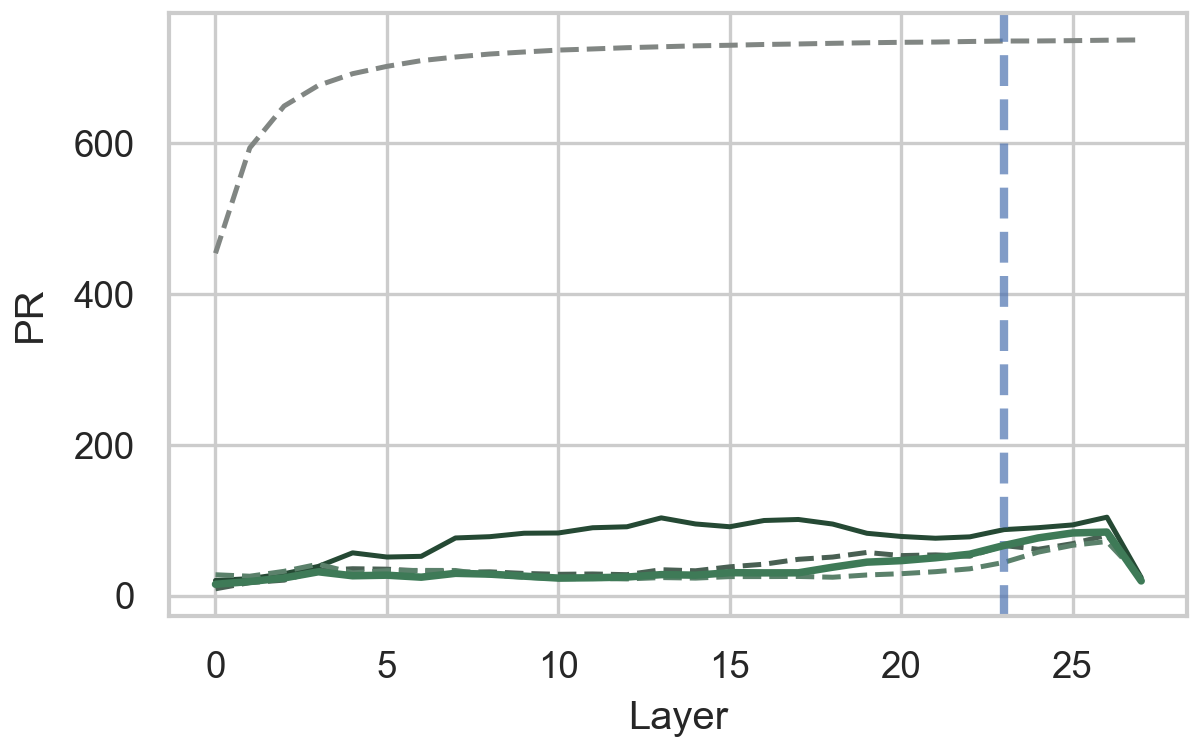}
    \caption{Qwen: Identical tokens PR}
    \label{fig:pr-qwen-identical}
  \end{subfigure}

  \vspace{0.6em}

  \begin{subfigure}[t]{0.31\textwidth}
    \centering
    \includegraphics[width=\linewidth]{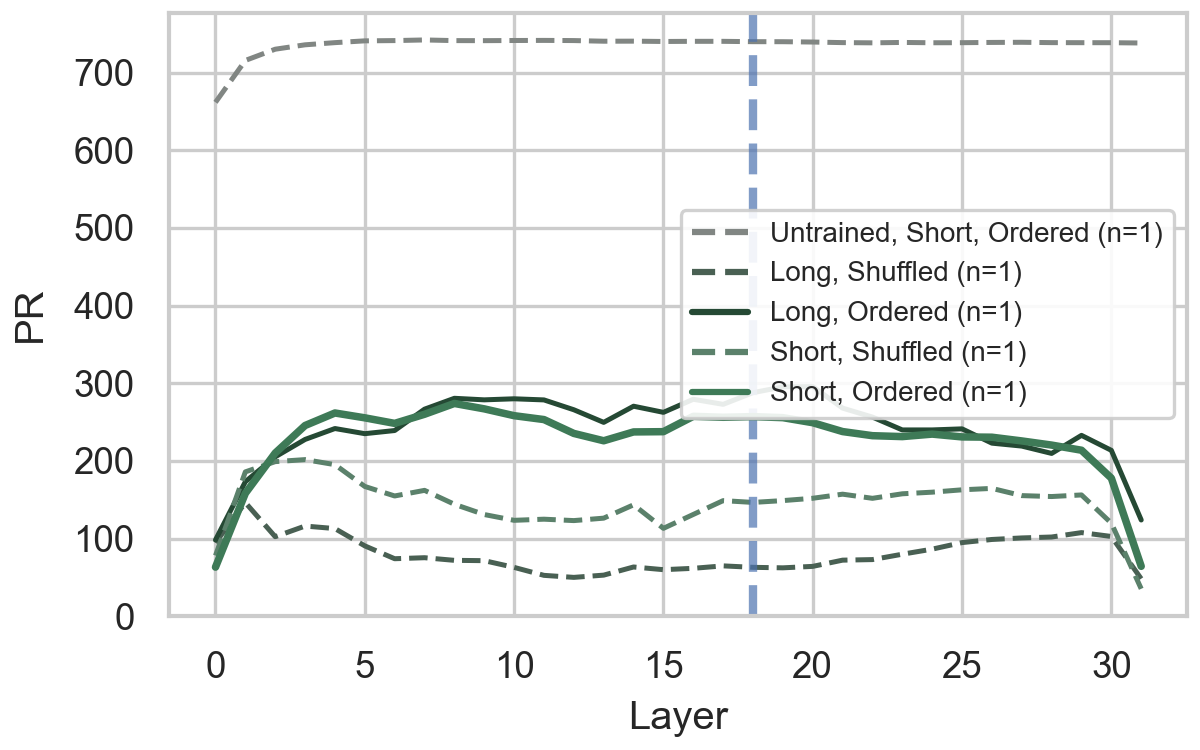}
    \caption{Llama: Non-identical tokens PR}
    \label{fig:pr-llama-nonidentical}
  \end{subfigure}\hfill
  \begin{subfigure}[t]{0.31\textwidth}
    \centering
    \includegraphics[width=\linewidth]{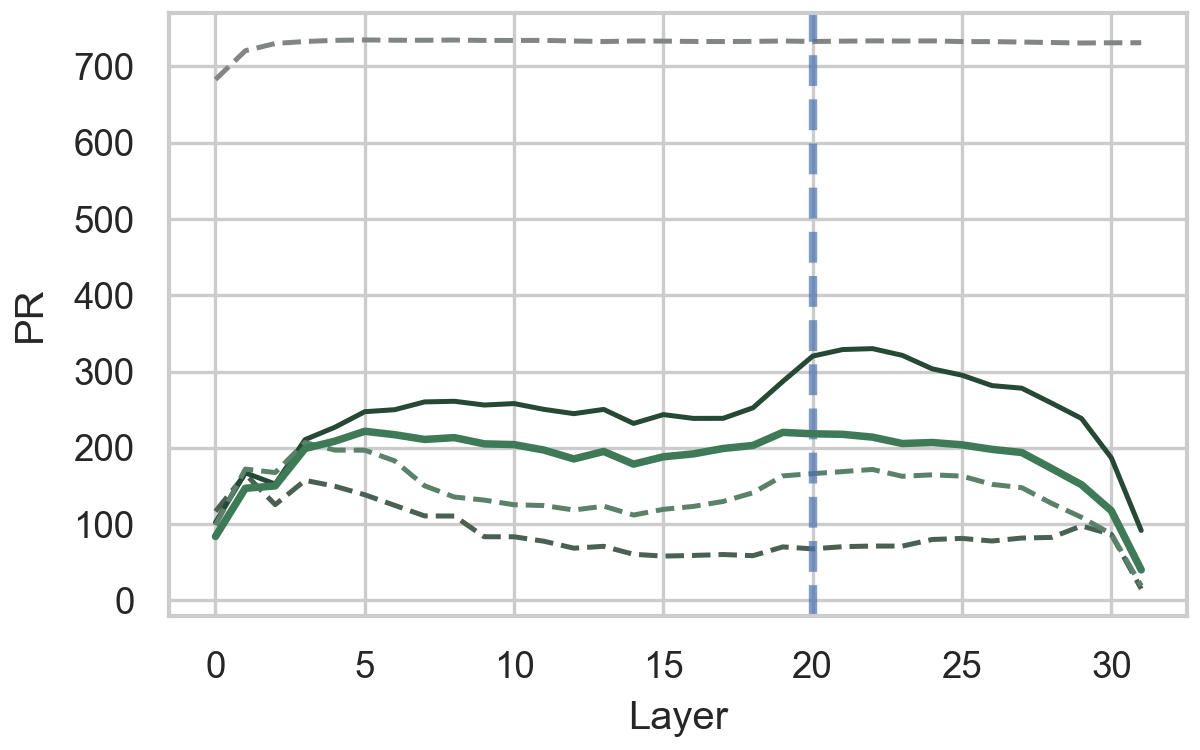}
    \caption{Mistral: Non-identical tokens PR}
    \label{fig:pr-mistral-nonidentical}
  \end{subfigure}\hfill
  \begin{subfigure}[t]{0.31\textwidth}
    \centering
    \includegraphics[width=\linewidth]{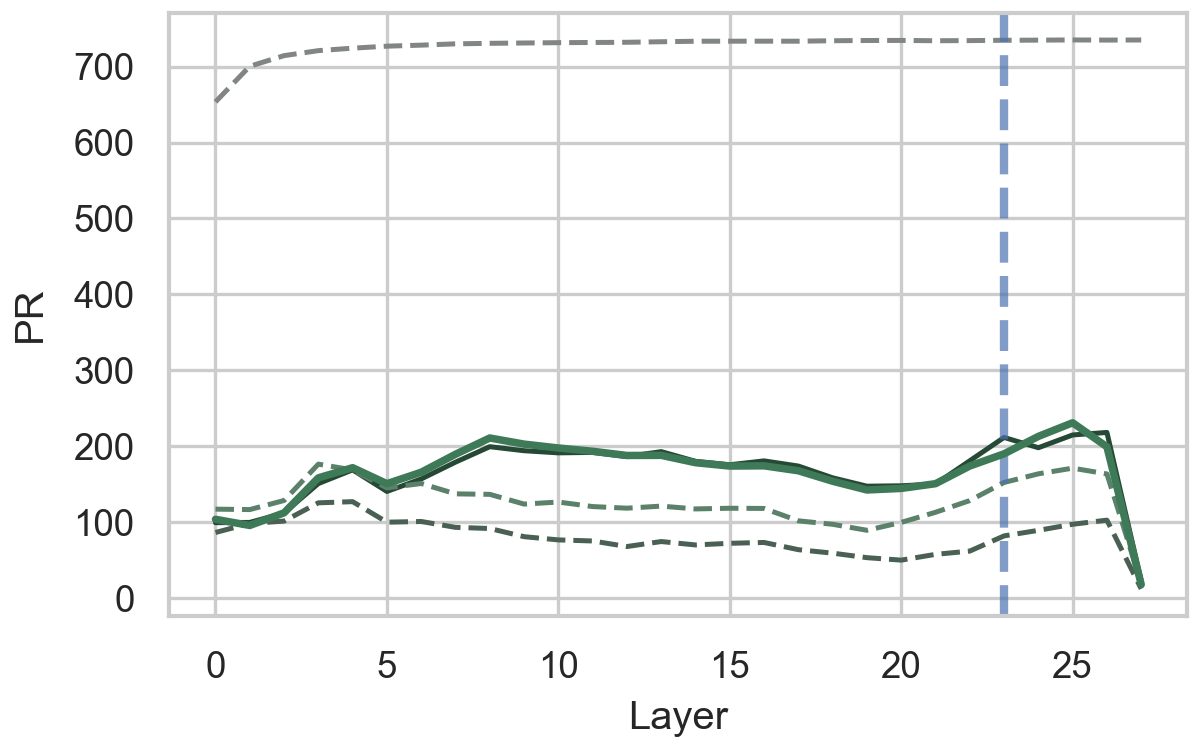}
    \caption{Qwen: Non-identical tokens PR}
    \label{fig:pr-qwen-nonidentical}
  \end{subfigure}

  \caption{
    Per-layer participation ratio (PR) for normalized token representations.
    Columns correspond to models, from left to right: Llama, Mistral, Qwen.
    The blue vertical line marks the perturbation-based phase-change point.
    The top row shows identical tokens, and the bottom row shows non-identical tokens.
  }
  \label{fig:Normalized_PR}
\end{figure}

\begin{figure}[t]
  \centering

  \begin{subfigure}[t]{0.31\textwidth}
    \centering
    \includegraphics[width=\linewidth]{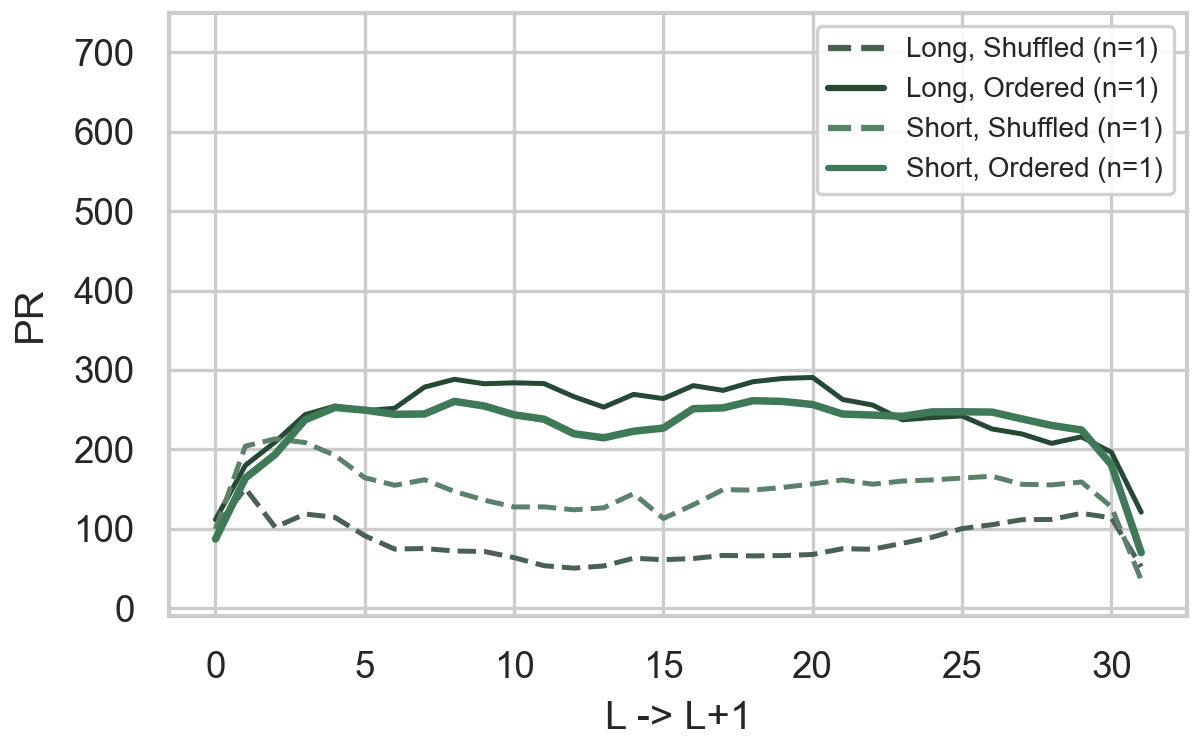}
    \caption{Llama: Non-identical tokens PR}
    \label{fig:third-pr-llama-nonidentical}
  \end{subfigure}\hfill
  \begin{subfigure}[t]{0.31\textwidth}
    \centering
    \includegraphics[width=\linewidth]{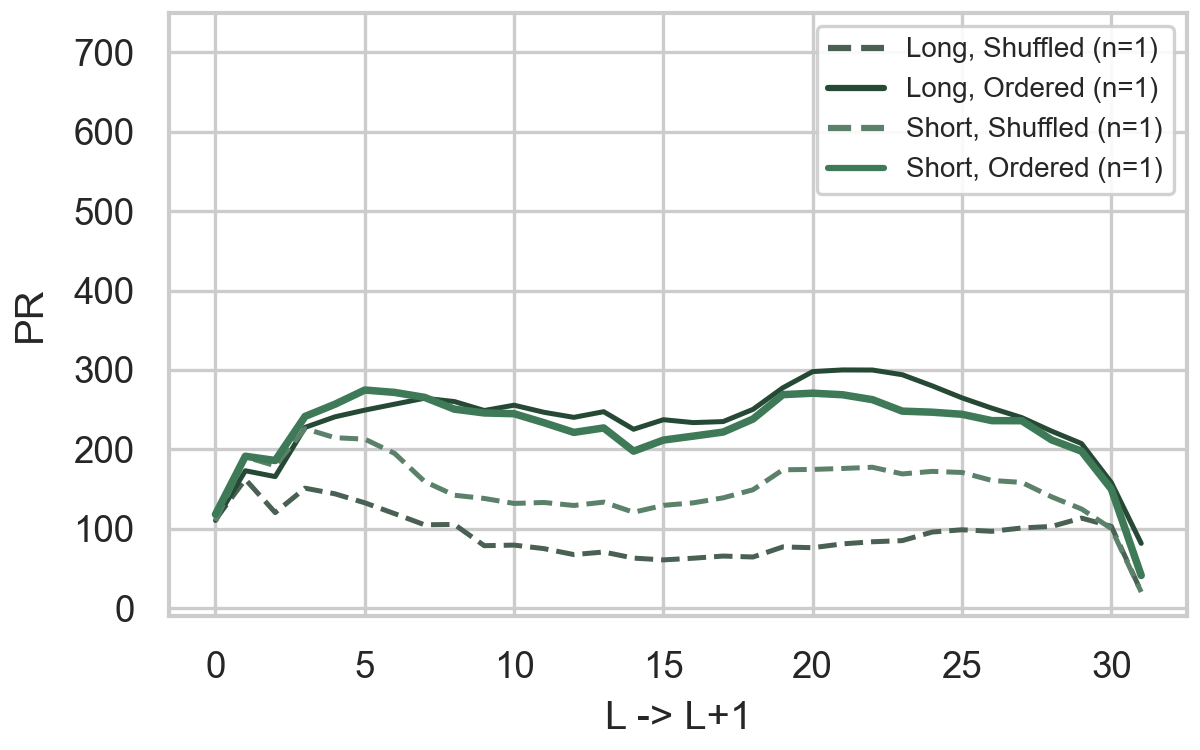}
    \caption{Mistral: Non-identical tokens PR}
    \label{fig:Third-pr-mistral-nonidentical}
  \end{subfigure}\hfill
  \begin{subfigure}[t]{0.31\textwidth}
    \centering
    \includegraphics[width=\linewidth]{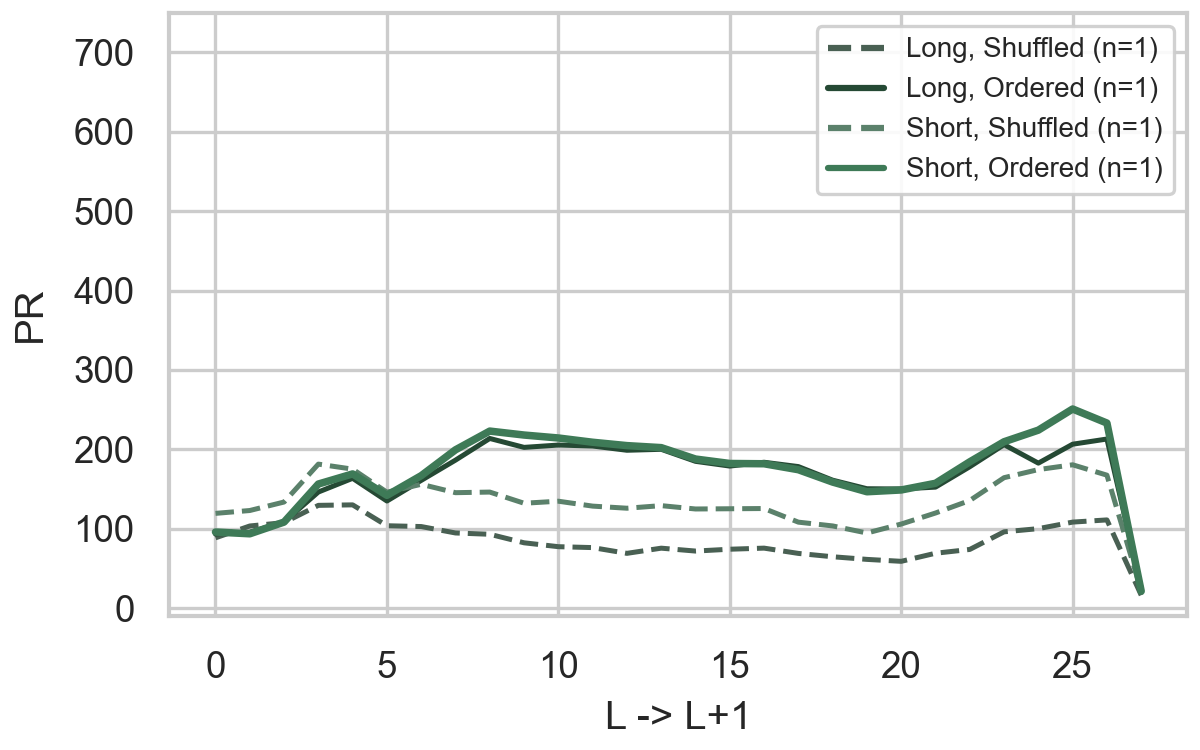}
    \caption{Qwen: Non-identical tokens PR}
    \label{fig:Third-pr-qwen-nonidentical}
  \end{subfigure}

  \caption{
    Per-layer participation ratio (PR) for normalized non-identical (\textit{third from last}) token representations.
    Columns correspond to models, from left to right: Llama, Mistral, Qwen.
  }
  \label{fig:third_from_Normalized_PR}
\end{figure}

\begin{figure}[t]
  \centering

  \begin{subfigure}[t]{0.31\textwidth}
    \centering
    \includegraphics[width=\linewidth]{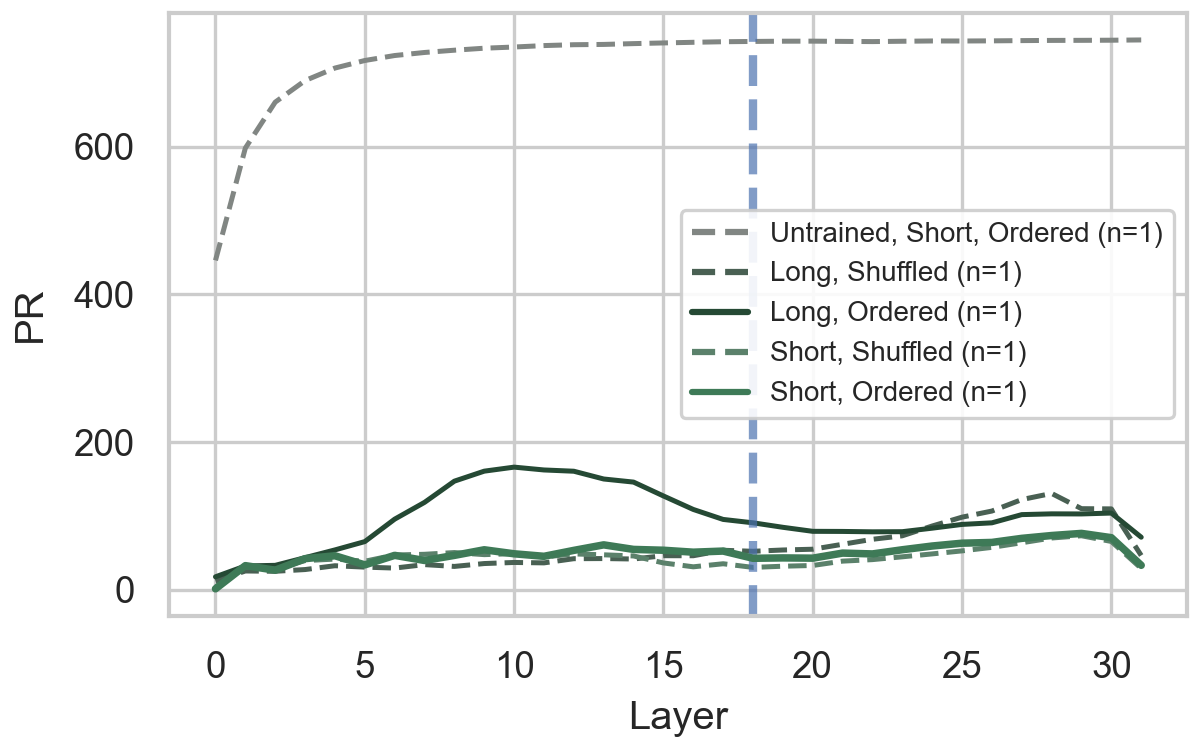}
    \caption{Llama: Identical tokens PR}
    \label{fig:raw-pr-llama-identical}
  \end{subfigure}\hfill
  \begin{subfigure}[t]{0.31\textwidth}
    \centering
    \includegraphics[width=\linewidth]{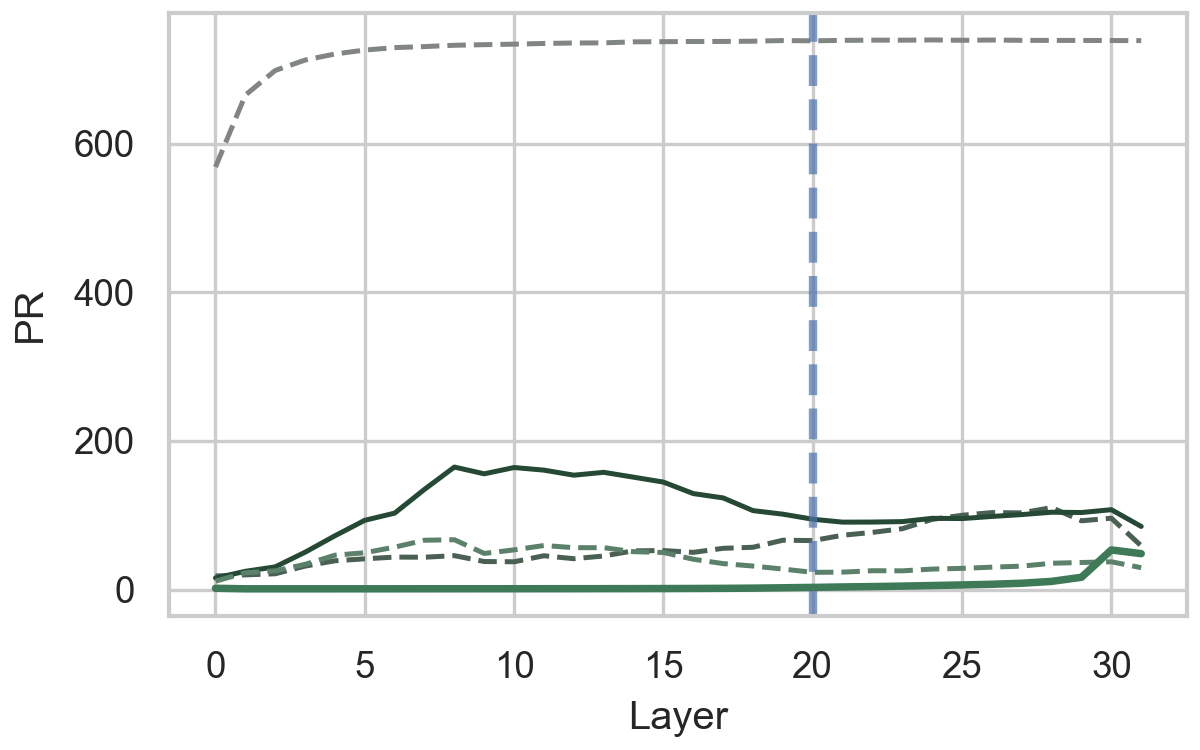}
    \caption{Mistral: Identical tokens PR}
    \label{fig:raw-pr-mistral-identical}
  \end{subfigure}\hfill
  \begin{subfigure}[t]{0.31\textwidth}
    \centering
    \includegraphics[width=\linewidth]{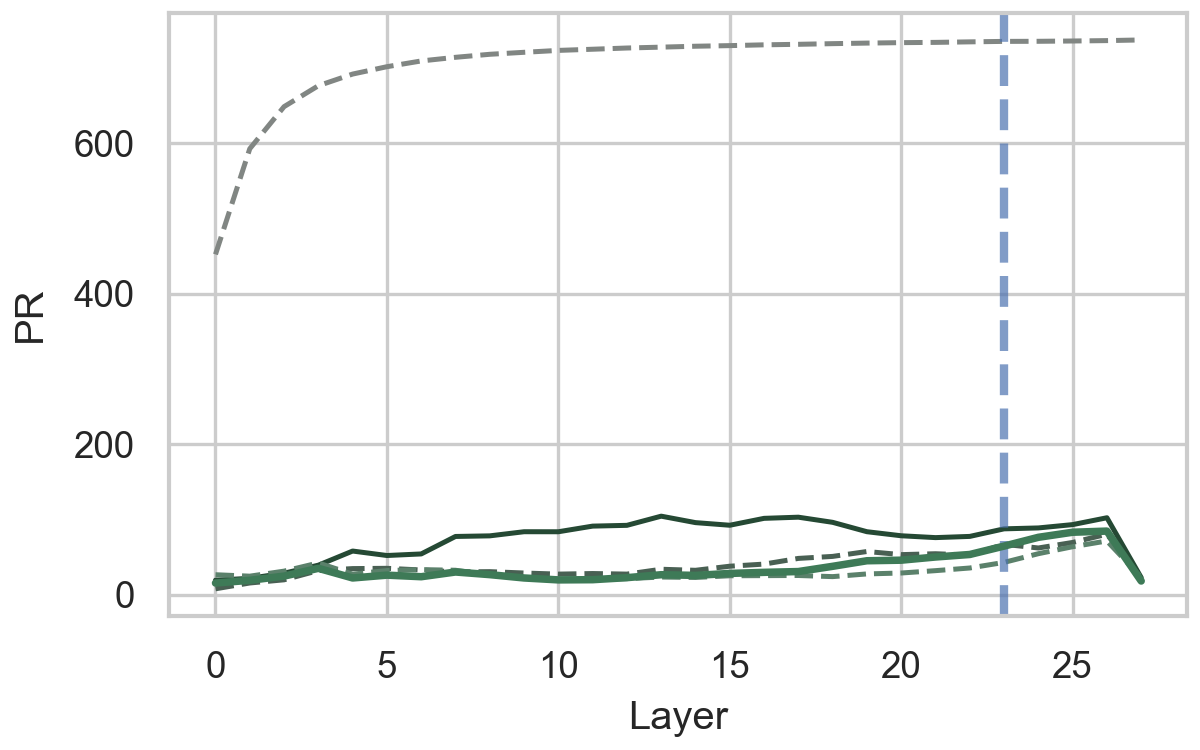}
    \caption{Qwen: Identical tokens PR}
    \label{fig:raw-pr-qwen-identical}
  \end{subfigure}

  \vspace{0.6em}

  \begin{subfigure}[t]{0.31\textwidth}
    \centering
    \includegraphics[width=\linewidth]{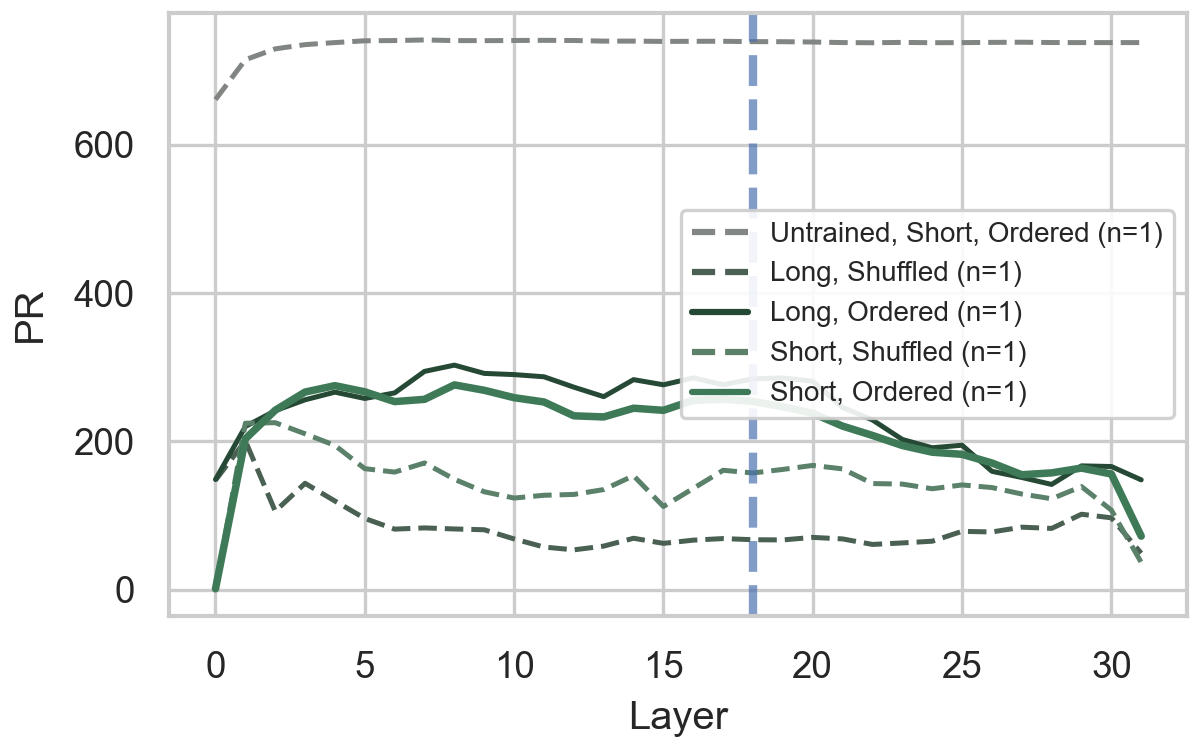}
    \caption{Llama: Non-identical tokens PR}
    \label{fig:raw-pr-llama-nonidentical}
  \end{subfigure}\hfill
  \begin{subfigure}[t]{0.31\textwidth}
    \centering
    \includegraphics[width=\linewidth]{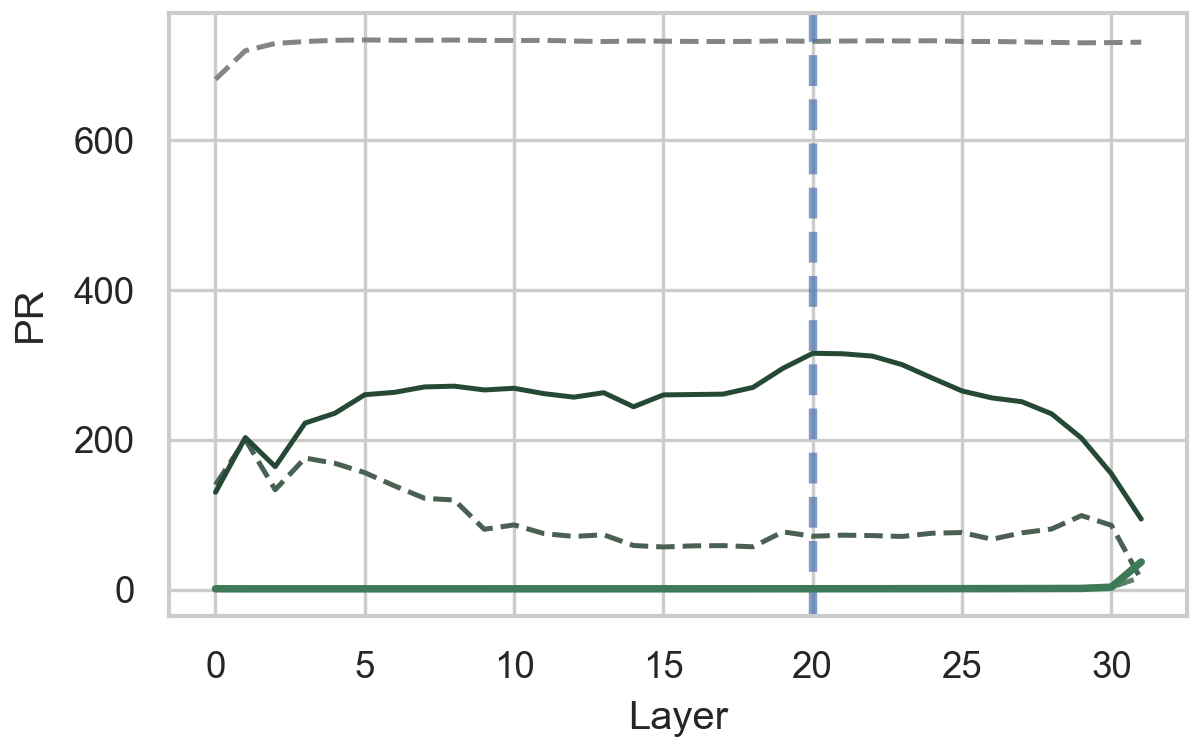}
    \caption{Mistral: Non-identical tokens PR}
    \label{fig:raw-pr-mistral-nonidentical}
  \end{subfigure}\hfill
  \begin{subfigure}[t]{0.31\textwidth}
    \centering
    \includegraphics[width=\linewidth]{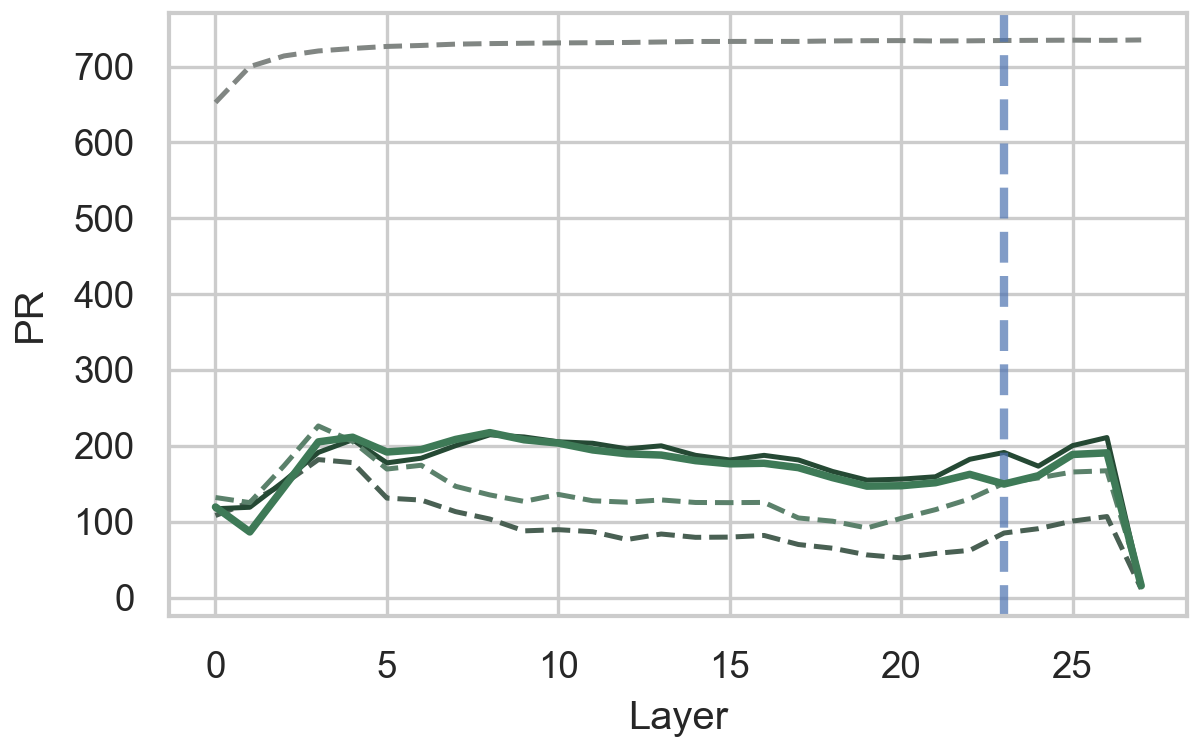}
    \caption{Qwen: Non-identical tokens PR}
    \label{fig:raw-pr-qwen-nonidentical}
  \end{subfigure}

  \caption{
    Per-layer participation ratio (PR) for token representations.
    Columns correspond to models, from left to right: Llama, Mistral, Qwen.
    The top row shows identical tokens, and the bottom row shows non-identical tokens.
    The blue vertical line marks the perturbation-based phase-change point.
  }
  \label{fig:PR}
\end{figure}

\begin{figure}[t]
  \centering

  \begin{subfigure}[t]{0.31\textwidth}
    \centering
    \includegraphics[width=\linewidth]{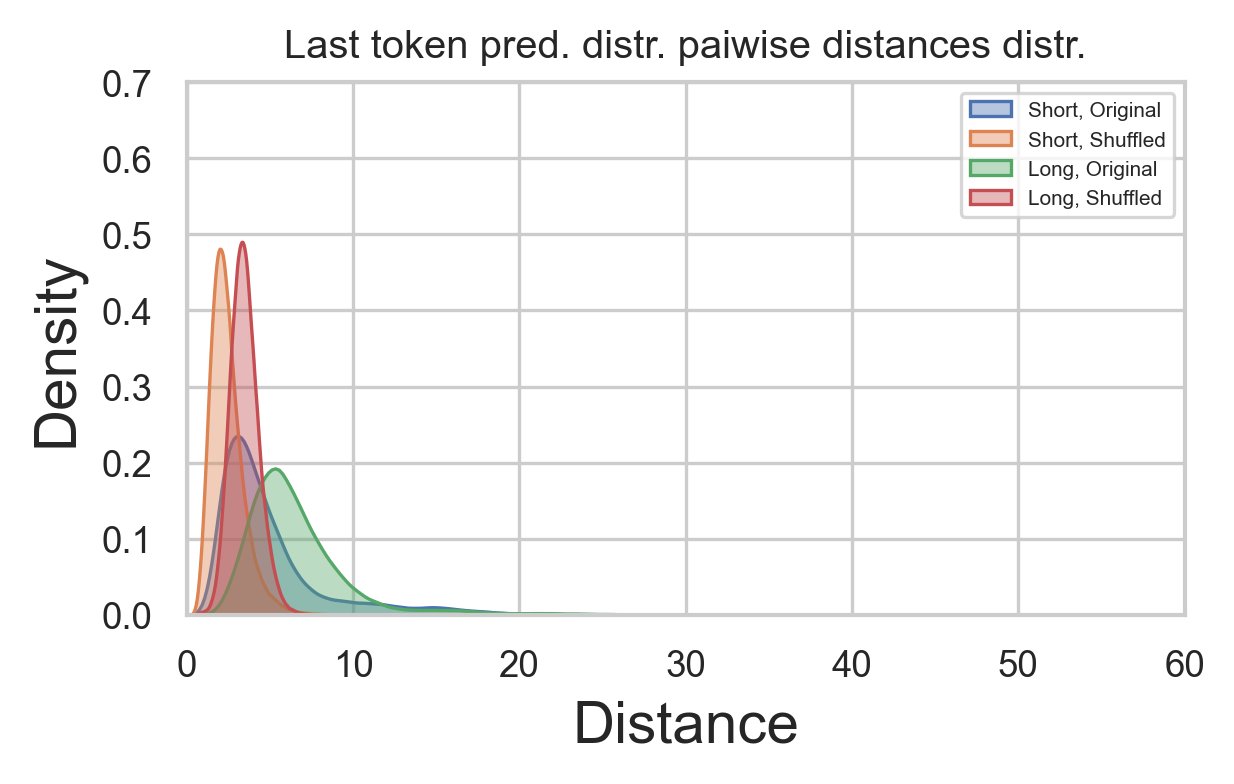}
    \caption{Llama: Identical tokens}
    \label{fig:id-pred-llama}
  \end{subfigure}\hfill
  \begin{subfigure}[t]{0.31\textwidth}
    \centering
    \includegraphics[width=\linewidth]{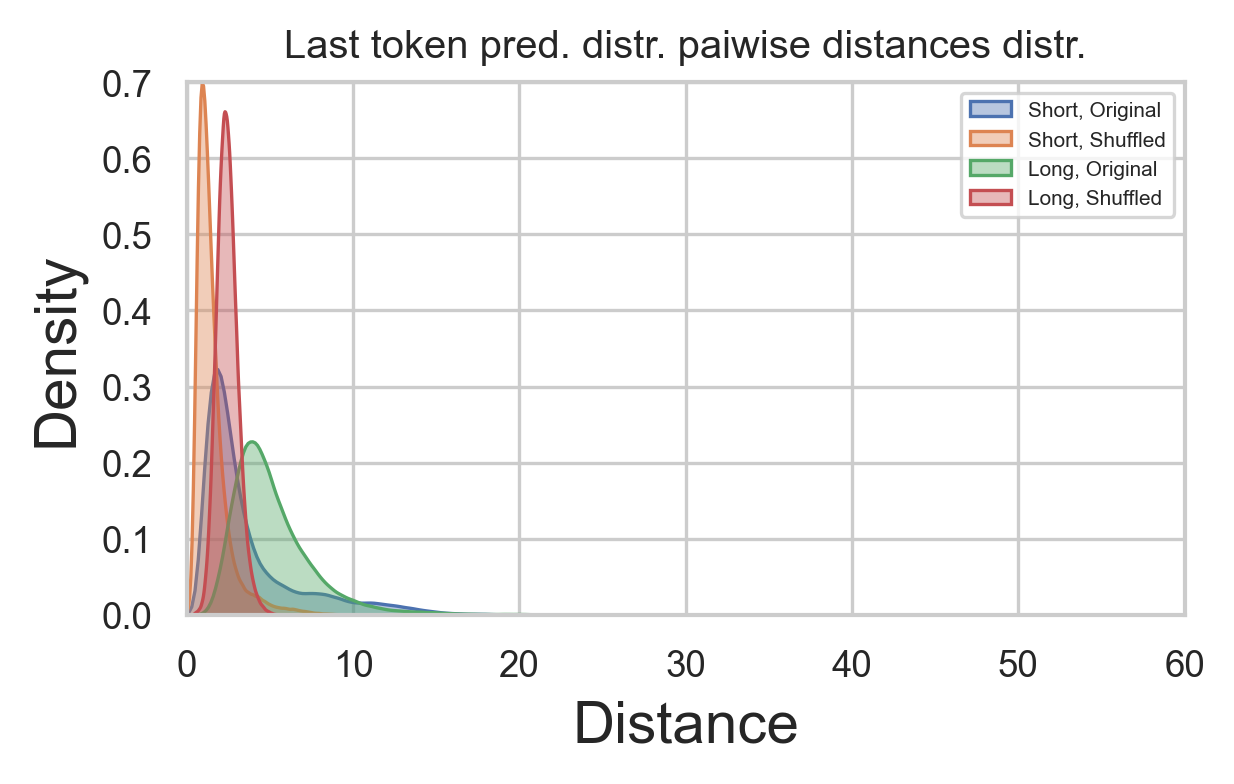}
    \caption{Mistral: Identical tokens}
    \label{fig:id-pred-mistral}
  \end{subfigure}\hfill
  \begin{subfigure}[t]{0.31\textwidth}
    \centering
    \includegraphics[width=\linewidth]{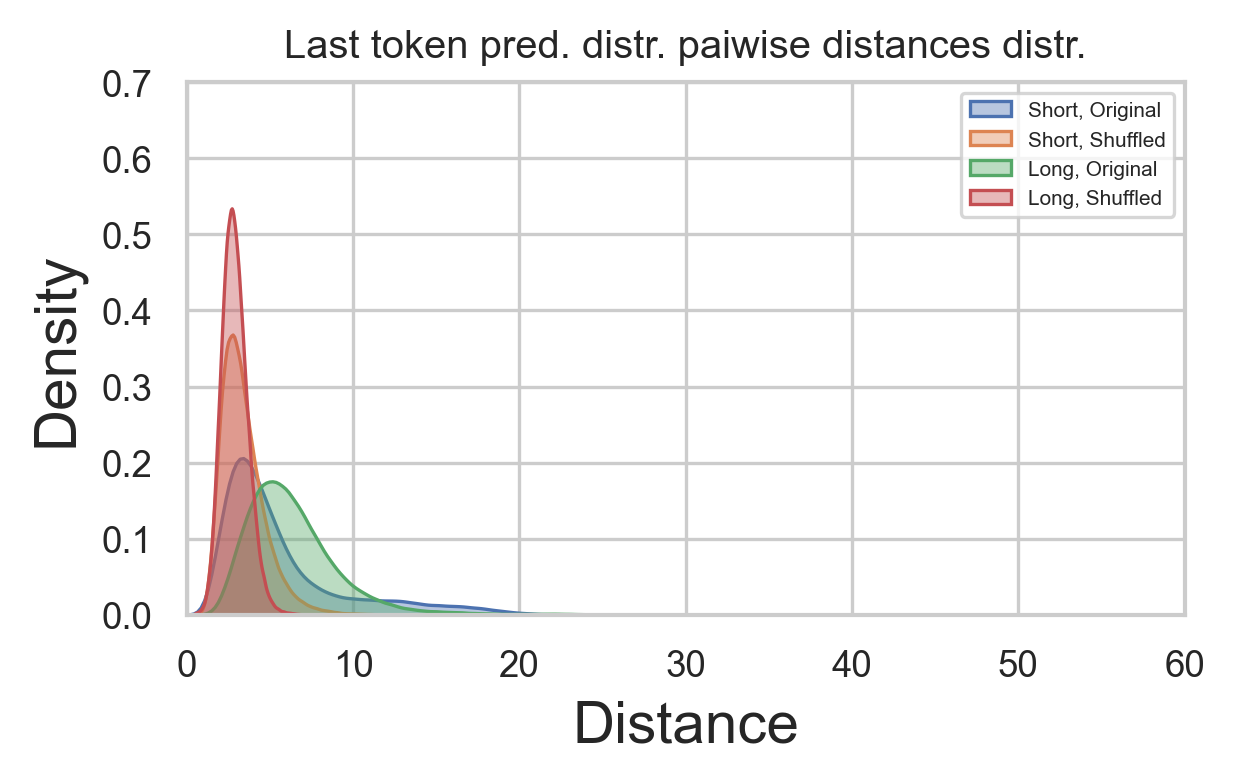}
    \caption{Qwen: Identical tokens}
    \label{fig:id-pred-qwen}
  \end{subfigure}

  \vspace{0.6em}

  \begin{subfigure}[t]{0.31\textwidth}
    \centering
    \includegraphics[width=\linewidth]{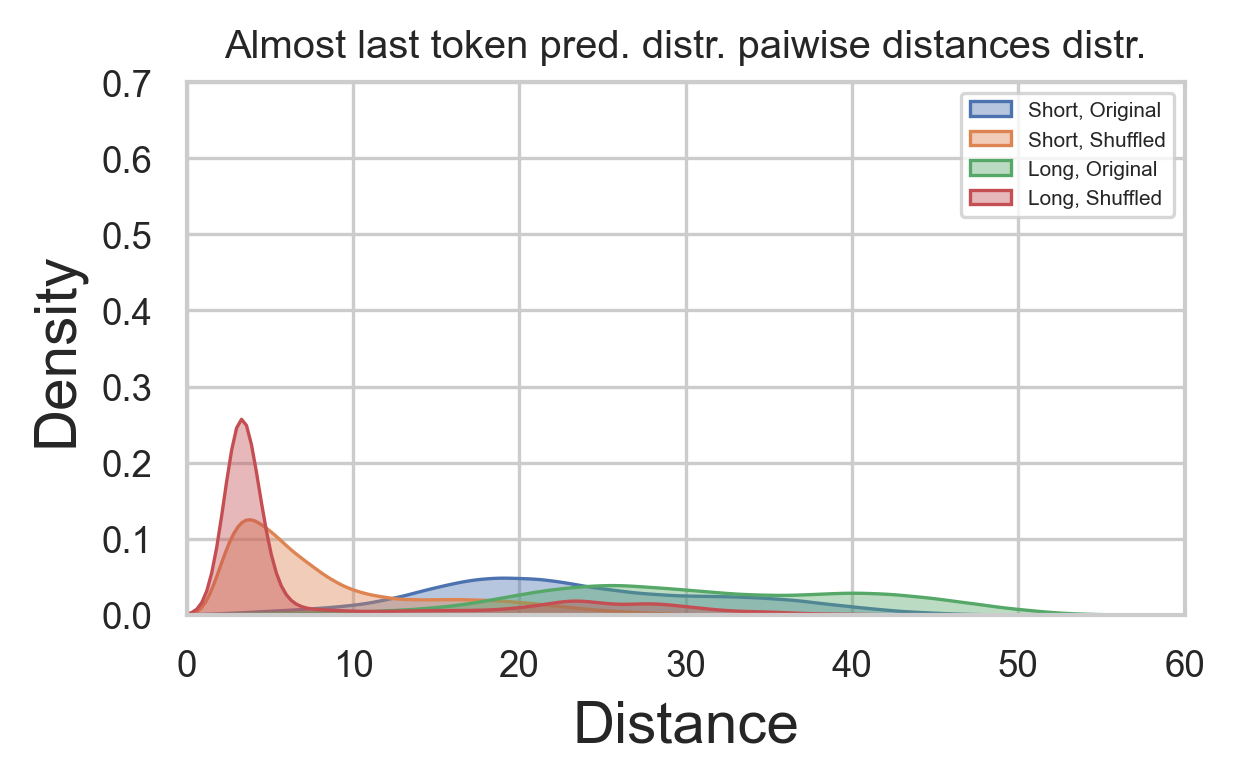}
    \caption{Llama: Non-identical tokens}
    \label{fig:nid-pred-llama}
  \end{subfigure}\hfill
  \begin{subfigure}[t]{0.31\textwidth}
    \centering
    \includegraphics[width=\linewidth]{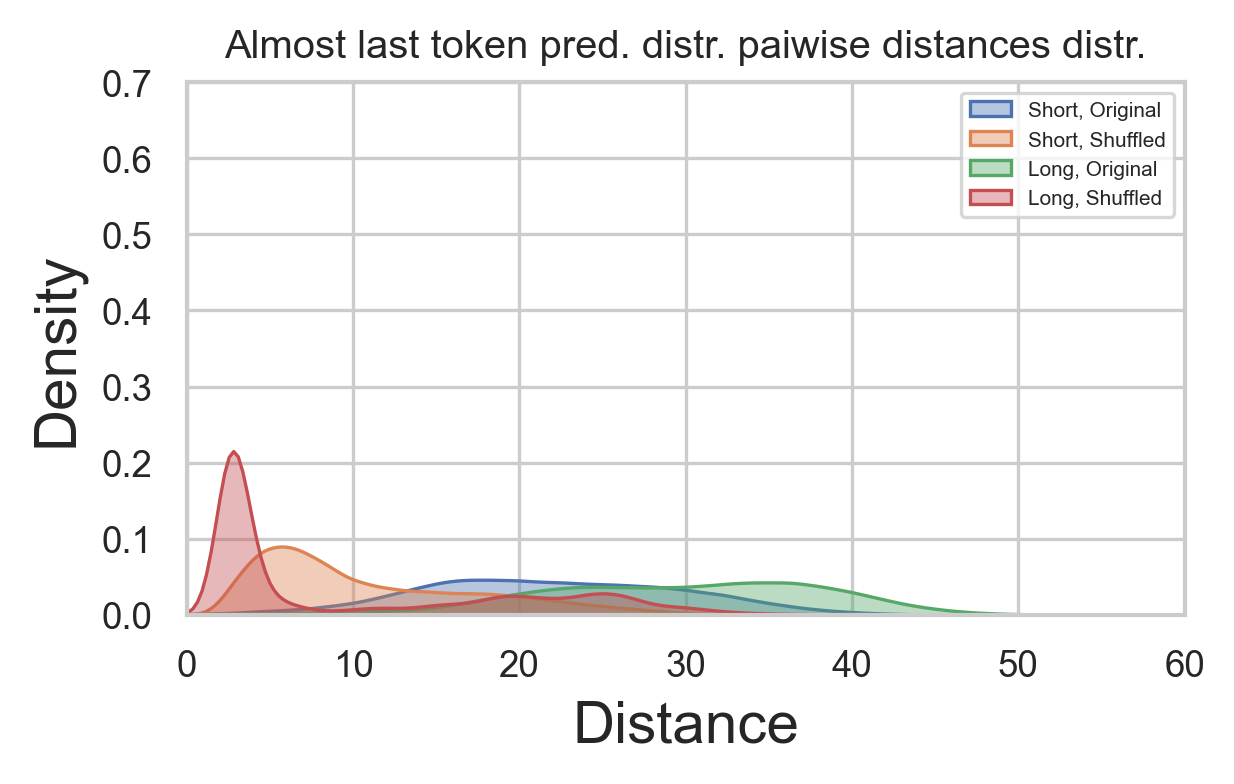}
    \caption{Mistral: Non-identical tokens}
    \label{fig:nid-pred-mist}
  \end{subfigure}\hfill
  \begin{subfigure}[t]{0.31\textwidth}
    \centering
    \includegraphics[width=\linewidth]{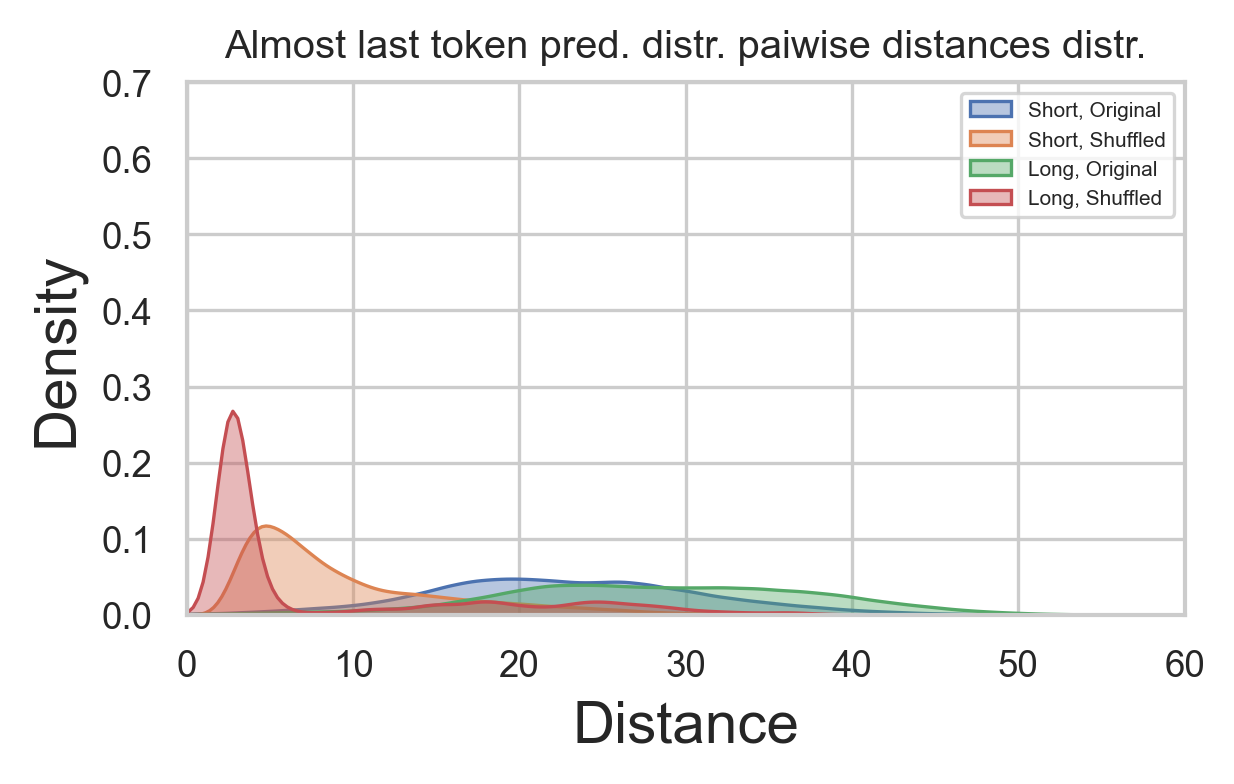}
    \caption{Qwen: Non-identical tokens}
    \label{fig:nid-pred-qwen}
  \end{subfigure}

  \caption{Prediction dissimilarity distributions. For each model, we show kernel density estimates of the empirical distributions of pairwise symmetric KL divergences between prediction distributions, separately for identical tokens (top) and non-identical tokens (bottom).}
  \label{fig:pred_dist_sim}
\end{figure}

\begin{figure}[t]
  \centering

  \begin{subfigure}[t]{0.31\textwidth}
    \centering
    \includegraphics[width=\linewidth]{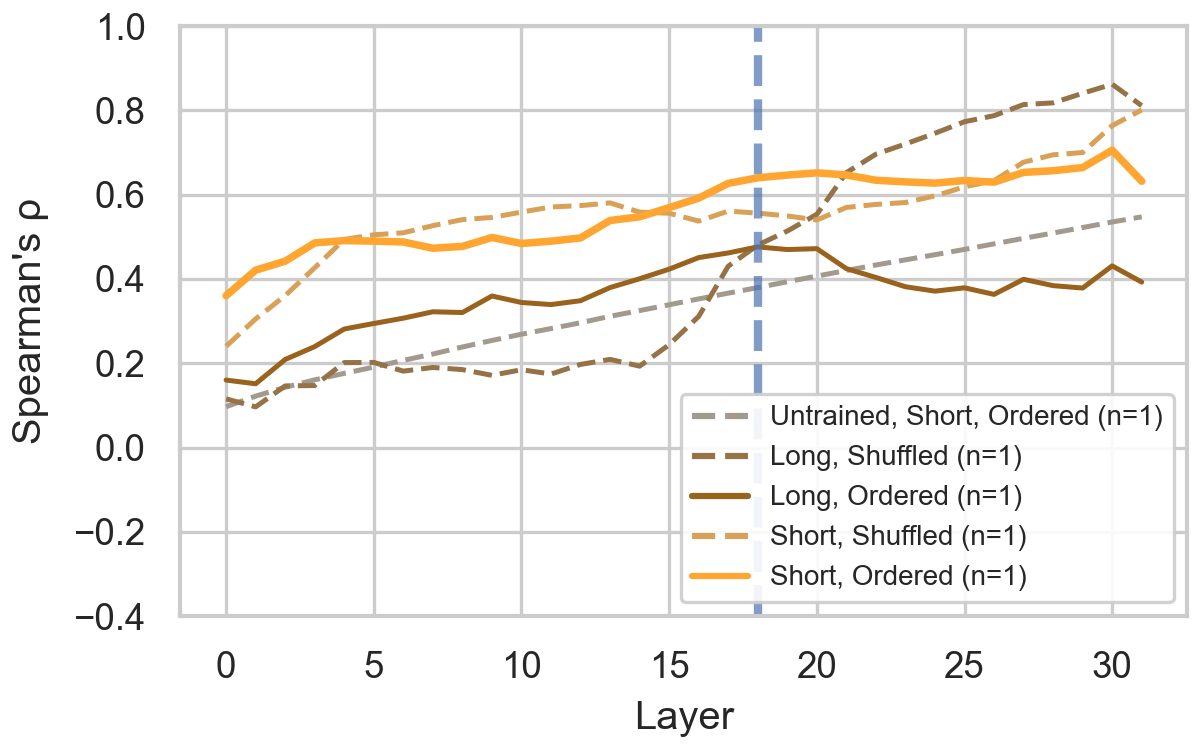}
    \caption{Llama: angular distances, predictions correlation}
    \label{fig:id-ang-llama-ang}
  \end{subfigure}\hfill
  \begin{subfigure}[t]{0.31\textwidth}
    \centering
    \includegraphics[width=\linewidth]{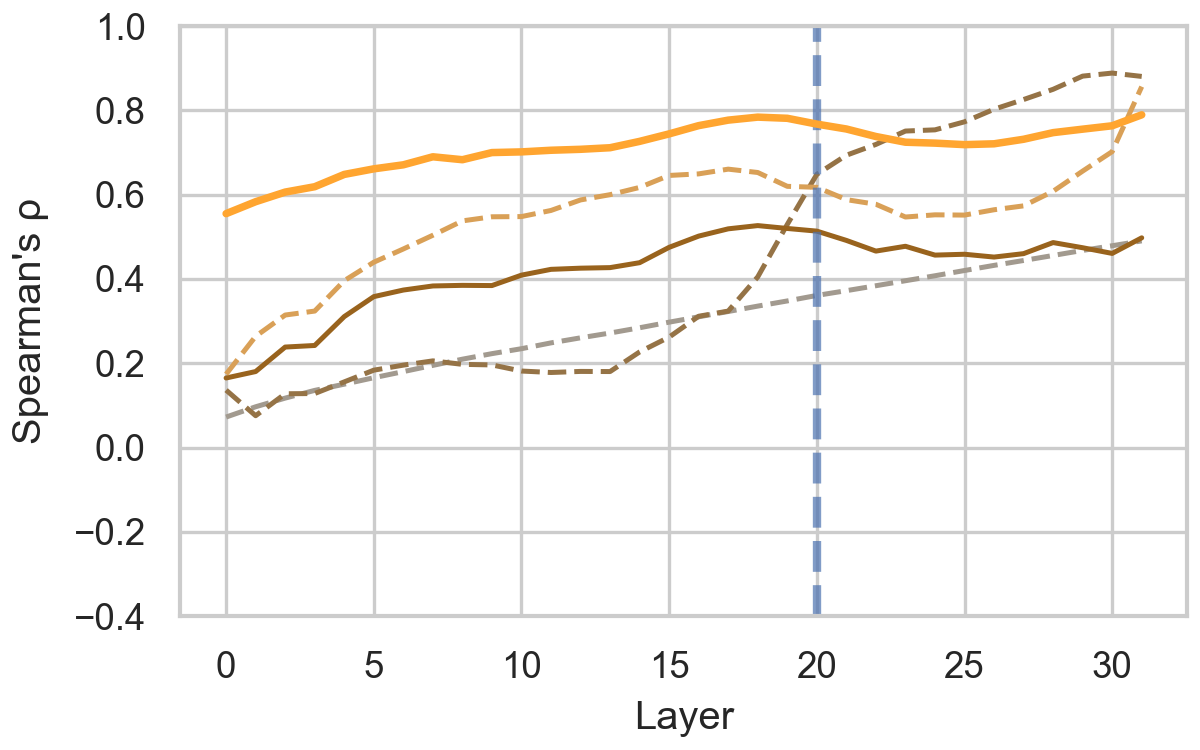}
    \caption{Mistral: angular distances, predictions correlation}
    \label{fig:id-ang-mistral-ang}
  \end{subfigure}\hfill
  \begin{subfigure}[t]{0.31\textwidth}
    \centering
    \includegraphics[width=\linewidth]{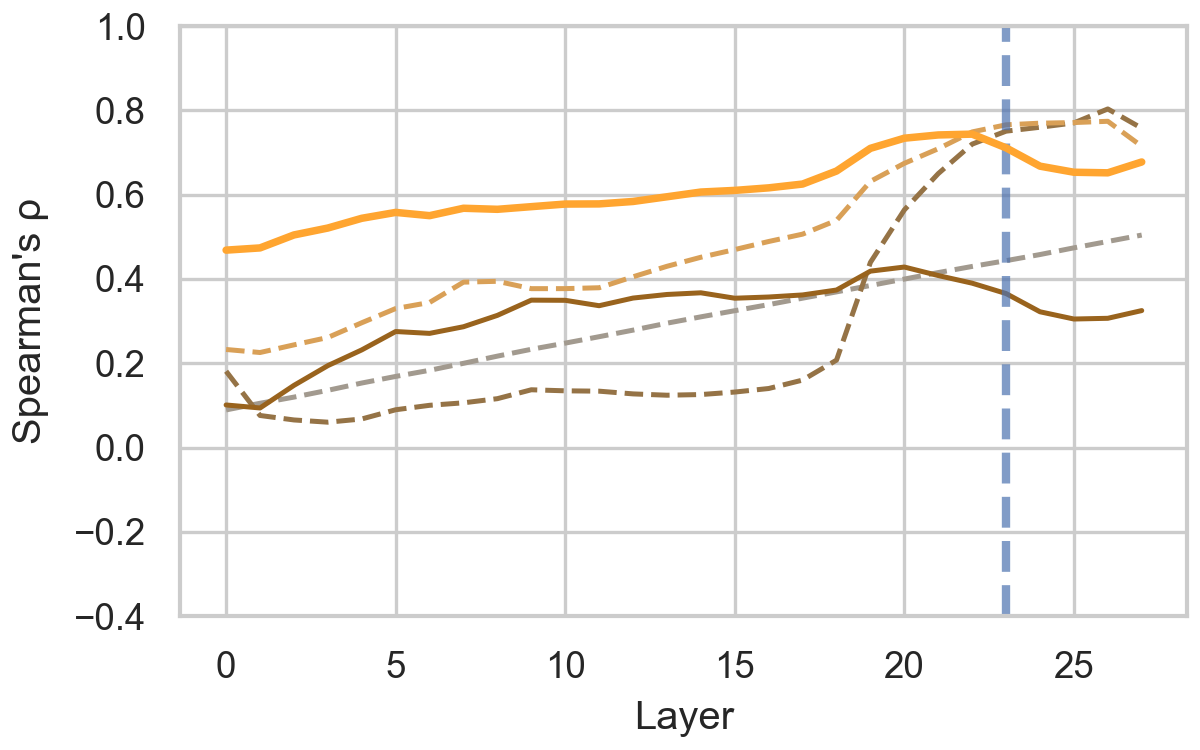}
    \caption{Qwen: angular distances, predictions correlation}
    \label{fig:id-ang-qwen-ang}
  \end{subfigure}

  \vspace{0.6em}

  \begin{subfigure}[t]{0.31\textwidth}
    \centering
    \includegraphics[width=\linewidth]{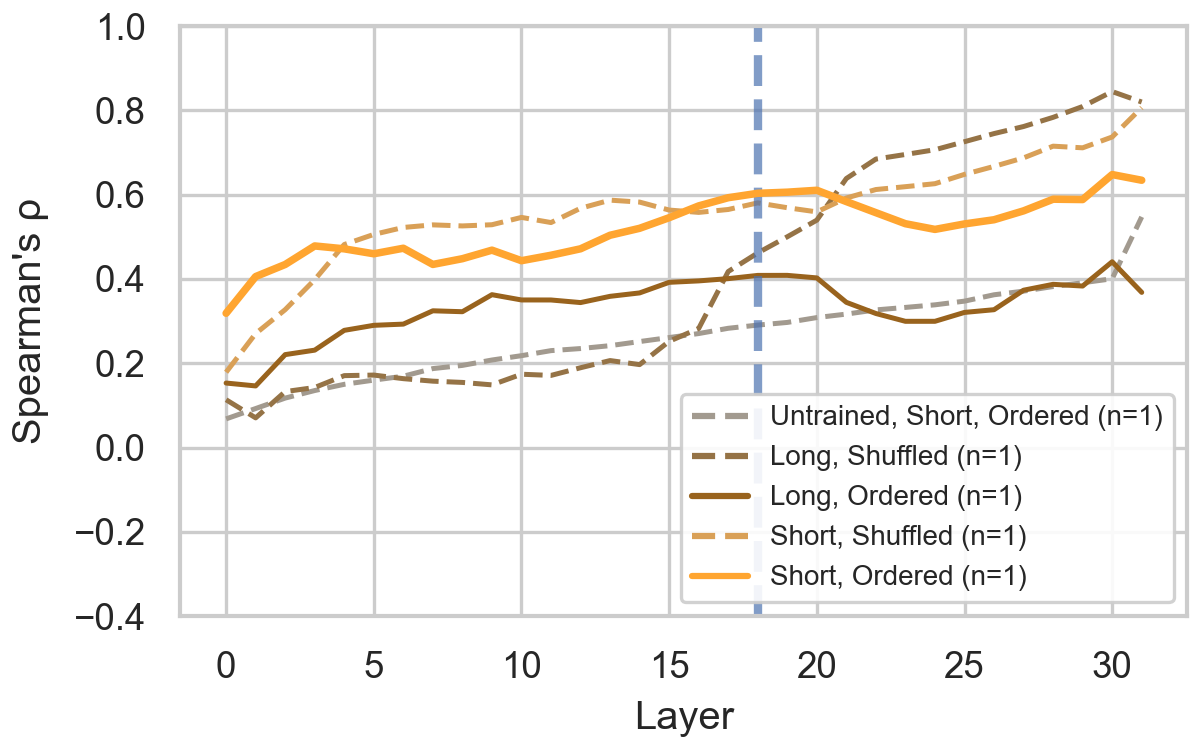}
    \caption{Llama: Euc. distances, predictions correlation}
    \label{fig:id-ang-llama-euc}
  \end{subfigure}\hfill
  \begin{subfigure}[t]{0.31\textwidth}
    \centering
    \includegraphics[width=\linewidth]{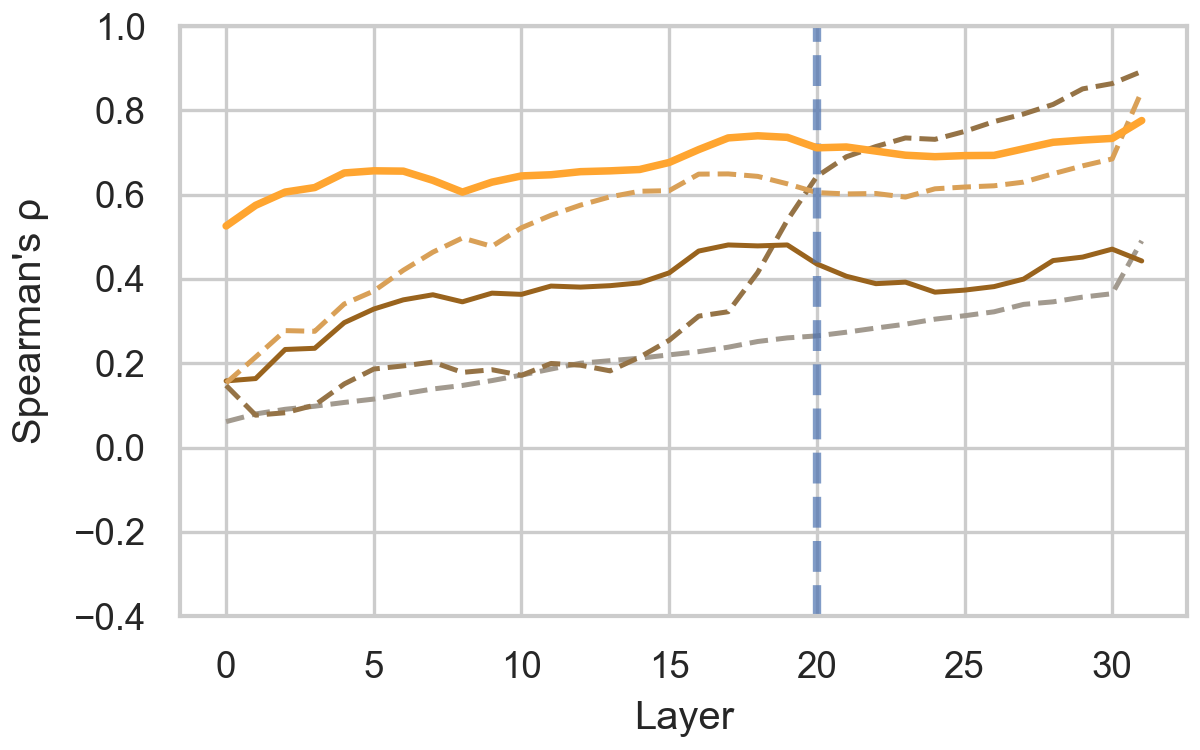}
    \caption{Mistral: Euc. distances, predictions correlation}
    \label{fig:id-ang-mistral-euc}
  \end{subfigure}\hfill
  \begin{subfigure}[t]{0.31\textwidth}
    \centering
    \includegraphics[width=\linewidth]{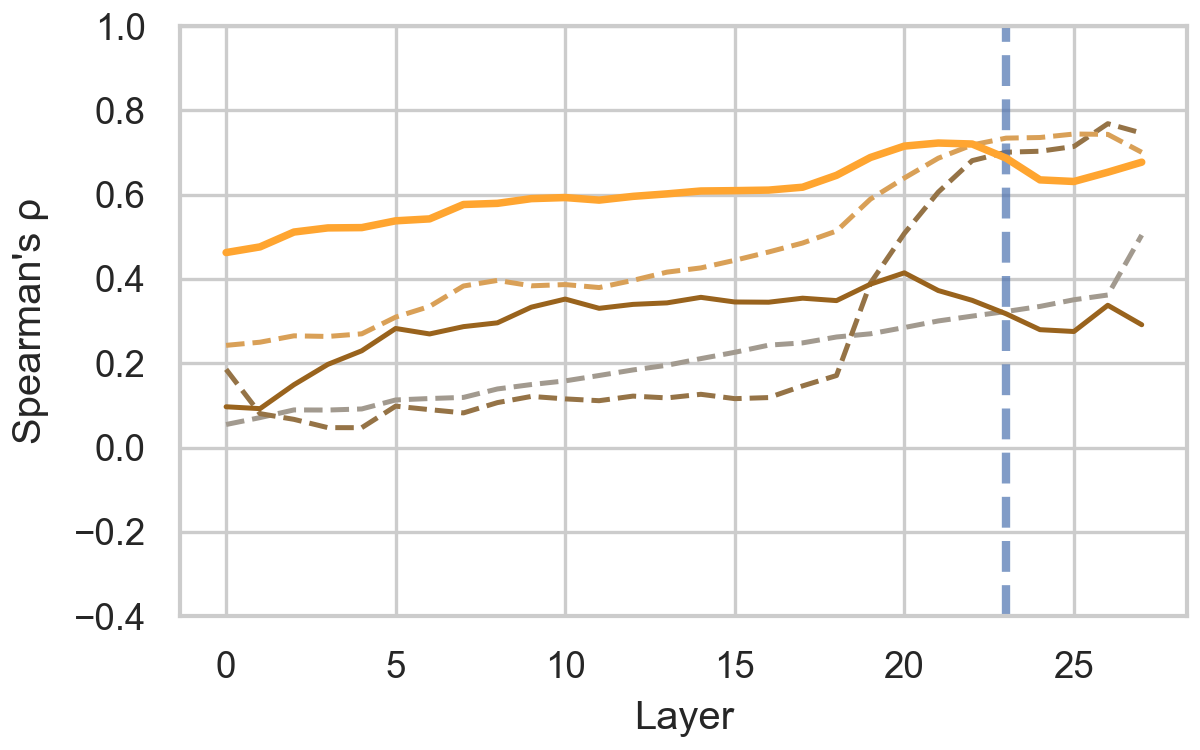}
    \caption{Qwen: Euc. distances, predictions correlation}
    \label{fig:id-ang-qwen-euc}
  \end{subfigure}

  \caption{
    Per-layer correlation between identical tokens' pairwise distances and pairwise prediction-distribution symmetric KL divergence.
    The top row shows angular distances, and the bottom row shows Euclidean distances.
    Columns correspond to models, from left to right: Llama, Mistral, Qwen.
    The blue vertical line marks the perturbation-based phase-change point.
  }
  \label{fig:identical_correlations}
\end{figure}

\begin{figure}[t]
  \centering

  \begin{subfigure}[t]{0.31\textwidth}
    \centering
    \includegraphics[width=\linewidth]{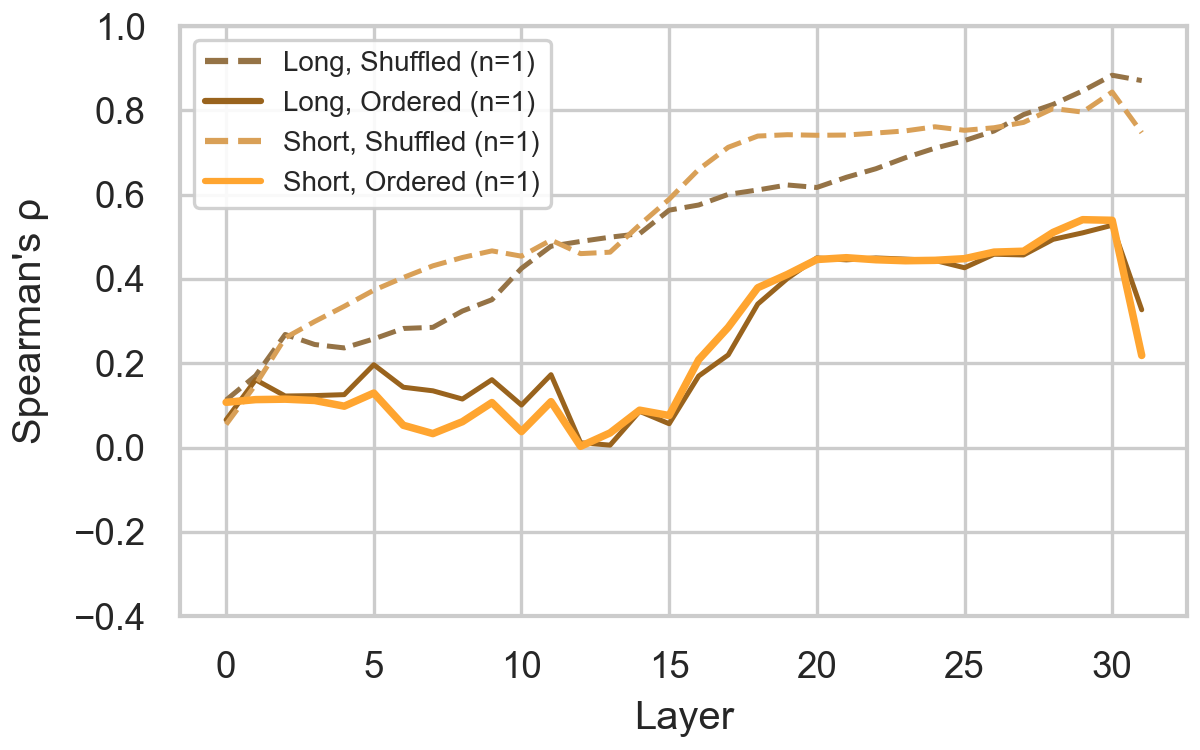}
    \caption{Llama: angular distances, predictions correlation}
    \label{fig:alt-ang-llama-ang}
  \end{subfigure}\hfill
  \begin{subfigure}[t]{0.31\textwidth}
    \centering
    \includegraphics[width=\linewidth]{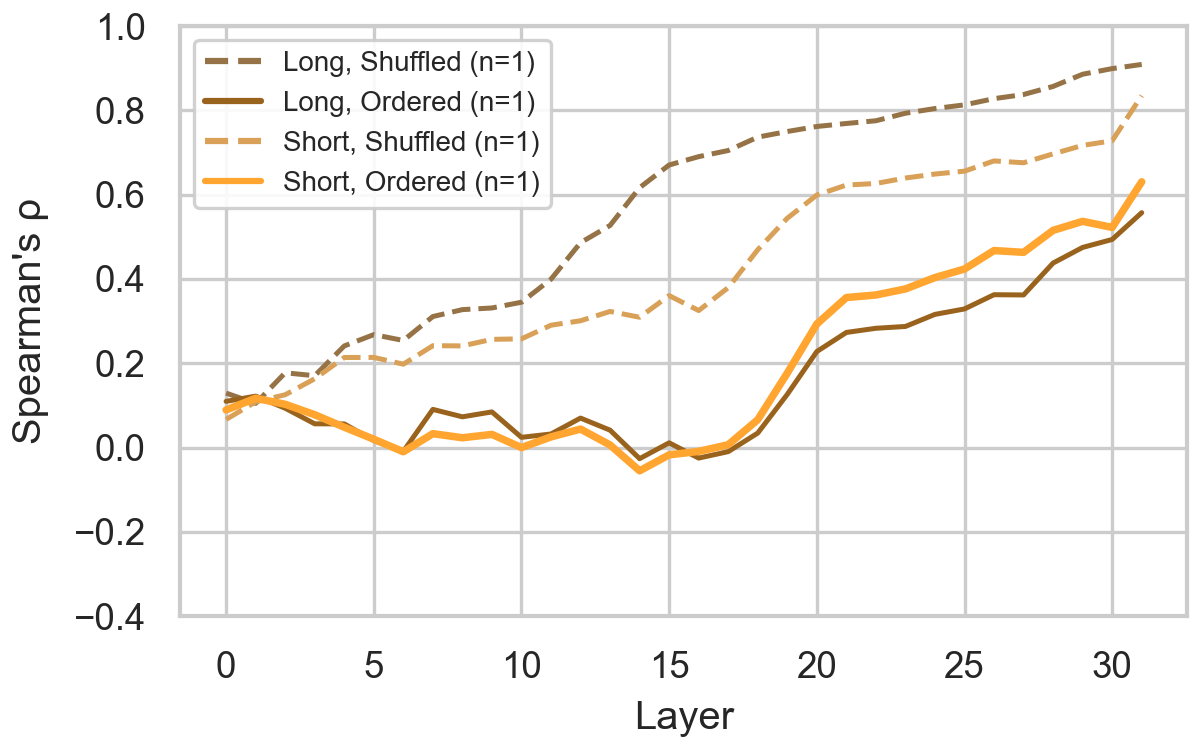}
    \caption{Mistral: angular distances, predictions correlation}
    \label{fig:alt-ang-mistral-ang}
  \end{subfigure}\hfill
  \begin{subfigure}[t]{0.31\textwidth}
    \centering
    \includegraphics[width=\linewidth]{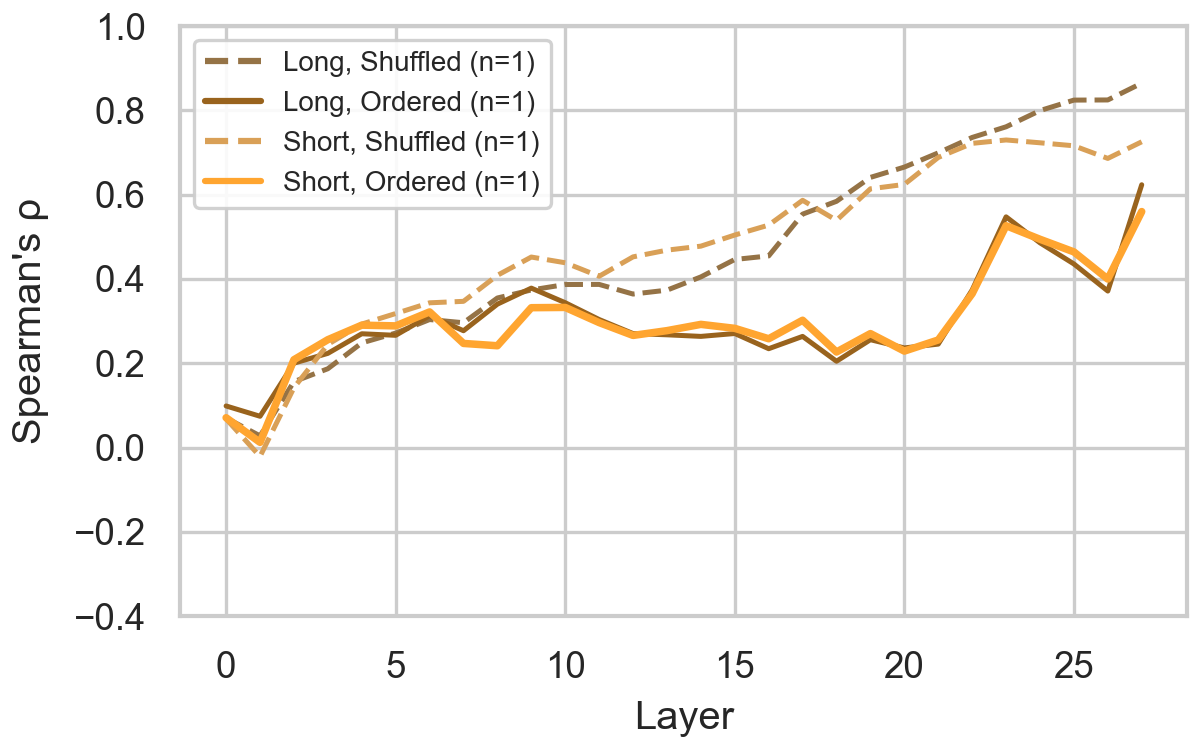}
    \caption{Qwen: angular distances, predictions correlation}
    \label{fig:alt-ang-qwen-ang}
  \end{subfigure}

  \vspace{0.6em}

  \begin{subfigure}[t]{0.31\textwidth}
    \centering
    \includegraphics[width=\linewidth]{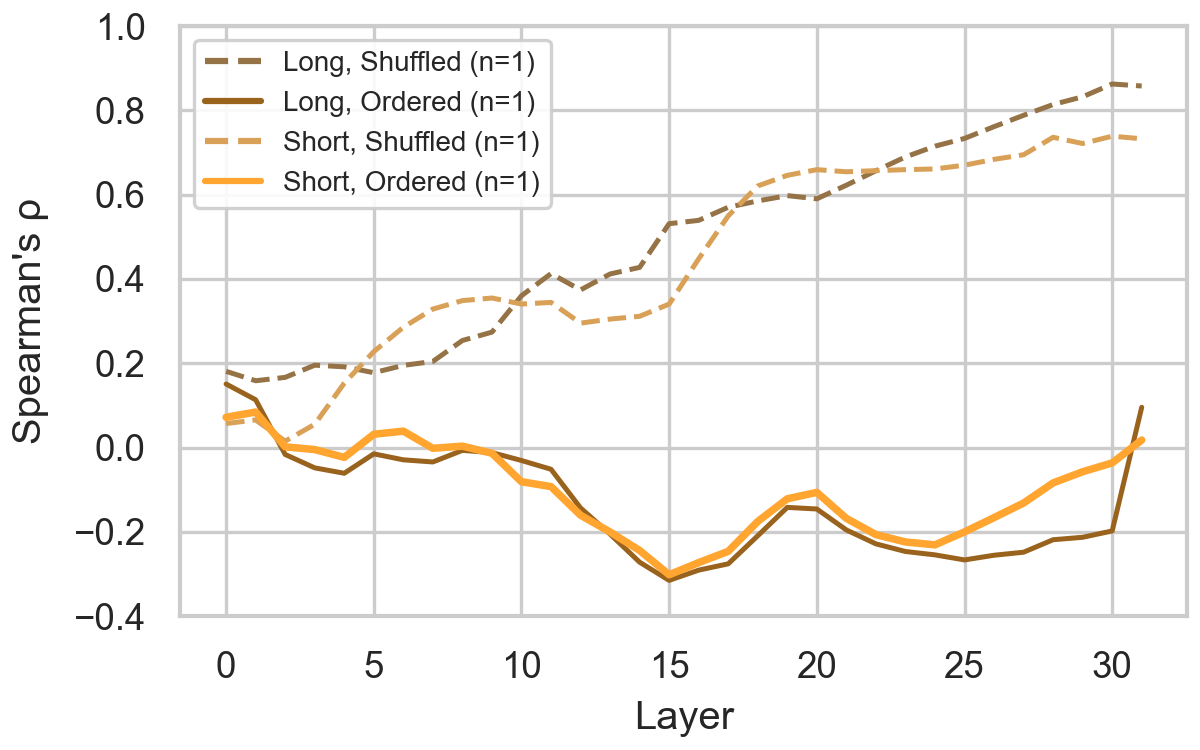}
    \caption{Llama: Euc. distances, predictions correlation}
    \label{fig:alt-ang-llama-euc}
  \end{subfigure}\hfill
  \begin{subfigure}[t]{0.31\textwidth}
    \centering
    \includegraphics[width=\linewidth]{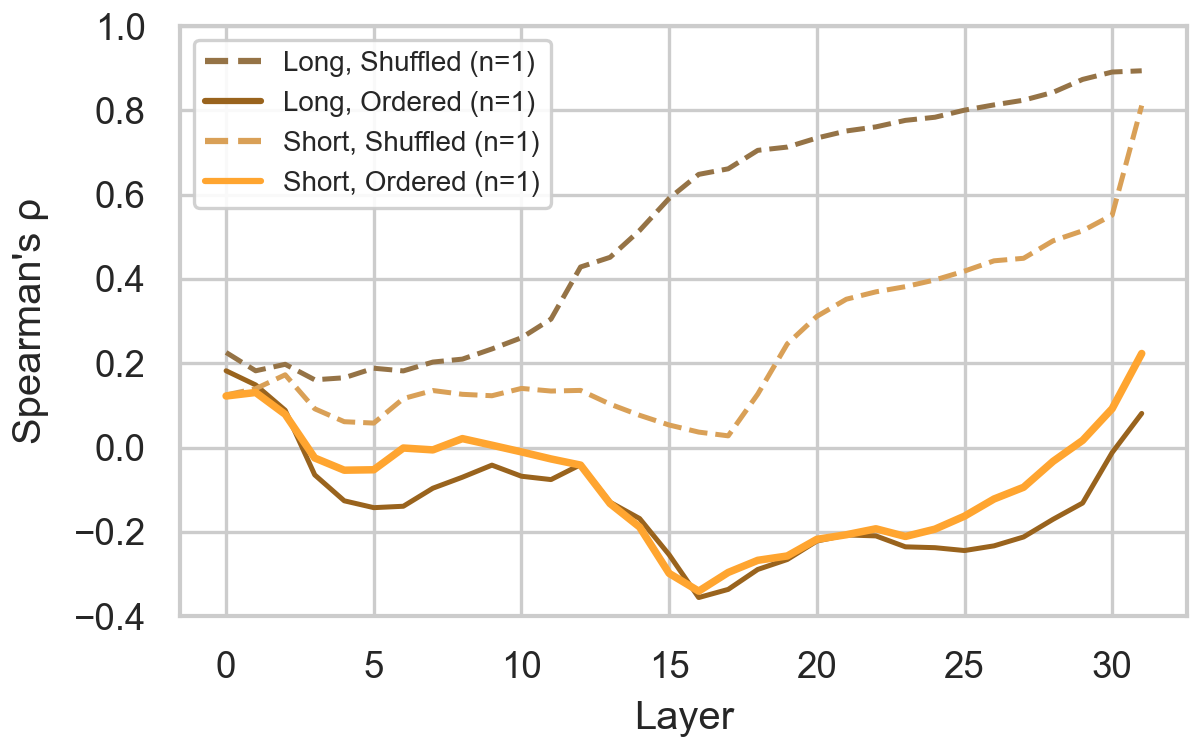}
    \caption{Mistral: Euc. distances, predictions correlation}
    \label{fig:alt-ang-mistral-euc}
  \end{subfigure}\hfill
  \begin{subfigure}[t]{0.31\textwidth}
    \centering
    \includegraphics[width=\linewidth]{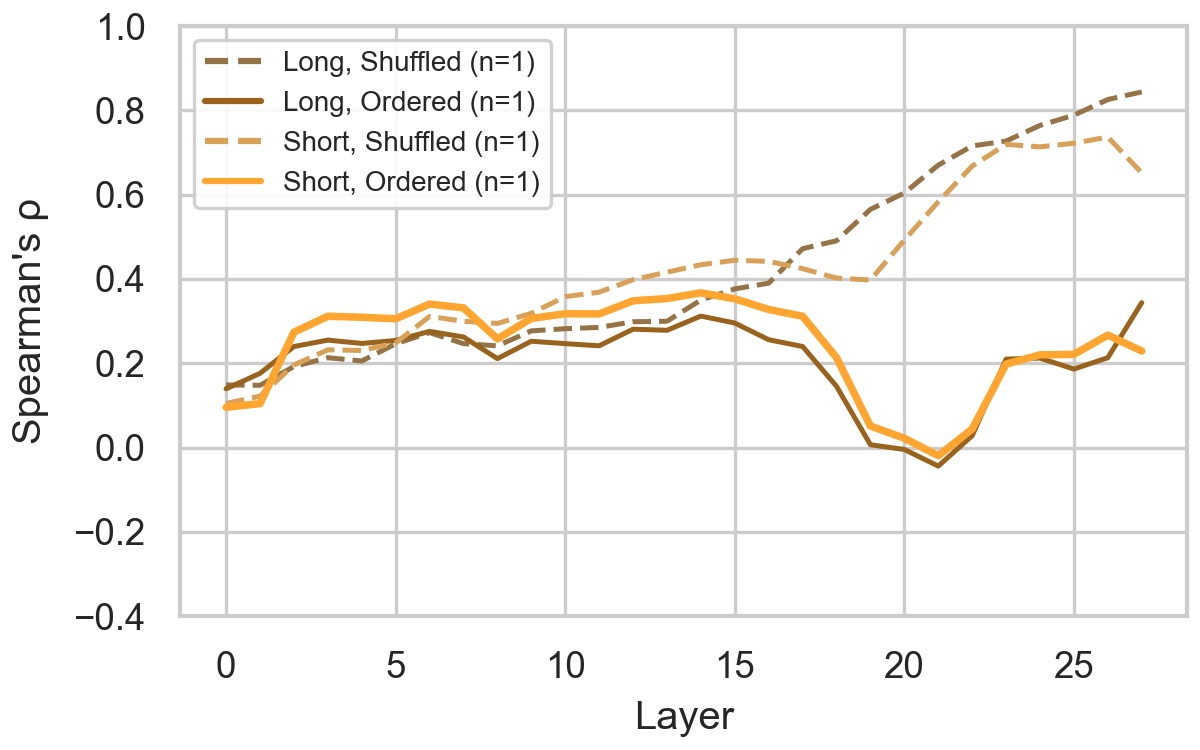}
    \caption{Qwen: Euc. distances, predictions correlation}
    \label{fig:alt-ang-qwen-euc}
  \end{subfigure}

  \caption{
    Per-layer correlation between \textit{third from last} non-identical tokens' pairwise distances and pairwise prediction-distribution symmetric KL divergence, shown across layers.
    \textbf{Top row}: angular distances.
    \textbf{Bottom row}: Euclidean distances.
    Rows correspond to distance metrics, and columns correspond to models.
    The blue vertical line marks the perturbation-based phase-change point.
  }
  
  \label{fig:third_non_identical_correlations}
\end{figure}

\end{document}